\documentclass[letterpaper,twocolumn,10pt]{article}
\usepackage{usenix2019_v3}

\usepackage{tikz}
\usepackage{amsmath}

\usepackage{filecontents}
\usepackage{enumitem}
\usepackage{subcaption}
\usepackage{booktabs}
\usepackage{amssymb}
\usepackage{makecell}
\usepackage{multirow}
\usepackage{balance}
\usepackage{hyperref}

\newcommand{\myparagraph}[1]{\vspace{4pt} \noindent \textbf{#1.}}

\DeclareMathOperator*{\argmin}{arg\,min}
\newcommand{\norm}[1]{\left\lVert#1\right\rVert}
\newcommand{\yolo}[1]{Yolov3}
\newcommand{\frcnn}[1]{Faster-RCNN}
\newcommand{\mrcnn}[1]{Mask-RCNN}
\newcommand{\lisa}[1]{Lisa-CNN}
\newcommand{\gtsrb}[1]{Gtsrb-CNN}

\newcommand\blfootnote[1]{%
	\begingroup
	\renewcommand\thefootnote{}\footnote{#1}%
	\addtocounter{footnote}{-1}%
	\endgroup
}

\begin{document}

\date{}


\title{\Large \bf SLAP: Improving Physical Adversarial Examples with Short-Lived Adversarial Perturbations}

 \author{
 {\rm Giulio Lovisotto}\\
 University of Oxford
 \and
 {\rm Henry Turner}\\
 University of Oxford
 \and
 {\rm Ivo Sluganovic}\\
University of Oxford
\and
{\rm Martin Strohmeier}\\
armasuisse
\and
{\rm Ivan Martinovic}\\
University of Oxford
}

\maketitle

\begin{abstract}

Research into adversarial examples (AE) has developed rapidly, yet static adversarial patches are still the main technique for conducting attacks in the real world, despite being obvious, semi-permanent and unmodifiable once deployed.

In this paper, we propose Short-Lived Adversarial Perturbations (SLAP), a novel technique that allows adversaries to realize physically robust real-world AE by using a light projector.
Attackers can project a specifically crafted adversarial perturbation onto a real-world object, transforming it into an AE.
This allows the adversary greater control over the attack compared to adversarial patches: (i) projections can be dynamically turned on and off or modified at will, (ii) projections do not suffer from the locality constraint imposed by patches, making them harder to detect.

We study the feasibility of SLAP in the self-driving scenario, targeting both object detector and traffic sign recognition tasks, focusing on the detection of stop signs.
We conduct experiments in a variety of ambient light conditions, including outdoors, showing how in non-bright settings the proposed method generates AE that are extremely robust, causing misclassifications on state-of-the-art networks with up to 99\% success rate for a variety of angles and distances.
We also demostrate that SLAP-generated AE do not present detectable behaviours seen in adversarial patches and therefore bypass SentiNet, a physical AE detection method.
We evaluate other defences including an adaptive defender using adversarial learning which is able to thwart the attack effectiveness up to 80\% even in favourable attacker conditions.\blfootnote{\textit{This paper was accepted at Usenix Security 2021}}
\end{abstract}

\section{Introduction}

Recent advances in computational capabilities and machine learning algorithms have led to deep neural networks (DNN) rapidly becoming the dominant choice for a wide range of computer vision tasks.
Due to their performance, DNNs are increasingly being used in security-critical contexts, such as biometric authentication or object recognition for autonomous driving. 
However, if a malicious actor controls the input to the network, DNNs are susceptible to carefully crafted adversarial examples (AE)~\cite{Szegedy2015}, which leverage specific directions in input space to create examples which whilst resembling legitimate images, will be misclassified at test time.

\begin{figure}[t]
	\centering
	\begin{subfigure}[b]{0.23\textwidth}
		\centering
		\includegraphics[width=\textwidth]{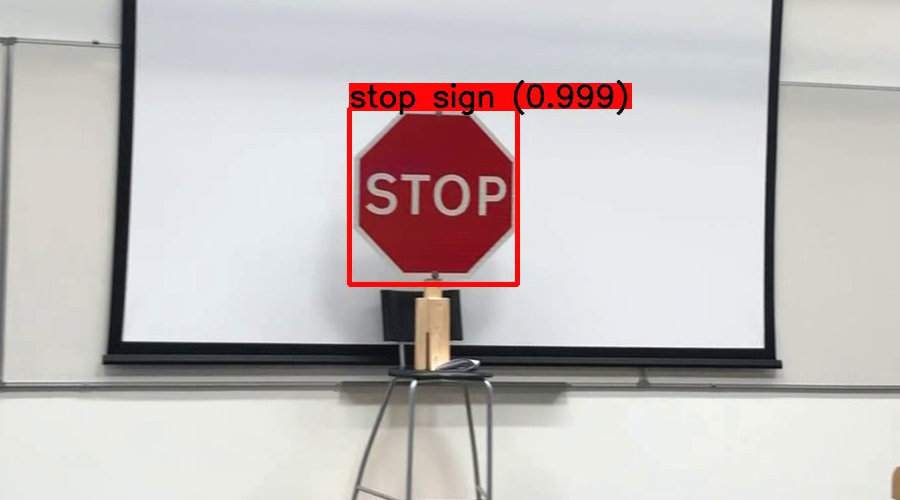}
		\caption%
		{{\small Non adversarial scenario.}}
		\label{fig:intro1}
	\end{subfigure}
	\hfill
	\begin{subfigure}[b]{0.23\textwidth}
		\centering
		\includegraphics[width=\textwidth]{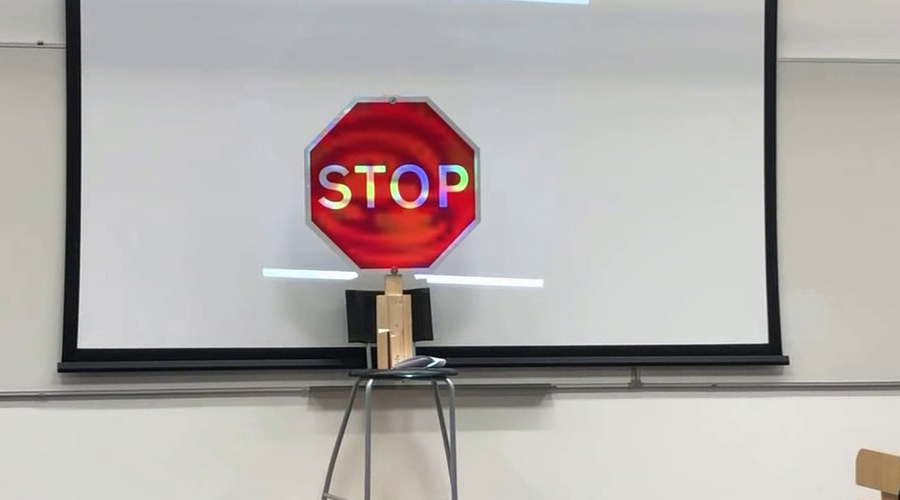}
		\caption[]%
		{{\small Adversarial projection.}}
		\label{fig:intro2}
	\end{subfigure}

	\caption{The attack visualized. A projector shines a specific pattern on the stop sign causing an object detector (\yolo{} in this picture) to misdetect the object.}
	\label{fig:introduction}
\end{figure}

A significant body of earlier research focused on analyzing AE in the digital domain, where an adversary has the capability of making pixel-specific manipulations to the input.
This concept has been further developed with the realization of physically robust AE~\cite{Sharif2016, eykholt2017robust, Eykholt2018-woot, chen2018shapeshifter, Zhao2019, 9230411}, which are examples that survive real-world environmental conditions, such as varied viewing distances or angles.
In order to realize AE, adversaries can either print patches (e.g. as stickers or glasses in the case of face recognition), or replace an entire object by overlaying the object with a printed version of it with subtle changes.
However, these techniques have multiple limitations.
Firstly, these methods typically generate highly salient areas in the network inputs, which makes them detectable by recent countermeasures~\cite{Chou2018}.
Secondly, in the autonomous driving scenario, sticking patches on a traffic sign leads to continuous misdetection of such signs, which is equivalent to removing the sign from the road or covering it.



In this paper, we focus on the autonomous driving scenario and propose using a light projector to achieve \emph{Short-Lived Adversarial Perturbations (SLAPs)}, a novel AE approach that allows adversaries to realize robust, dynamic real-world AE from a distance.
SLAP-generated AE provide the attacker with multiple benefits over existing patch-based methods, in particular giving fine-grained control over the timing of attacks, allowing them to become \textit{short-lived}.

As part of designing the SLAP attack, we propose a method to model the effect of projections under certain environmental conditions, by analyzing the absolute changes in pixel colors captured by an RGB camera as different projections are being shown.
The method consists of fitting a differentiable model, which we propagate the derivatives of the projection through during the AE crafting phase.
Our method improves the established \textit{non printability score}~\cite{Sharif2016} (NPS) used in patch-based AE by modelling a three-way additive relationship between the projection surface, the projection color, and the camera-perceived output.
Furthermore, we improve the robustness of AE in the physical world by systematically identifying and accounting for a large set of environmental changes.
We empirically analyze the feasibility of SLAP on two different use-cases: (i) object detection and (ii) traffic sign recognition.

To understand the relationship between ambient light and attack feasibility, we collect extensive measurements in different light conditions, including outdoors.
We conduct our attack on four different models: \yolo{}, \mrcnn{}, \lisa{}, and \gtsrb{}, demonstrating the attack can successfully render a stop sign undetected in over 99\% of camera frames, depending on ambient light levels.

We also evaluate the transferability of our attack, showing that depending on the model used during the AE crafting phase, SLAP could be used to conduct black-box attacks.
In particular, we show that AE generated with \mrcnn{} and \yolo{} transfer onto the proprietary Google Vision API models in up to 100\% of cases.

Finally, we evaluate potential defences.
We show that SLAP can bypass SentiNet~\cite{Chou2018}, a recent defence tailored to physical AE detection. 
Since SLAP does not present a locality constraint in the same way as adversarial patches, SLAP AE bypass SentiNet over 95\% of the time.
We investigate other countermeasures and find that an adaptive defender using adversarial learning can prevent most attacks, but at the cost of reduced accuracy in non-adversarial conditions.

\myparagraph{Contributions}
\begin{itemize}
	\item We propose SLAP, a novel attack vector for the realizability of AE in the physical world by using an RGB projector. This technique gives the attacker new capabilities compared to existing approaches, including short-livedness and undetectability.
	\item We propose a method to craft robust AE designed for use with a projector. The method models a three-way additive relationship between a surface, a projection and the camera-perceived image. We enhance the robustness of the attacks by systematically identifying and accounting for varying environmental conditions during the optimization process.
	\item We evaluate the SLAP attack on two different computer vision tasks related to autonomous driving: (i) object detection and (ii) traffic sign recognition. We conduct an extensive empirical evaluation, including in- and out-doors, showing that under favourable lighting conditions the attack leads to the target object being undetected.
	\item We evaluate countermeasures. We firstly show that SLAP AE bypass locality-based detection measures such as SentiNet~\cite{Chou2018}, which is tailored for the detection of physical AE. We then show that an adaptive defender using adversarial learning can thwart most of the attacks.
\end{itemize}

\section{Background and Related Work}

We start by introducing the necessary background on LCD projectors and object detection.
We then cover the related work in physically-realizable adversarial examples.

\subsection{Projector technology}


A common LCD (liquid crystal display) projector works by sending light through a series of dichroic filters in order to form the red, green and blue components of the projected images.
As the light passes through, individual pixels may be opened or closed to allow the light to pass~\cite{fischetti2007two}, creating a wide range of colors.
The total amount of light that projectors emit (measured in lumens), as well as the amount of light per area (measured in lux) is an important factor for determining the image quality, with stronger output leading to more accurate images in a range of conditions.
Common office projectors are in the range of 2,000-3,000 lumens of emitted light, while the higher-end projectors can achieve up to tens of thousands of lumens (e.g., the projectors used during the London 2012 Olympics~\cite{olympics2020panasonic}).
As lumens only measure the total quantity of visible light emitted from the projector, the current ambient light perceived on the projection surface has an important role in determining the formed image contrast and color quality.
The brighter the ambient light, the less visible will the image formed by a projector be due to weaker contrast and narrower range of colors.

As an example, a 2,000 ANSI lumens projector can emit enough light to obtain a light intensity of 2,000 lux on a square meter area (measured for white light~\cite{ansi-lumens}).
Such a projector would reproduce an image in an office quite well (ambient \textasciitilde500 lux), but could hardly make the image visible if it was placed outside in a sunny day (\textasciitilde18,000 lux).
Additionally, projectors are generally used and tested while projecting on a (white) projection screen, which are designed to optimize the resulting image quality.
When projecting on different materials and non-white surfaces, the resulting image will vary greatly given that light propagation significantly changes depending on the material in use and the background color.
In Section~\ref{sec:projectability} we explain how we model such changes in an empirical way that accounts for many variability factors.

\subsection{Object Detection}

Object detection refers to the task of segmenting instances of semantic objects in an image.
The output of object detectors is generally a set coordinates of bounding boxes in the input image that contain specific objects.
In the following we detail two object detectors, \yolo{}~\cite{redmon2018yolov3} and \frcnn{}~\cite{ren2015faster} which are used throughout this paper.

\yolo{} is a single-shot detector which runs inputs through a single convolutional neural network (CNN).
The CNN uses a back-bone network to compute feature maps for each cell in a square grid of the input image.
Three grid sizes are used in \yolo{} to increase accuracy of detecting smaller objects (13x13, 26x26, 52x52).
\yolo{} is used in many real-time processing systems~\cite{apolloauto2017, bmwinnovationlab2019, van2018you}.

\frcnn{} is the result of a series of improvements on the initial R-CNN object detector network~\cite{girshick2014rich}.
\frcnn{} uses a two-stage detection method, where an initial network generates region proposals and a second network predicts labels for proposals.
More recently, \mrcnn{}~\cite{he2017mask} extended \frcnn{} in order to add object segmentation to object detection.
Both \yolo{} and \mrcnn{} use non-maximum suppression in post-processing to remove redundant boxes with high overlap.

\myparagraph{Traffic Sign Recognition}
The task of traffic sign recognition consists in distinguishing between different traffic signs.
Differently from object detection, in traffic sign recognition the networks typically require a cutout of the sign as input, rather than the full scene.
Several datasets of videos from car dash cameras are available online, such as LISA~\cite{mogelmose2012vision} or GTSRB~\cite{houben2013detection}, in which a region of interest that identifies the ground-truth position of the traffic sign in each video frame is generally manually annotated.
In this paper, for continuity, we consider two different models for traffic sign recognition, \lisa{} and \gtsrb{}, both introduced in~\cite{eykholt2017robust}, one of the earliest works in real-world robust AE.

\subsection{Physical Adversarial Examples}

\begin{figure}[t]
	\centering
	\includegraphics[width=0.4\textwidth]{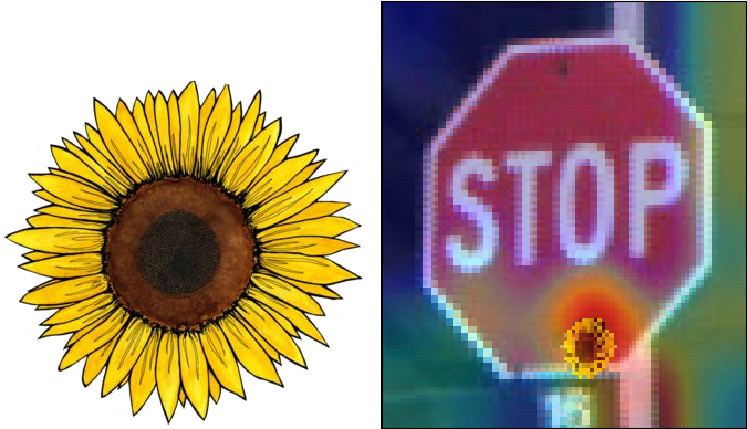}
	\caption{Example of an adversarial patch attack~\cite{Gu2017}. The network has been compromised and reacts to the sunflower being placed in the input by misclassifying the stop sign. SentiNet~\cite{Chou2018} leverages the locality of the patch to detect regions with high saliency, and can therefore detect the attack. The figure is taken from Figure~5 in~\cite{Chou2018}. }
	\label{fig:sentinet_eg}
\end{figure}

Kurakin et al.~\cite{Kurakin2016} showed that perturbations computed with the fast gradient descent~\cite{Szegedy2015} method can survive printing and re-capture with a camera.
However, these perturbations would not be realizable on a real (3D) input, therefore other works on physical attacks against neural networks have focused on adversarial patches~\cite{Brown2018, Karmon2018}.
Evtimov et al.~\cite{Eykholt2017, Eykholt2018-woot} showed how to craft robust physical perturbations for stop signs, that survive changes when reproduced in the physical world (e.g., distance and viewing angle).
The perturbation is in the form of a poster overlaid on the stop sign itself or a sticker patch that the authors apply to the sign.
Sharif et al.~\cite{Sharif2016} showed that physical AE for face recognition can be realized by using colored eye-glass frames, further strengthening the realizability of the perturbation in the presence of input noise (e.g., different user poses, limited color gamut of printers).
Although most of these attacks are focused on evasion attacks, localized perturbations have also been used in poisoning attacks~\cite{Liu2018, Gu2017} both by altering the training process or the network parameters post-training.

More recent works have focused on AE for object detection~\cite{Eykholt2018-woot, Zhao2019, chen2018shapeshifter}.
All these works use either printed posters or patches to apply on top of the traffic signs as an attack vector.
As discussed in the previous section, patches suffer from several disadvantages that can be overcome with a projector, in particular projections are short-lived and dynamic.
This allows adversaries to turn the projection on/off as they please, which can be used to target specific vehicles and allows them to leave no traces of the malicious attack.

\myparagraph{Physical AE Detection}
Differently from a digital scenario, where input changes are simply limited by $L_p$-norms, the realization of physically robust AE is more constrained.
Adversarial patches are one technique for phsyical AE, however, they have drawbacks which enable their detection.
In fact, Chou et al.~\cite{Chou2018} exploited the locality of adversarial patches to create an AE-detection method named SentiNet, which detects physical AE leveraging the fact that adversarial patches generate localized areas of high saliency in input, as shown in Figure~\ref{fig:sentinet_eg}.
These highly salient areas successfully capture the adversarial patch in input, and therefore can be used for the detection of an AE by using the fact that such salient areas will cause misclassifications when overlaid onto other benign images.
For example, Figure~\ref{fig:sentinet_eg} shows that an adversarial patch shaped as a flower will cause the stop sign to be misclassified as a warning sign.
The same flower patch can be applied to different images and will also cause misclassifications in other classes, which is an unusual behavior which can be detected.
SentiNet can capture this behavior just by looking at the saliency masks of benign images, and fitting a curve to the accepted behavior range, rather than fitting a binary model for the detection.
This way SentiNet can adapt for unseen attacks and therefore claims to generalize to different attack methods.
In this paper, we show how AE generated with SLAP can bypass such detection.

\section{Threat Model}
\begin{figure}[t]
	\centering
	\includegraphics[width=0.4\textwidth]{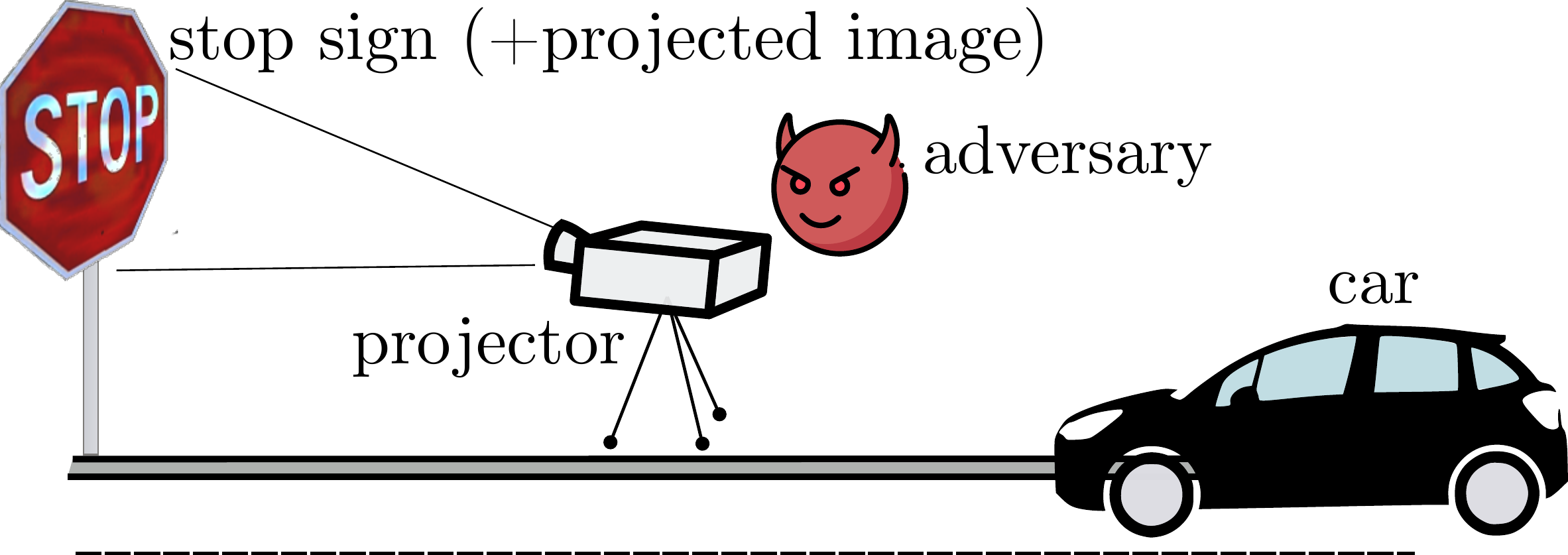}
	\caption{Attack scenario. An adversary points a projector at a stop sign and controls the projection in order to cause the sign to be undetected by an approaching vehicle.}
	\label{fig:threatmodel}
\end{figure}

We focus on an autonomous driving scenario, where cameras are placed in vehicles and the vehicle makes decisions based on the cameras' inputs.
The vehicle uses camera(s) to detect and track the objects in the scene, including traffic signs.

\myparagraph{Goal} The adversaries' goal is to cause a stop sign to be undetected by the neural networks processing the camera feeds within the car, which will cause vehicles approaching the stop sign to ignore them, potentially leading to accidents and dangerous situations.
The adversary may want to target specific vehicles, therefore using adversarial patches to stick on the stop sign is not a suitable attack vector.
Patches would lead to the stop sign always being undetected by each passing vehicle and would cause suspicion among the passengers realizing that cars did not stop because of an altered sign.

\myparagraph{Capabilities and Knowledge} The adversary has access to the general proximity of the target stop sign and can control a projector so that it points to the sign, see Figure~\ref{fig:threatmodel}.
We note that the adversary does not necessarily need to have direct physical access to the sign itself -- rather to a position from which a visual line of sight exists.
In the paper we analyze both adversaries with white-box knowledge and a black-box scenario based on the transferability of adversarial examples.

\section{Method}

In this section, we explain our method to carry out the attack.

\subsection{Modelling projectable perturbations}\label{sec:projectability}

Often, to realize physical AE, researchers use the non-printability-score introduced by Sharif.~\cite{Sharif2016}, which models the set of colors a printer is able to print.
In our case, when shining light with the projector, the resulting output color as captured by a camera depends on a multitude of factors rather than just the printer (as in NPS).
These factors include: (i) projector strength, (ii)  projector distance, (iii) ambient light, (iv) camera exposure, (v) color and material properties (diffusion, reflections) of the surface the projection is being shone on (hereafter, \textit{projection surface}).
The achievable color spectrum is significantly smaller than the spectrum available to printed stickers as a result of these factors (e.g., a patch can be black or white, while most projections on a stop sign will result in red-ish images).
In order to understand the feasibility of certain input perturbations, we model these phenomena as follows.

\myparagraph{Formalizing the problem}
We wish to create a model which, given a certain projection and a projection surface, predicts the resulting colors in output (as captured by a camera).
We describe this model $\mathcal{P}$ as follows:
\begin{equation}\label{eq:pm}
\mathcal{P}(\theta_1, S, P) = O,
\end{equation}
where $S$ is the projection surface, $P$ is the projected image, $O$ is the image formed by projecting $P$ on $S$ and $\theta_1$ are the model parameters, respectively.

Finding a perfect model would require taking all of the factors listed above into account, some of which may not be available to an adversary and is also likely to be time consuming due to the volume of possible combinations.
Therefore, we opt for a sampling approach, in which we iteratively shine a set of colors on the target surface (the target object) and collect the outputs captured by the camera.
We then fit a model to the collected data, which approximates the resulting output color for given projected images and projection surfaces.

\myparagraph{Collecting projectable colors}
We define \textit{projectable colors} for a given pixel in $S$ as the set of color which are achievable in output for that pixel given all possible projection images.
To collect the projectable colors, we do as follows:
\begin{enumerate}
	\item collect an image of the projection surface ($S$ in Eq.~\ref{eq:pm}). This is an image of the target object.
	\item select a color $c_p = [r,g,b]$, shine an image of that color $P_{c_p}$ over the projection surface, collect the output $O_{c_p}$.
	\item repeat the previous step with different colors until enough data is collected.
\end{enumerate}
In practice, with $r, g$ and $b$ $\in [0, 255]$ we choose a certain quantization per-color channel and project all possible colors consecutively, while recording a video of the projection surface.
This allows us to collect enough information about the full color space.
With this method, we found that a quantization of 127 is enough to obtain sufficient accuracy for our method, so that we only need to project $3^3=27$ colors to obtain enough data for our model.

\begin{figure}[t]
	\centering
	\includegraphics[width=0.45\textwidth]{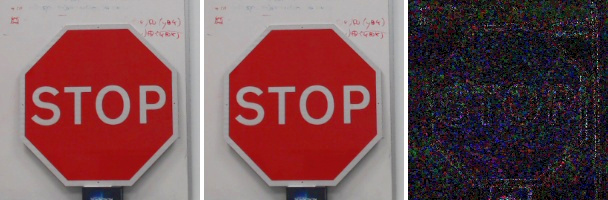}
	\caption{Camera light sensor noise visualized. The first two images show consecutive frames, while the third image shows the absolute pixel-wise difference ($\times20$) between the two frames. Such sensor noise is accounted for with smoothing over many frames during the data collection step.}
	\label{fig:camera_noise}
\end{figure}

\myparagraph{Camera noise}
In order to collect accurate data, our modelling technique has to account for noise that is being introduced by the camera.
At first, we remove noise originating from the sensitivity of the light sensor (ISO~\cite{mancuso2001introduction}), shown in Figure~\ref{fig:camera_noise}.
In fact, in non-bright lighting conditions, the camera increases the light-sensitivity of image sensor, which generates subtle pixel changes across consecutive (static) frames~\cite{exposure-triangle}.
To overcome this factor, instead of collecting individual frames for $S, P_{c_p}, O_{c_p}$, we collect 10 consecutive frames and compute and use the median of each pixel as our final image, the camera is static during this process.

\begin{figure}[t]
	\centering
	\includegraphics[width=0.475\textwidth]{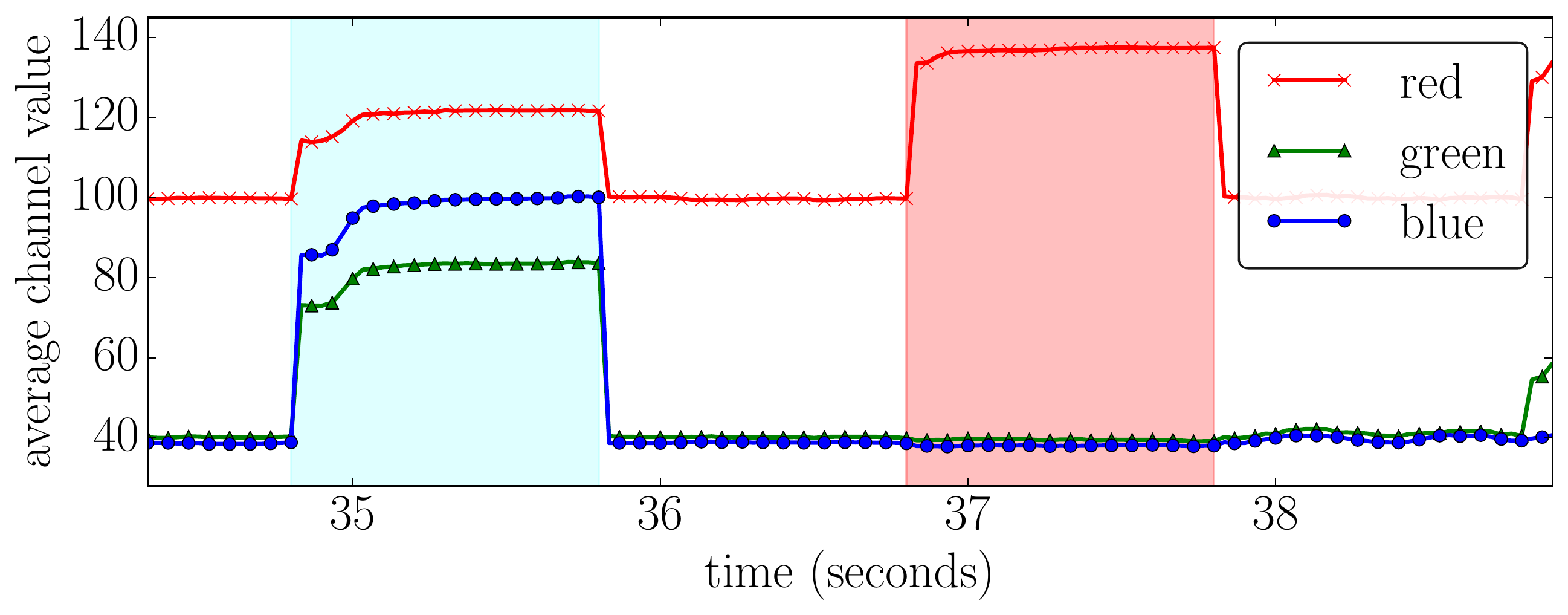}
	\caption{Plot showing how the average value of a pixel per channel (RGB) changes when a certain projection is shown. Immediately after the projection is shown, the camera requires a few frames (the lines are marked every 2 frames) to converge to a stable value. The two shaded areas cover the time the projection is being shown and are colored with the projection color.}
	\label{fig:camera_fading}
\end{figure}

Secondly, we found that there is a smoothing over-time effect in the sensor readings while recording the video, so that the sensor does not update immediately when a certain color is being shown.
Figure~\ref{fig:camera_fading} shows how the average pixel color per channel changes over time in relation to the timing of certain projections being shown.
The camera does not immediately stabilize to the resulting color when a projection is shown, but adjusts over a few frames.
To account for this adaptation, during the data collection, we interleave each projected color with 10 frames of no projection, so that the camera re-adapts to the unaltered image of the projection surface.

\begin{figure}[t]
	\centering
	\includegraphics[width=0.45\textwidth]{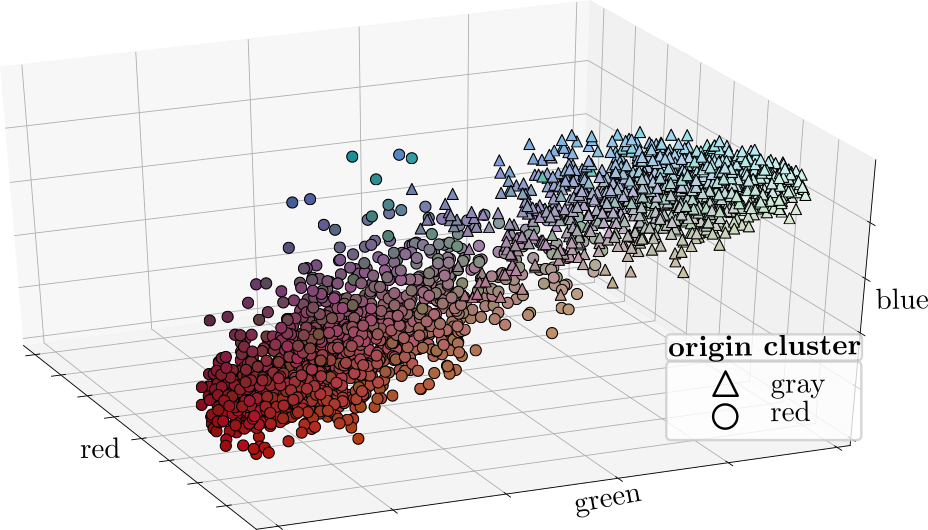}
	\caption{Plot showing the output space of the learned projection model. Each data point correspond to a color in $S$ and its color is the model output $\mathcal{P}(c_s, c_p)$ for a random $c_p$. }
	\label{fig:projection_model}
\end{figure}

\myparagraph{Fitting a projection model}
Once we have collected a set of $S, P_{c_p}, O_{c_p}$ for the chosen set of colors, we construct a training dataset as follows.
First we group together pixels of the same color by creating a mask for each unique color in the projection surface.
In other words, we find the set of unique colors present in $S$, i.e., $c_s \in S_{unq}$ and then create a mask for each color $M^{(c_s)} = \{i_j, ..., i_k\}$ such that:
\[i \in M^{(c_s)} \; \text{i.f.f.}\;\;  i^{\:th} \text{ pixel in } S == c_{s}.\]
Then, for each unique source color $c_s$, we extract all the mask-matching pixels from the output $O_{c_p}$, average their colors to get an output color $c_o^{(s, p)}$, and save the following triple for our training data $\{c_s, c_p, c_o^{(s, p)}\}$.
Such triple indicates that by projecting $c_p$ on pixels of color $c_s$ we obtained (on average) the color $c_o^{(s, p)}$.
We then use the triples to fit a neural network composed of two hidden layers with ReLU activation, we re-write Equation~\ref{eq:pm} as an optimization problem as follows:
\begin{equation}\label{eq:pm2}
 Loss_{\mathcal{P}} = \argmin_{\theta_1} \sum_{\forall c_s, c_p}\norm{\mathcal{P}(c_s, c_p) - c_o^{(s, p)}}_1,
 \end{equation}
where $\mathcal{P}$ is the model.
We optimize the network using gradient descent and Adam optimizer.
Using $\mathcal{P}$ we have a differentiable model which can be used to propagate the derivatives through it during the AE generation, see Section~\ref{sec:ae-generation}.

\myparagraph{Visualizing the Learned Model}
When the projection surface $S$ is a stop sign (as mainly investigated in this paper), pixels in $S$ generally can be separated into two clusters based on their color, corresponding to the ``red'' and ``white'' part of the sign.
The presence of these two clusters is reflected in the outputs of the projection model, as different colors will be achievable in output for the red and white parts of the stop sign.
We visualize the outputs of the projection model in Figure~\ref{fig:projection_model}, where we use a learned projection model $\mathcal{P}$, the captured source image $S$ and we compute a set of output colors for random projection colors $c_p$.
Each data point in Figure~\ref{fig:projection_model} corresponds to the color of an output pixel and is marked by a different marker (either triangle or circle) based on whether the corresponding source pixel was into the red or white cluster.
Figure~\ref{fig:projection_model} shows that the model learns a different function for red or white source pixels, obtaining in output more blue tones for white pixels while different shades of red for the remaining red pixels.

\subsection{AE Generation}\label{sec:ae-generation}

\begin{figure}[t]
	\centering
	\includegraphics[width=0.45\textwidth]{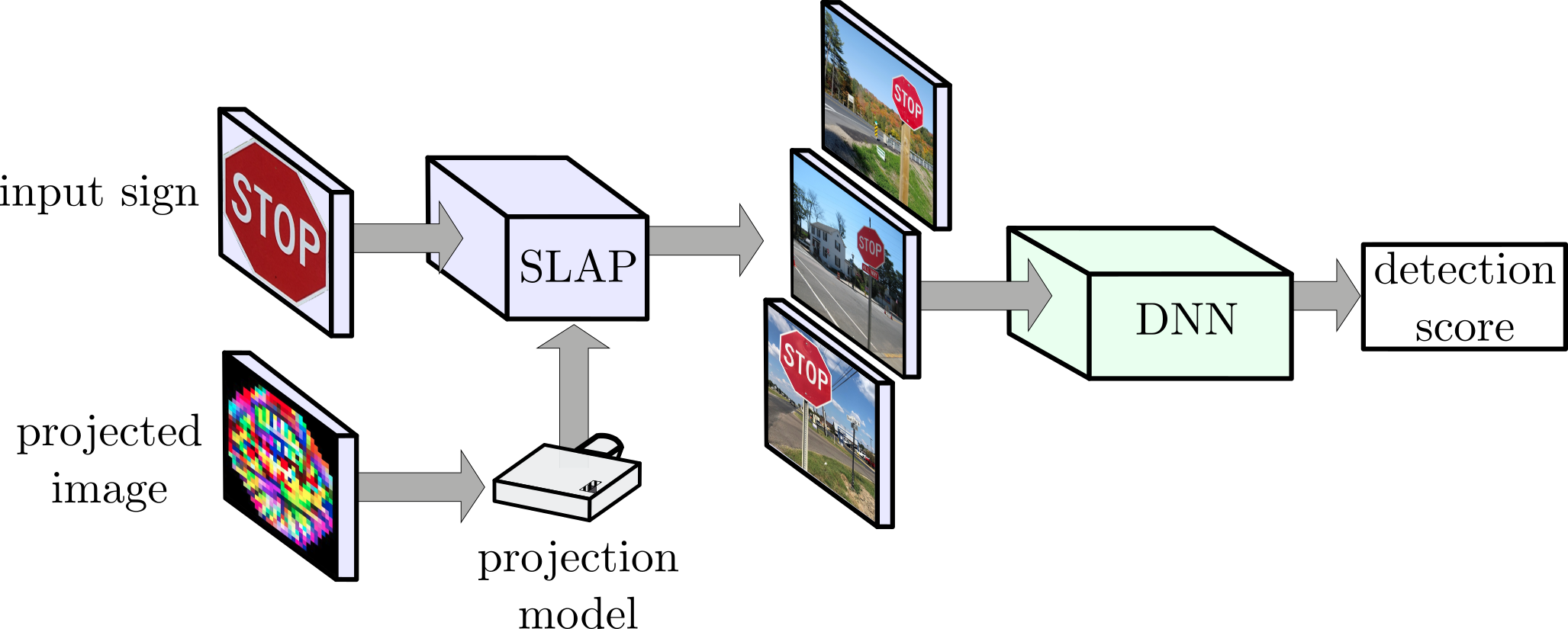}
	\caption{Overview of the adversarial samples generation pipeline. We optimize the projected image which passes through the projection model in order to minimize the target detection score on a given DNN for a set of randomly generated permutations of the input.}
	\label{fig:graph}
\end{figure}

In this section we describe our method for generating the adversarial projection.
As a starting point, we combine the projection model described in Section~\ref{sec:projectability} with the target network and use gradient descent along both to optimize the projected image.
In its basic form, we optimize the following loss function:
\[
\argmin_{\delta_{x}} \;J(f(t + \mathcal{P}(x,  \delta_{x}))) \;\;\;\text{s.t.} \;\;0\leq \delta_{x} \leq 1,
\]
where $\delta_{x}$ is the projected image, $f$ the detection network, $\mathcal{P}$ the projection model, $x$ the input image background, $x$ a stop sign image, and $J$ the detection loss, described later.
In the following we describe how we augmented the loss function in order to facilitate the physical feasibility of the adversarial perturbation and the convergence of the optimization.

\myparagraph{Physical Constraints}
In order to maintain the physical realizability of the projection, we do the following.
At first, we restrict the granularity of the projection in a fixed grid of $n\times n$ cells, so that each cell contains pixels of the same color.
This allows us to use the same projection for different distances of viewing the stop sign.
Secondly, we include the \textit{total variation} of the projection in the loss function in order to reduce the effect of camera smoothing and/or blurring~\cite{Mahendran2015}.

\myparagraph{Variable Substitution}
Since the optimization problem for the projection is bounded in [0,1] (space of RGB images) to ease the flowing of gradients when backpropagating we remove this box constraint.
Given the image to project $\delta_{x}$,
we substitute $\delta_{x}$ with a new variable $w$ such that
\[w = \frac{\tanh \delta_{x}}{2}+0.5\]
and instead optimize for $w$.
Since $\tanh\delta_{x}$ is bounded in $[-1, 1]$ we find that this substitution leads to faster convergence in the optimization.

\myparagraph{Loss Function}
We also limit the amount of perturbation in our loss  so that our final optimization looks as follows:
\[
\argmin_{w} \;J(f(t + \mathcal{P}(x,  w)))+\lambda\norm{\mathcal{P}(x, w) - x}_p + \text{TV}(w),
\]
where $\lambda$ is  a parameter used to control the importance of the $p$-norm $\norm{\cdot}_p$ and TV is the total variation described above.
Since we operate on both object detectors and traffic sign recognizers, we use two different losses $J$ depending on the target network.
For object detectors, we consider that the network returns a finite set of boxes $b \in B$ where for each box there is an associated probability output of the box containing a semantic object of class $j$, i.e.,  $p^{(b)}_j$.
For traffic sign recognizers, the network returns a probability vector containing the probability of the input image being traffic sign of class $j$, i.e., $p_j$.
We then use the following loss functions in the two cases:
\begin{itemize}
	\item \textbf{Object Detectors:} the loss is the sum of the detection probabilities for stop signs, i.e., $\sum_{b \in B} p^{(b)}_j $;
	\item \textbf{Traffic Sign Classification:} the loss is the probability for the stop sign class $p_j$.
\end{itemize}

\subsection{Training Data Augmentation}~\label{sec:training_data_augmentation}

Generating adversarial examples that work effectively in the physical world requires taking into account different environmental conditions.
Adversarial examples computed with straightforward approaches such as in~\cite{Szegedy2015} do not survive different viewing angles or viewing distances~\cite{Eykholt2018-woot}.
In order to enhance the physical realizability of these samples, different input transformations need to be accounted for during the optimization.
We use the \textit{Expectation over Transformation} (EOT) method~\cite{eykholt2017robust}, which consists in reducing the loss over a set of training images computed synthetically.
These training images are generated using linear transformations of the desired input, i.e., an image containing stop signs, so that different environmental conditions can be accounted for during the optimization.
Using EOT, our final loss becomes:
\begin{equation}\label{eq:optimization}
\begin{split}
Loss_f = \argmin_{w} \; \mathbb{E}_{t_i\sim T, m_j\sim M} \;\; J(f(t_i + m_j \cdot \mathcal{P}(x,  w))) \\
+\lambda\norm{\mathcal{P}(x, w) - x}_p + \text{TV}(w),
\end{split}
\end{equation}
where $T$ is a distribution over several background images and $M$ is an alignment function that applies linear transformations to the perturbed sign.
In this work, we augment the set of the transformations to account for additional environmental conditions that are disregarded in previous work.

\myparagraph{Background and Traffic Sign Post} Similarly to~\cite{Zhao2019} we select a set of road backgrounds and carefully place the stop sign on a post at the edge of the road. In~\cite{Zhao2019} it is shown that the post provides useful information to the detector and should therefore be included when crafting the adversarial perturbation.

\myparagraph{Perspective} We vary the angle at which the camera is looking at the stop sign. Since we do not want to account for all perspective transforms, we use the following observations.
Firstly, a traffic sign is mostly placed on one side of the lane (to the right in right-driving countries), meaning that rarely a camera mounted on a car would see a sign on the left-part of the frame.
Secondly, traffic signs are mounted at specific heights (e.g., 5 or 7 feet in the US~\cite{manual-traffic-control}),  which normally exceed the height of cars for better visibility.
Given these two observations, we prioritize perspective transforms that match these conditions.

\myparagraph{Distance} As the car is approaching the stop sign, the sign will appear with different sizes in the camera frame. Our goal is for the car to misclassify the stop sign in every frame, therefore we place stop signs with different sizes during the optimization. We test the detection of the stop sign in non-adversarial settings with decreasing stop sign sizes and we set the minimum size of the sign to be the smallest size at which the sign is detected with high confidence. In other words, we only optimize for signs sizes that are large enough to be detected by the classifier.

\myparagraph{Rotation}
As shown in~\cite{Engstrom2019}, simple rotations may lead to misclassifications when those transformations are not captured in the training dataset. We therefore add rotation to the stop sign when crafting the adversarial perturbation.

\myparagraph{Brightness}
The color of the stop sign changes based on a combination of ambient light and camera settings, e.g., in sunny days the colors appear brighter to the camera.
To account for this, we apply different brightness transformations to the stop sign, so that we include a wider range of color tones.
Since different colors contribute differently to an image brightness, we transform the stop sign image from RGB to YCrCb format~\cite{jpegcolorspace1992}, increase the luma component (Y) by a specified delta and then bring the image back into RGB.

\myparagraph{Camera Aspect Ratio}
We observe that popular object detectors resize the input images to be squared before being processed by the network (e.g., Yolov3 resizes images to 416x416 pixels), to speed up the processing.
However, the typical native aspect ratio of cameras, i.e., the size of the sensor, is 4:3 (e.g., the Aptina AR0132 chip used in the front-viewing cameras by Tesla, has a resolution of 1280x960~\cite{AR0132AT}).
This leads to objects in the frames to being distorted  when the frames are resized to squared.
To account for this distortion, we choose the dimension of the stop sign so that its height is greater than its width, reflecting a 4:3 to 1:1 resizing.

\subsection{Remarks}

We use AdamOptimizer to run the AE generation.
We optimize a single variable that is the image to project with the projector (its substitute, see Section~\ref{sec:ae-generation}).
We use batches of size 20.
All the training images are created synthetically by placing a stop sign on a road background and applying the transformations described in the previous section.
We do not use a fixed pre-computed dataset, a new batch with new images is created after every backpass on the network.
The parameters for the transformations are chosen uniformly at random in the ranges shown in Table~\ref{tab:parameters}.
For all operations that require resizing, we use cubic interpolation, finding that it provides more robust results compared to alternatives in this use-case.
We run the optimization for 50 epochs, in one epoch we feed 600 generated images containing a stop sign in the network.
For each epoch we optimize the 20\% worst-performing batches by backpropagating twice, convergence is usually reached before the last epoch.
Compared to similar works~\cite{Zhao2019}, our method runs significantly faster requiring only 50 modifications of the perturbation (compared to 500), which takes less than 10 minutes on an NVIDIA Titan V GPU for \yolo{}.

\begin{table}[t]
	\small
	\centering
	\begin{tabular}{c@{\hskip6pt}|c@{\hskip6pt}c@{\hskip6pt}c@{\hskip6pt}c}
		\toprule
		\textit{Parameter} & \yolo{} & \mrcnn{} & \lisa{} & \gtsrb{} \\
		\midrule
		learning rate & 0.005 & 0.005 & 0.05 & 0.05\\\midrule
		brightness & \multicolumn{4}{c}{$[-13, +13]$ (with range [0, 255])} \\\midrule
		perspective & \multicolumn{4}{c}{$x$-axis $[-30^{\circ}, +30^{\circ}]$, \hspace{.1cm} $y$-axis $[-30^{\circ}, +30^{\circ}] $} \\\midrule
		rotation & \multicolumn{4}{c}{$[-5^{\circ}, +5^{\circ}]$}\\\midrule
		aspect ratio& \multicolumn{4}{c}{from 4:3 to 16:9} \\\midrule
		sign size & \multicolumn{4}{c}{[25, 90] pixels} \\\midrule
        grid size & \multicolumn{4}{c}{$25\times25$} \\
		\bottomrule

    \end{tabular}
	\caption{
		Parameters used for the AE generation and the training data augmentation. The values for brightness, perspective, rotation, aspect ratio indicate the ranges for the applied transformations. All parameters are picked uniformly at random  (with the exception of perspective) during the AE generation for each sample in the generated training data.
	}
	\label{tab:parameters}
\end{table}

\section{Evaluation}\label{sec:evaluation}

In this section, we test the feasibility of the attack in real-world settings.

\subsection{Experimental Setup}\label{sec:experimental_setup}

\myparagraph{Projector Setup}
To test our projection, we buy a real stop sign of size 600x600mm.
For all of our experiments, we use a Sanyo PLC-XU4000 projector~\cite{sanyoprojector}, which is a mid-range office projector (roughly \$1,500) with 4,000 maximum lumens.
We carry out the experiment in a large lecture theatre in our institution.
We measure the projector light intensity with a Lux Meter Neoteck, following the 9-points measuring procedure used to measure ANSI lumens~\cite{ansi-lumens}, which reports that in default settings the projector emits around 2,200 lumens.
For the experiments, we place the projector 2 meters away from the stop sign, which, at maximum zoom, allows us to obtain roughly 800 lux of (white) light on the stop sign surface.
We use this 800 lux white value to make considerations on the attack feasibility in Section~\ref{sec:discussion}.
A similar amount of projected light can be obtained from greater distances by using long throw projectors, available for few thousand dollars (e.g., \$3,200 for Panasonic PT AE8000~\cite{panasonicprojector}, see Section~\ref{sec:discussion}).
We align the projection to match the stop sign outline by transforming the perspective of the image.

\begin{table}[t]
	\small
	\centering
	\begin{tabular}{cccc@{\hskip9pt}c@{\hskip9pt}c@{\hskip9pt}c}
		& & & \multicolumn{4}{c}{$Loss_f$} \\
		\textit{lux} & \makecell{camera\\exposure (ms)} & $Loss_{\mathcal{P}}$  & \rotatebox{90}{\small{\yolo{}}} & \rotatebox{90}{\small{\mrcnn{}}} & \rotatebox{90}{\small{\gtsrb{}}} & \rotatebox{90}{\small{\lisa{}}} \\\toprule
\textbf{120}  & 33  & 0.020 & 0.09 & 0.08 & 0.01 & 0.06 \\\midrule
\textbf{180}  & 25  & 0.023 & 0.11 & 0.52 & 0.00 & 0.07 \\\midrule
\textbf{300}  & 18  & 0.017 & 0.68 & 0.86 & 0.89 & 1.03 \\\midrule
\textbf{440}  & 12  & 0.015 & 1.44 & 4.24 & 5.31 & 2.45 \\\midrule
\textbf{600}  & 9  & 0.011 & 1.80 & 5.92 & 9.12 & 8.16 \\
\bottomrule

	\end{tabular}
	\caption{Preliminary results for the various light settings considered in the experiment. The camera exposure is the exposure of the camera used for profiling (set automatically). The table shows the optimization losses: $Loss_{\mathcal{P}}$ refers to the loss in Equation~\ref{eq:pm2}, while $Loss_{f}$ refers to the loss in Equation~\ref{eq:optimization}.
	}
	\label{tab:prelim_results}
\end{table}

\myparagraph{Ambient Light}
As mentioned in Section~\ref{sec:projectability}, the amount of ambient light limits the control on the input space for the adversary.
In fact, as the ambient light increases, fewer colors are achievable as the projector-emitted light becomes less  in the resulting appearance of the sign.
To account for different ambient light levels, we conduct our experiments indoor and we control the amount of light hitting the stop sign (Section~\ref{sec:indoor_results}).
We further evaluate the attack outdoors with a road driving test (Section~\ref{sec:road_driving_test}).
To reproduce various light settings indoors, we use both the ceiling lights mounted in the indoor hall and by using an additional 60 Watts LED floodlight pointed at the sign.
We measure the attack in five different light settings: 120, 180, 300, 440 and 600 lux.
The darker setting (120 lux) corresponds to slightly dimming the ceiling lights only.
The 180 lux setting corresponds to the normal indoor lighting found in the lecture theatre where we carry out the measurements.
Higher settings are achieved by adding the LED floodlight pointed directly at the sign at different distances (from roughly 4m away at 300 lux to <2m away at 600 lux).
For reference, on a clear day at sunrise/sunset the ambient light is roughly 400 lux, while on an overcast day at the same hours there are roughly 40 lux~\cite{wikipedia-lux}.

\myparagraph{Networks and Detection Thresholds}
We consider four different networks in our experiments: two object detectors, (1) \yolo{} and (2) \mrcnn{}, two traffic sign recognizers, (3) \lisa{} and (4) \gtsrb{}.
For \yolo{}, we use the Darknet-53 backbone of the original paper~\cite{redmon2018yolov3}.
For \mrcnn{}, we use Resnet-101 as a backbone and feature pyramid network~\cite{lin2017feature} for the region proposals.
We download the weights for \lisa{} and \gtsrb{} from the GitHub~\cite{eykholt2017github} of the paper authors~\cite{eykholt2017robust}.
As \mrcnn{} and \yolo{} return a list of boxes with a confidence score threshold for the output class, we set the threshold for detection at 0.6 and 0.4 respectively (i.e., we count detection as "there is a box labeled stop sign with score higher than \textit{x}").
These are the thresholds that bring the highest mean Average Precision (mAP) in the coco object detection benchmark~\cite{lin2014microsoft}.
For \lisa{} and \gtsrb{} we set the detection threshold as 0.5.
The input images are resized to 416x416 for \yolo{} and \mrcnn{} and to 32x32 for \lisa{} and \gtsrb{}.

\begin{figure}[t]
	\centering
	\includegraphics[width=0.46\textwidth]{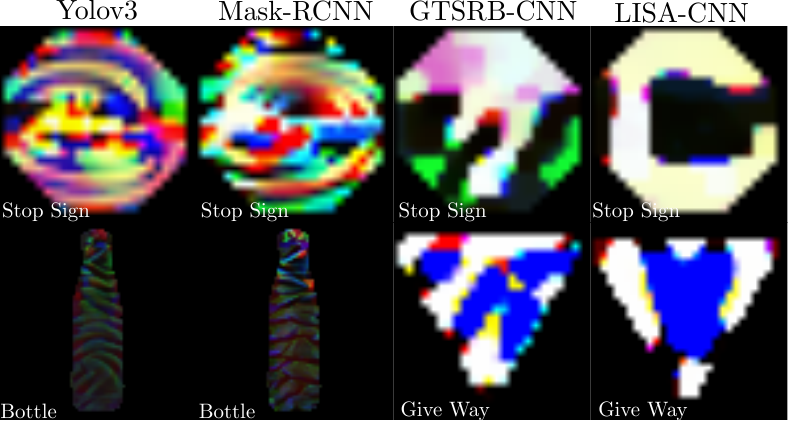}
	\caption{Examples of the projected images computed with the optimization. Bottom-right of each image specifies the target class fed to Equation~\ref{eq:optimization}. These images are computed within the 180 lux setting.}
	\label{fig:used_projections}
\end{figure}

\begin{figure*}[t]
	\centering
	\begin{subfigure}[b]{0.22\textwidth}
		\centering
		\includegraphics[width=\textwidth]{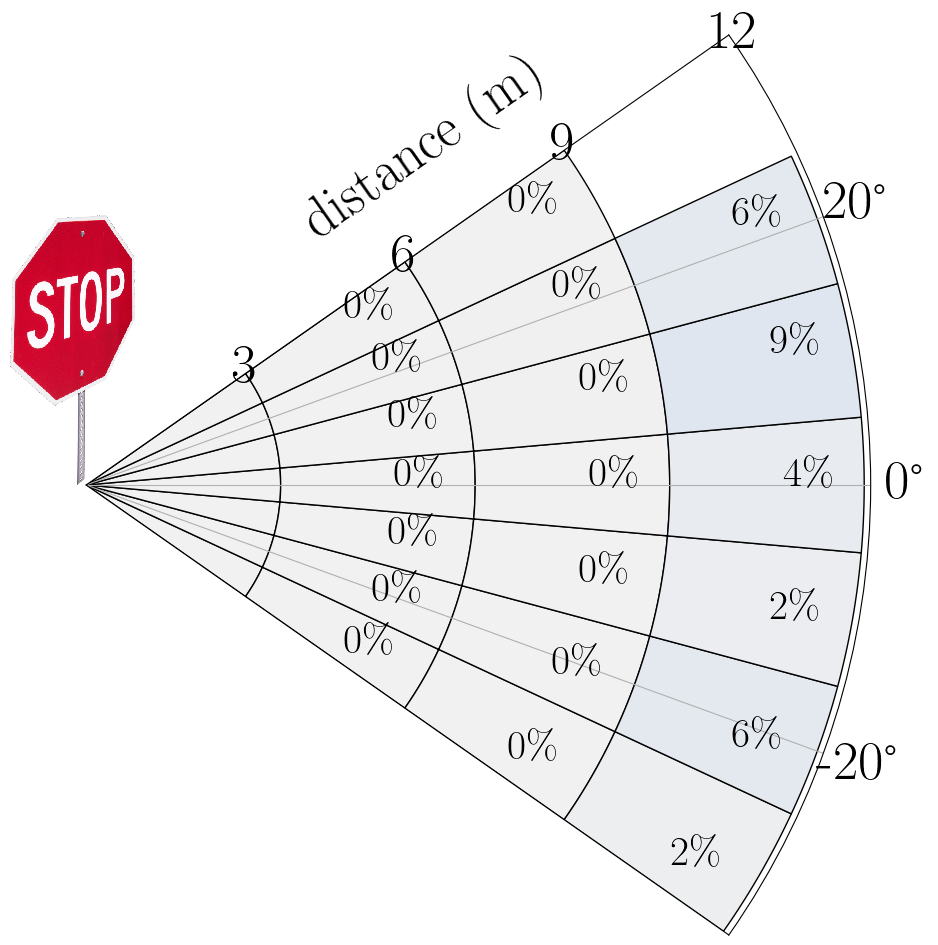}
		\caption%
		{{\small \yolo{}.}}
		\label{fig:baseline_cone_yolov3}
	\end{subfigure}
	\hfill
	\begin{subfigure}[b]{0.22\textwidth}
		\centering
		\includegraphics[width=\textwidth]{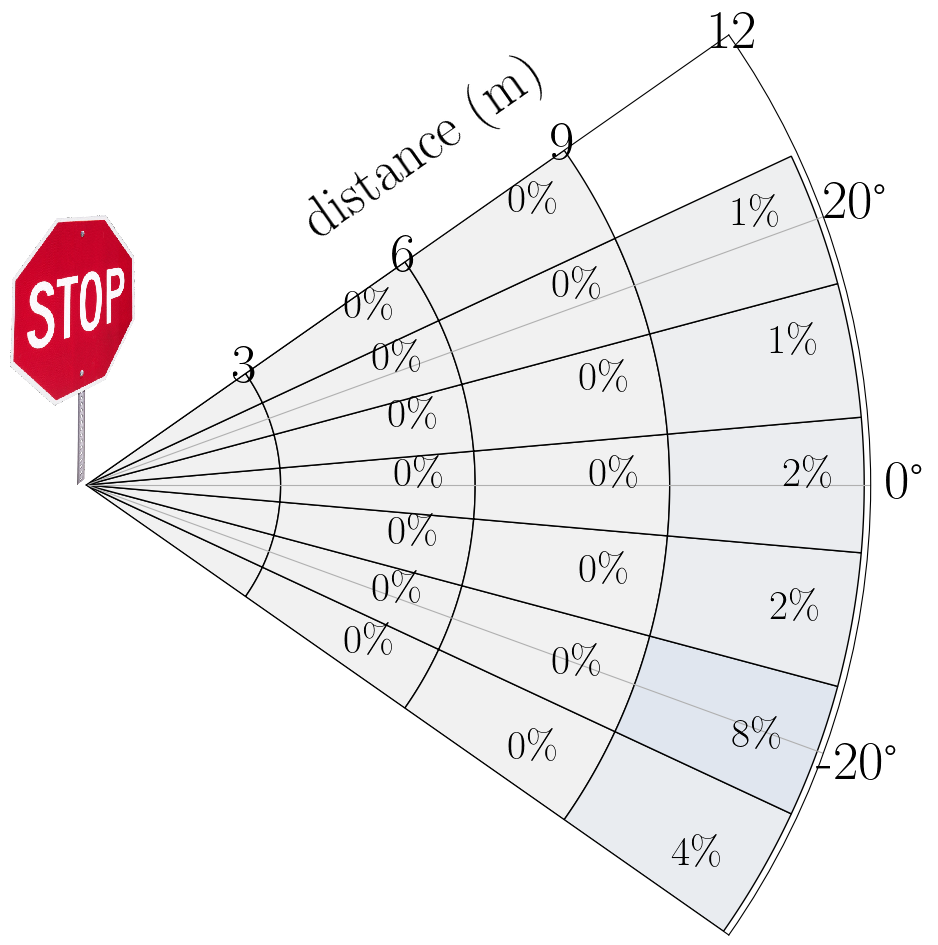}
		\caption[]%
		{{\small \mrcnn{}.}}
		\label{fig:baseline_cone_mrcnn}
	\end{subfigure}
	\hfill
	\begin{subfigure}[b]{0.22\textwidth}
		\centering
		\includegraphics[width=\textwidth]{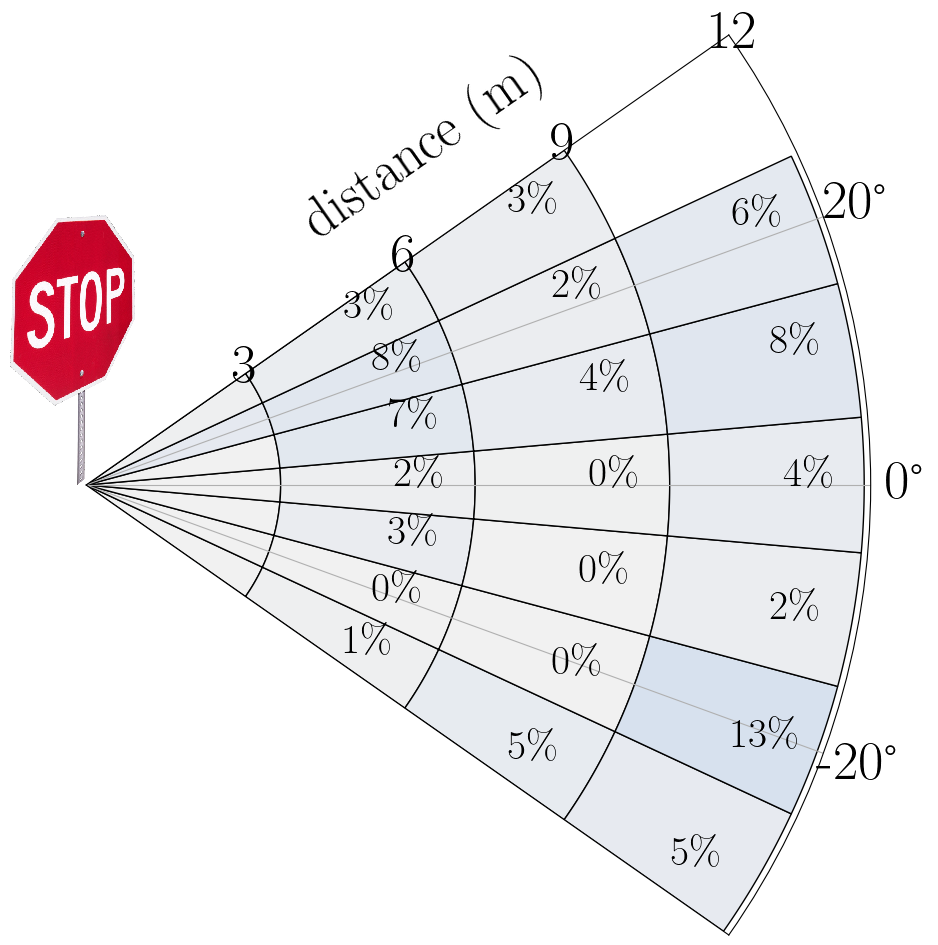}
		\caption[]%
		{{\small \gtsrb{}.}}
		\label{fig:baseline_cone_gtsrbcnn}
	\end{subfigure}
	\hfill
	\begin{subfigure}[b]{0.22\textwidth}
		\centering
		\includegraphics[width=\textwidth]{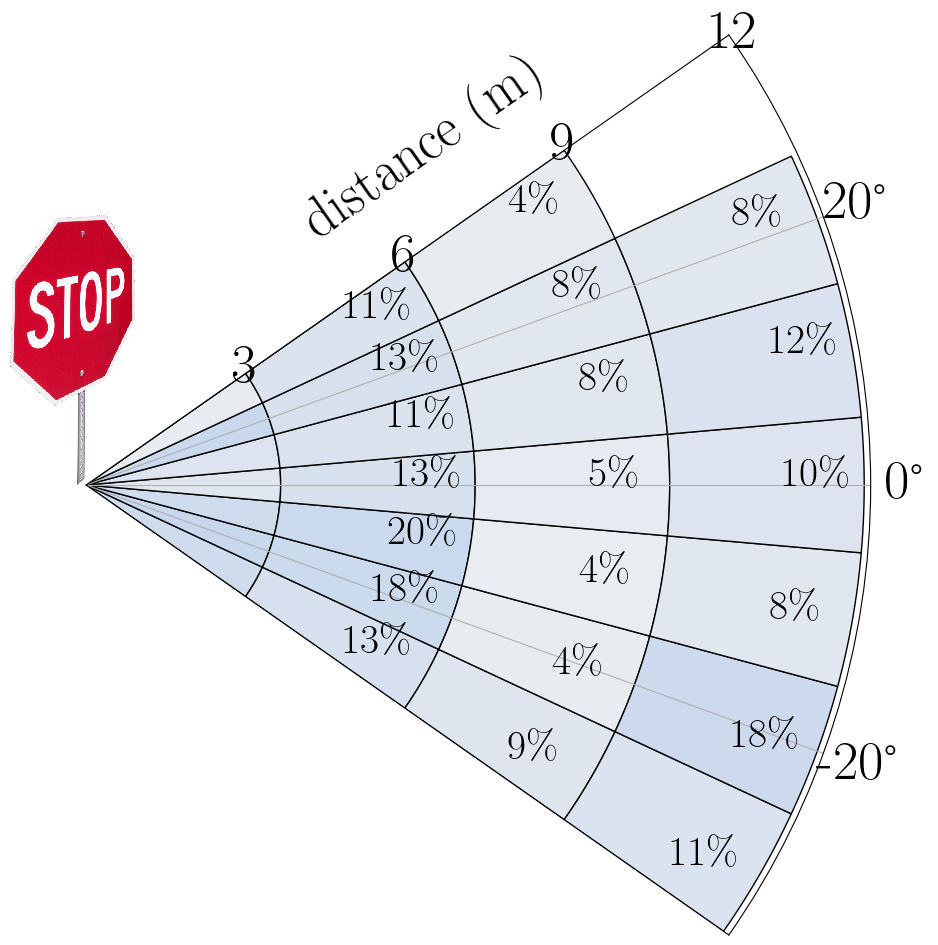}
		\caption[]%
		{{\small \lisa{}.}}
		\label{fig:baseline_cone_lisacnn}
	\end{subfigure}

	\caption{Baseline  mis-detection rate in absence of the adversary for the 180 lux setting at different angles, distances and for different networks, as the percentage of frames where a stop sign is \textbf{not} detected. Brighter shades represent higher detection rates. Percentages for 0-3m are omitted for clarity, but the corresponding cone section is colored accordingly.}
	\label{fig:baseline_cone}
\end{figure*}

\myparagraph{Metrics and Measurements}
For object detectors (\yolo{} and \mrcnn{}), we feed each frame into the network and we count how many times a stop sign is detected in the input.
For traffic sign recognizer (\gtsrb{} and \lisa{}), the network expects a cutout of a traffic sign rather than the full frame.
In order to obtain the cutout, we manually label the bounding box surrounding the stop sign and use a CSRT tracker~\cite{lukezic2017discriminative} to track the stop sign over the frames.
We then count how often the predicted label is a stop sign.
In order to monitor viewing angle and distance from the sign, we reconstruct the angle of view and distance based on the distortion on the octagonal outline of the sign and our recording camera field-of-view.
We use the default camera app on an iPhone X to record a set of videos of the stop sign at different distances and angles, with the projection being shone.
The iPhone is mounted on a stabilizing gimbal to avoid excessive blurring.
As mentioned in Section~\ref{sec:training_data_augmentation}, to match the 4:3 aspect ratio, we crop the 1080p video from the iPhone X (which has a resolution of 1920x1080) to 1440x1080 by removing the sides.

\myparagraph{Experimental Procedure}
Experiments follow this pipeline:
\begin{itemize}[topsep=3pt,itemsep=-1ex,partopsep=1ex,parsep=1ex]
	\item \textbf{Step 1:} We setup the stop sign and measure the amount of lux on the stop sign surface;
	\item \textbf{Step 2:} We carry out the profiling procedure to construct a projection model (Section~\ref{sec:projectability}); this uses a separate Logitech C920 HD Pro Webcam rather than the iPhone X camera (on which the attack is later evaluated).
	\item \textbf{Step 3:} We use the projection model to run the AE generation (Section~\ref{sec:ae-generation}) and optimize the image to project;
	\item \textbf{Step 4:} We shine the image on the sign and we take a set of videos at different distances and angles.
\end{itemize}
Unless otherwise stated, the parameters used for the optimization (Step 3) are those of Table~\ref{tab:parameters}.
Recording the profiling video of Step 2 requires less than 2 minutes, so does fitting the projection model.

\myparagraph{Preliminary Results}
Table~\ref{tab:prelim_results} shows parameters and resulting value of the loss functions at the end of the optimizations for the various light settings.
The table shows that our projection model fits the collected color triples: $Loss_{\mathcal{P}}<0.03$ shows that the error in the predicted colors, which are in [0, 1] color space, is less than 1\% per channel.
As expected, we found that the results of the optimization match the reduced capability to reproduce colors for higher ambient light settings: $Loss_f$ goes from roughly 0 to higher values as the light increases.
Note that the reported $Loss_f$ is summed over the batches of 20 and computed before non-maximum suppression.
We report in Figure~\ref{fig:used_projections} examples of projected images output of the optimization process, for three different target objects: stop sign, give way sign and bottle.
Whilst we limit the rest of the experiments to attacks on stop signs, we report considerations and results on generalizing the attack to various objects in
 Appendix~\ref{sec:app:other_objects} and in Section~\ref{sec:discussion}.

\newcommand{\mysize}{0.23}

\subsection{Indoor Results}\label{sec:indoor_results}

\begin{figure*}[h]
	\centering
	\begin{subfigure}[t]{\mysize\textwidth}
		\centering
		\includegraphics[width=\textwidth]{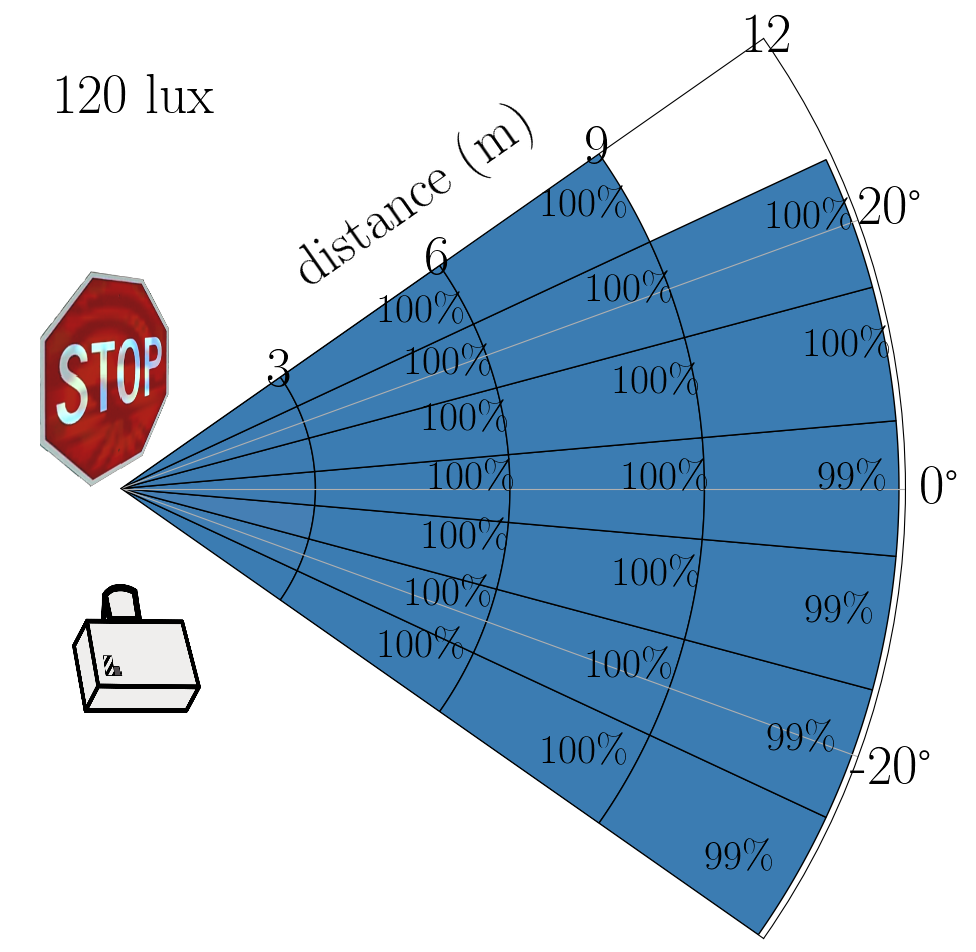}
		\caption%
		{{\small \yolo{}.}}
		\label{fig:120_cone_yolov3}
	\end{subfigure}
	\hfill
	\begin{subfigure}[t]{\mysize\textwidth}
		\centering
		\includegraphics[width=\textwidth]{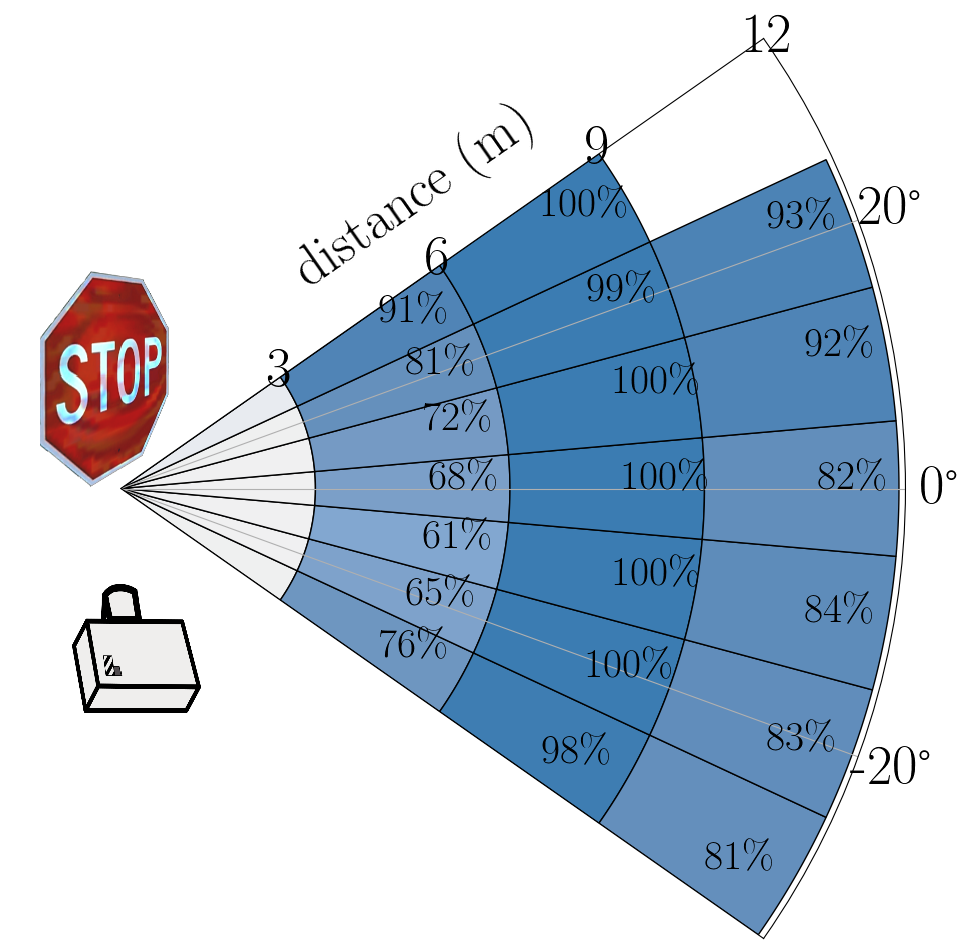}
		\caption[]%
		{{\small \mrcnn{}.}}
		\label{fig:120_cone_mrcnn}
	\end{subfigure}
	\hfill
	\begin{subfigure}[t]{\mysize\textwidth}
		\centering
		\includegraphics[width=\textwidth]{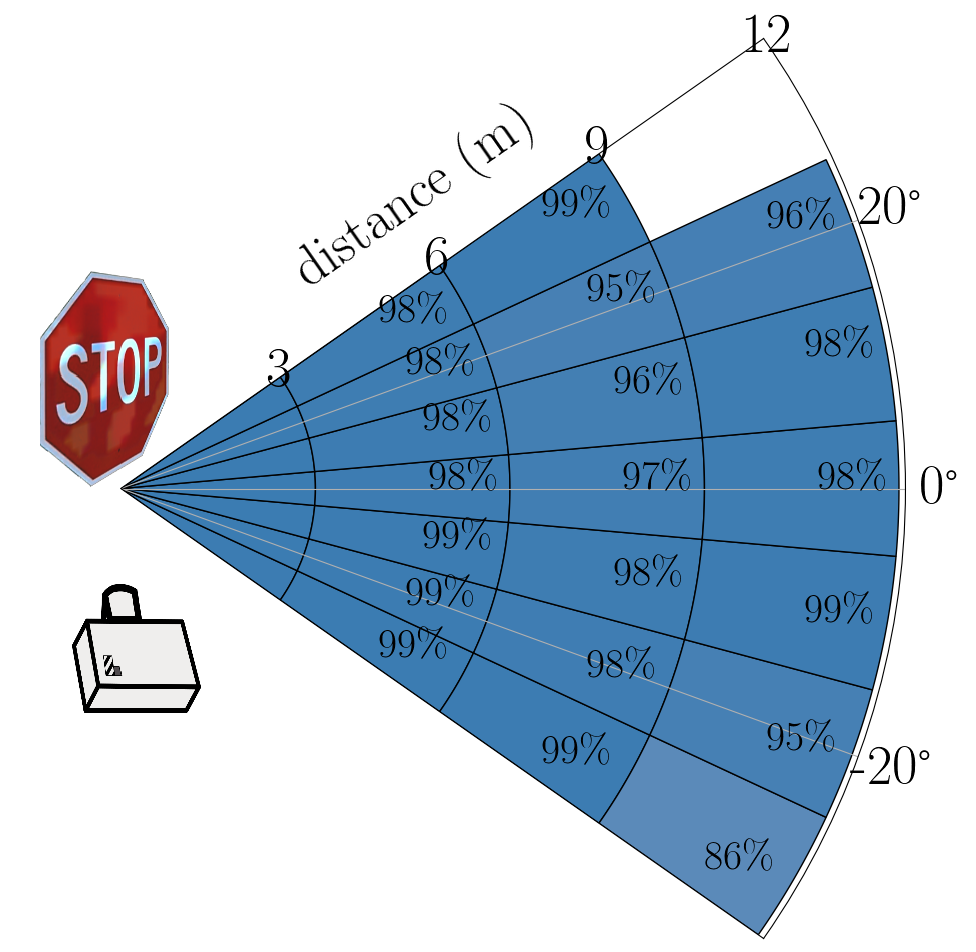}
		\caption[]%
		{{\small \gtsrb{}.}}
		\label{fig:120_cone_gtsrbcnn}
	\end{subfigure}
	\hfill
	\begin{subfigure}[t]{\mysize\textwidth}
		\centering
		\includegraphics[width=\textwidth]{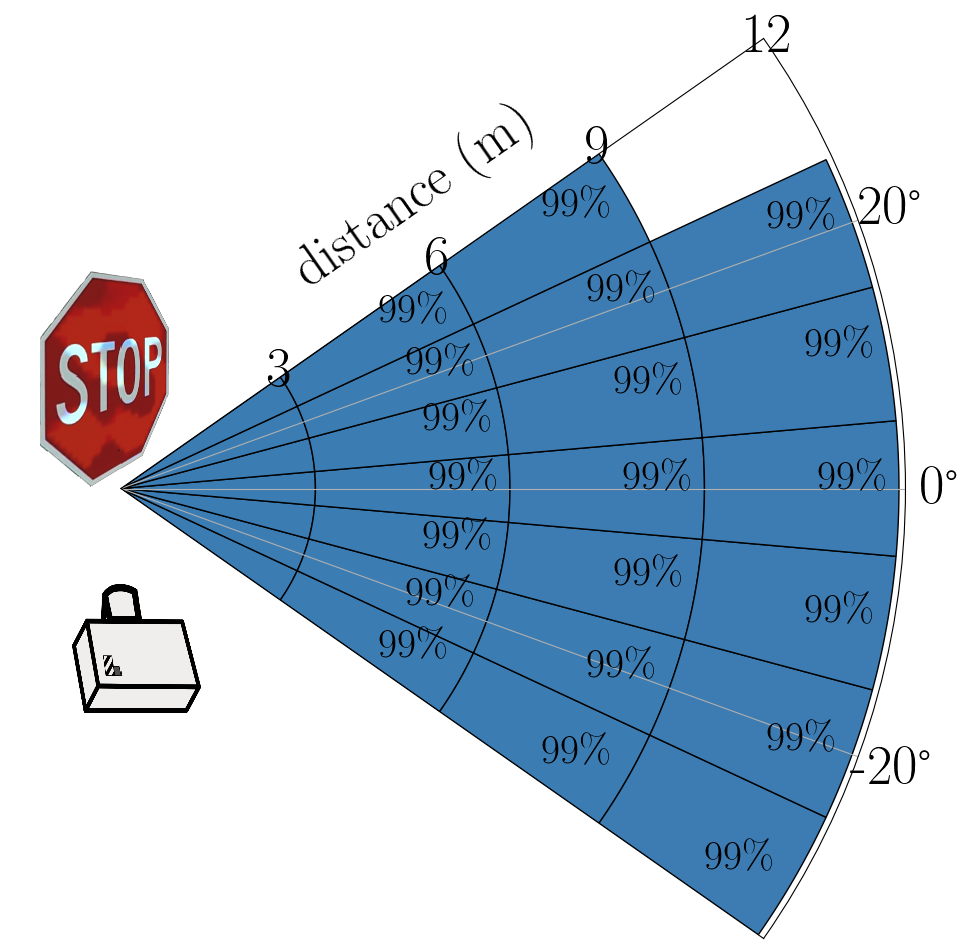}
		\caption[]%
		{{\small \lisa{}.}}
		\label{fig:120_cone_lisacnn}
	\end{subfigure}
	\hfill
	
	\begin{subfigure}[t]{\mysize\textwidth}
		\centering
		\includegraphics[width=\textwidth]{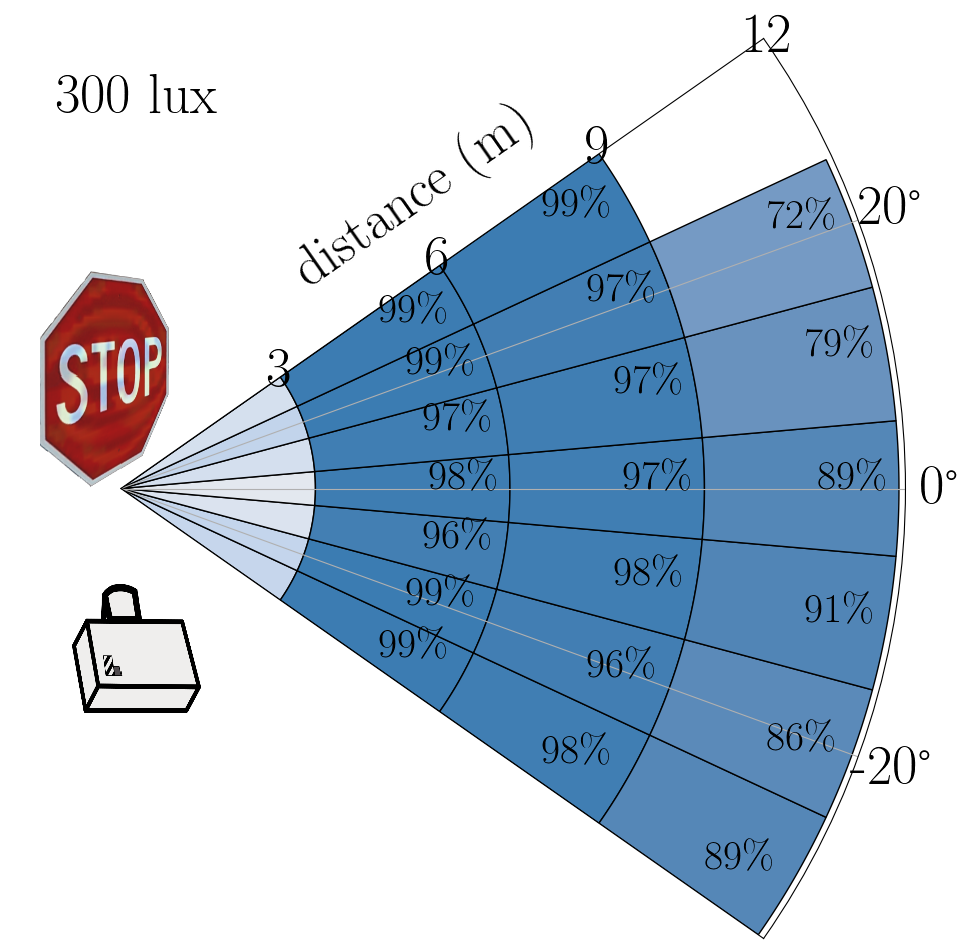}
		\caption%
		{{\small \yolo{}.}}
		\label{fig:300_cone_yolov3}
	\end{subfigure}
	\hfill
	\begin{subfigure}[t]{\mysize\textwidth}
		\centering
		\includegraphics[width=\textwidth]{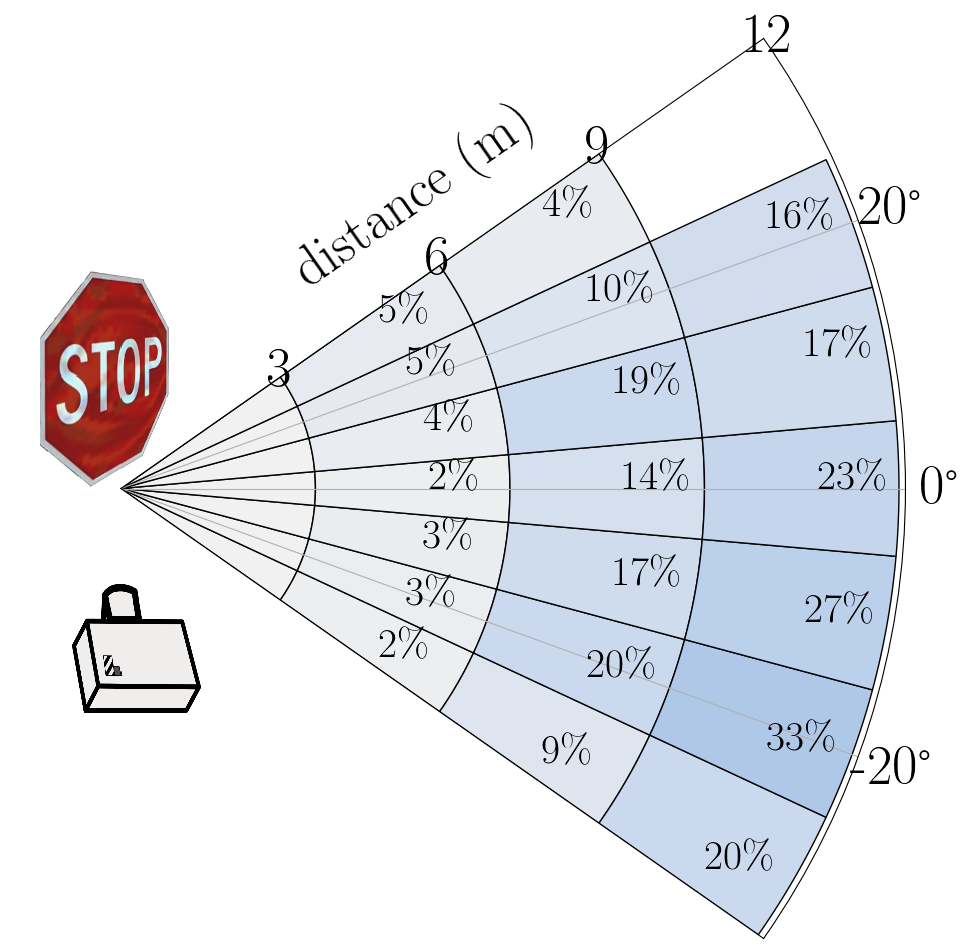}
		\caption[]%
		{{\small \mrcnn{}.}}
		\label{fig:300_cone_mrcnn}
	\end{subfigure}
	\hfill
	\begin{subfigure}[t]{\mysize\textwidth}
		\centering
		\includegraphics[width=\textwidth]{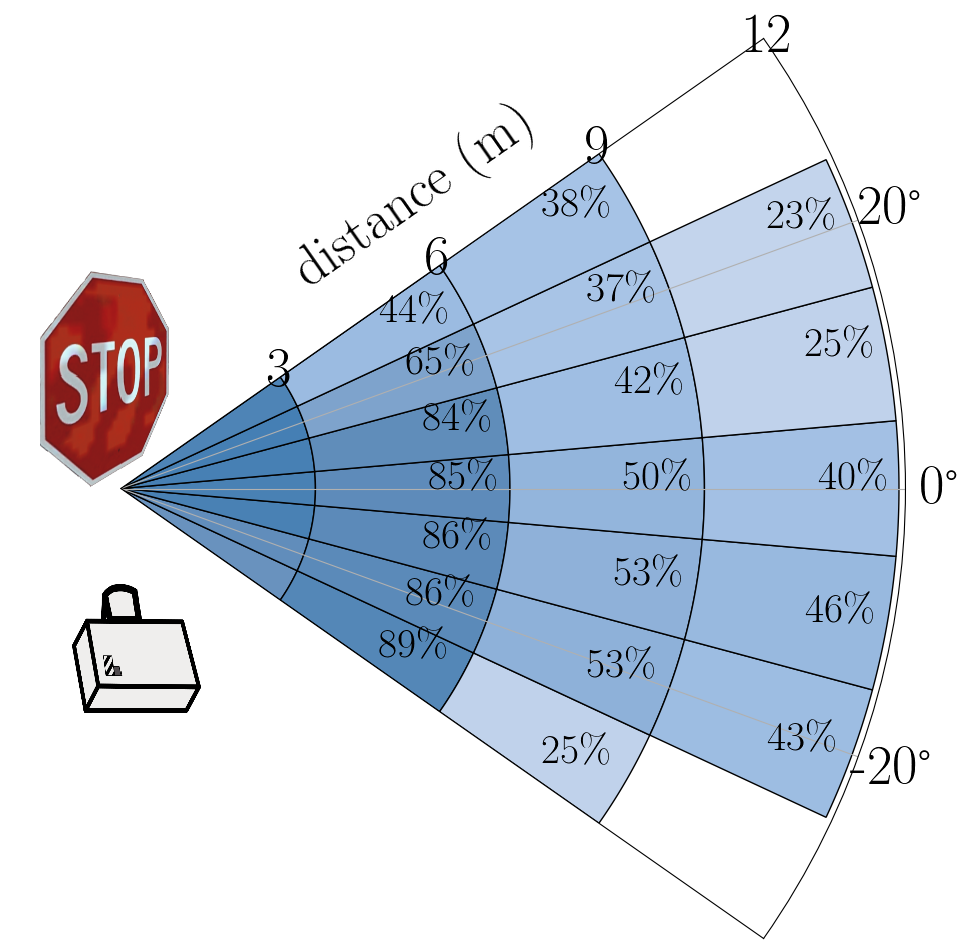}
		\caption[]%
		{{\small \gtsrb{}.}}
		\label{fig:300_cone_gtsrbcnn}
	\end{subfigure}
	\hfill
	\begin{subfigure}[t]{\mysize\textwidth}
		\centering
		\includegraphics[width=\textwidth]{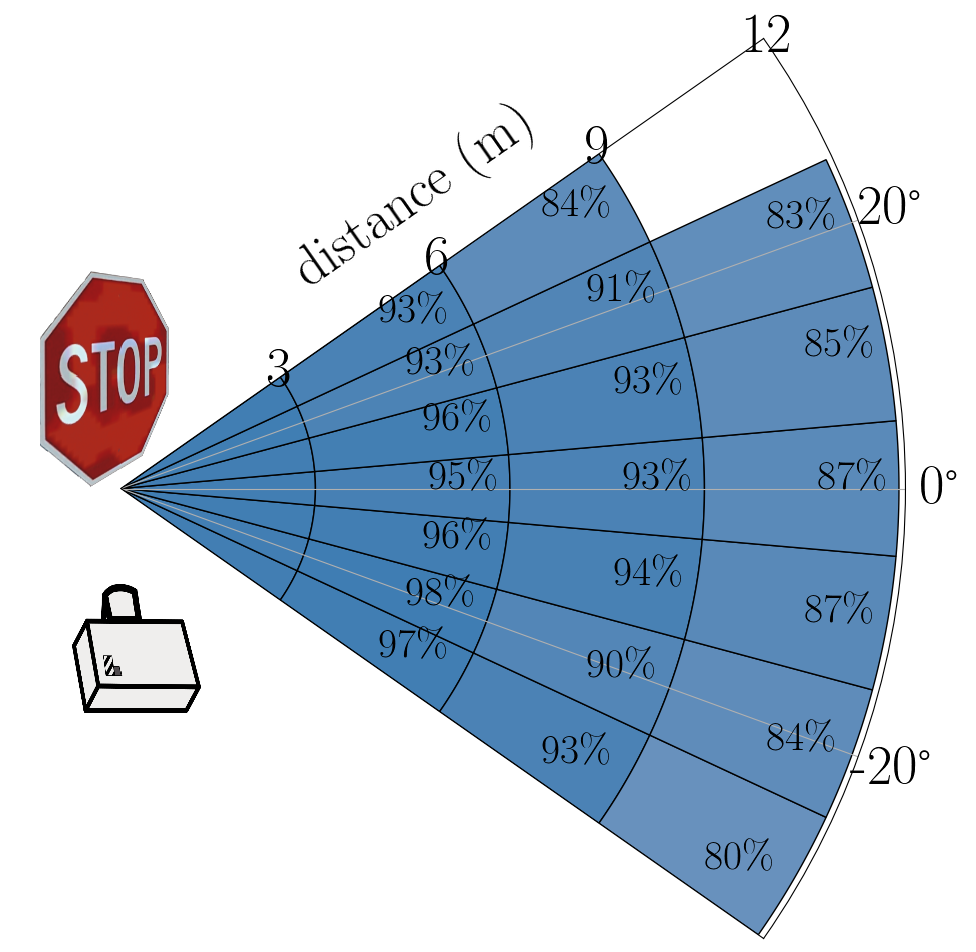}
		\caption[]%
		{{\small \lisa{}.}}
		\label{fig:300_cone_lisacnn}
	\end{subfigure}
	\hfill
	
	\begin{subfigure}[t]{\mysize\textwidth}
		\centering
		\includegraphics[width=\textwidth]{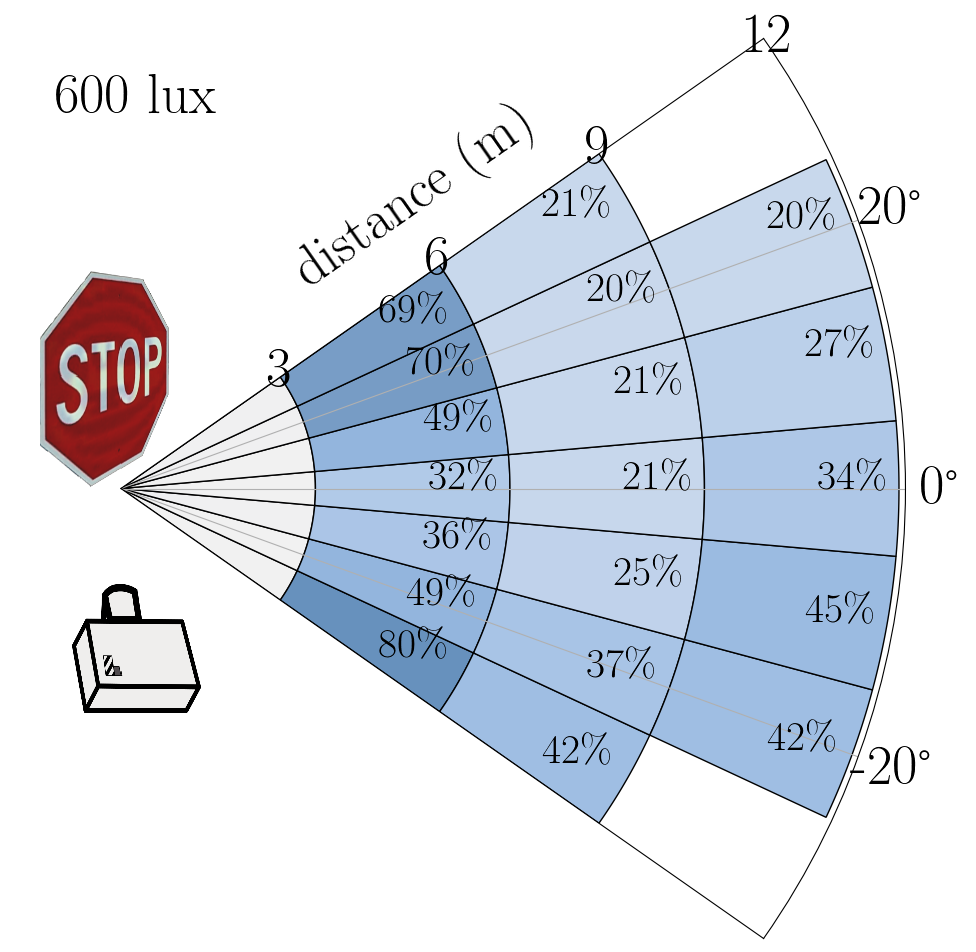}
		\caption%
		{{\small \yolo{}.}}
		\label{fig:600_cone_yolov3}
	\end{subfigure}
	\hfill
	\begin{subfigure}[t]{\mysize\textwidth}
		\centering
		\includegraphics[width=\textwidth]{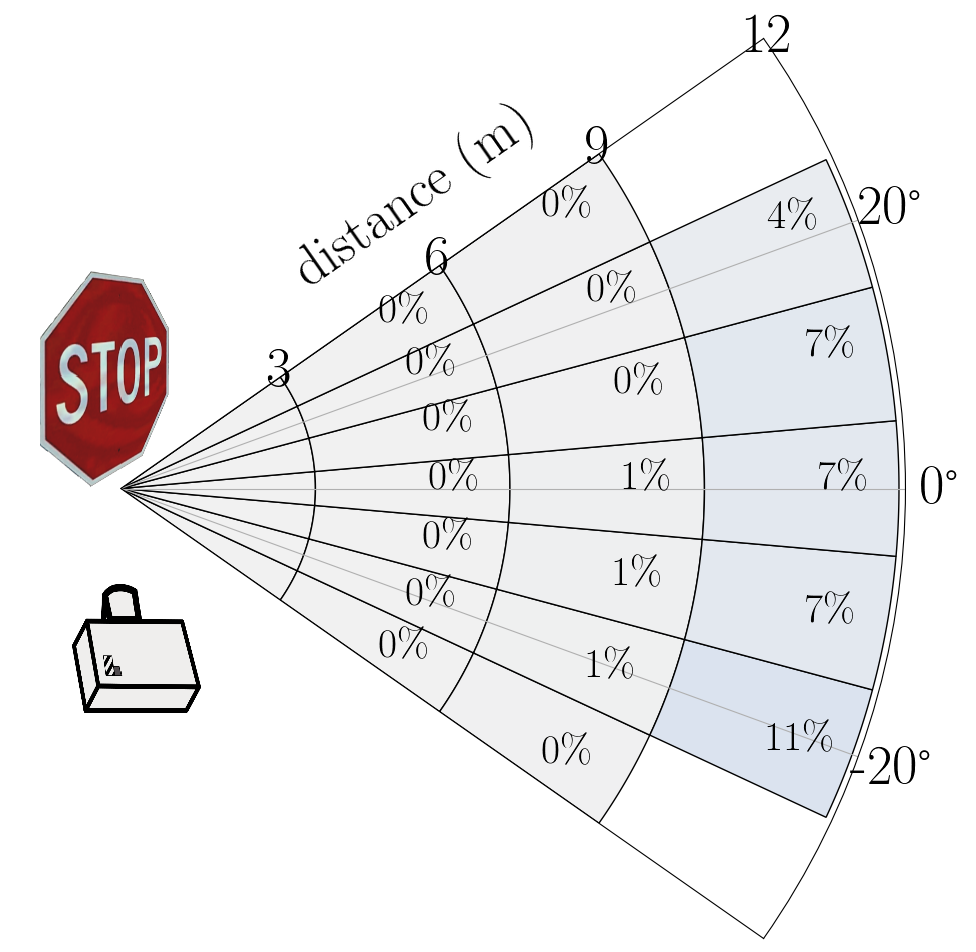}
		\caption[]%
		{{\small \mrcnn{}.}}
		\label{fig:600_cone_mrcnn}
	\end{subfigure}
	\hfill
	\begin{subfigure}[t]{\mysize\textwidth}
		\centering
		\includegraphics[width=\textwidth]{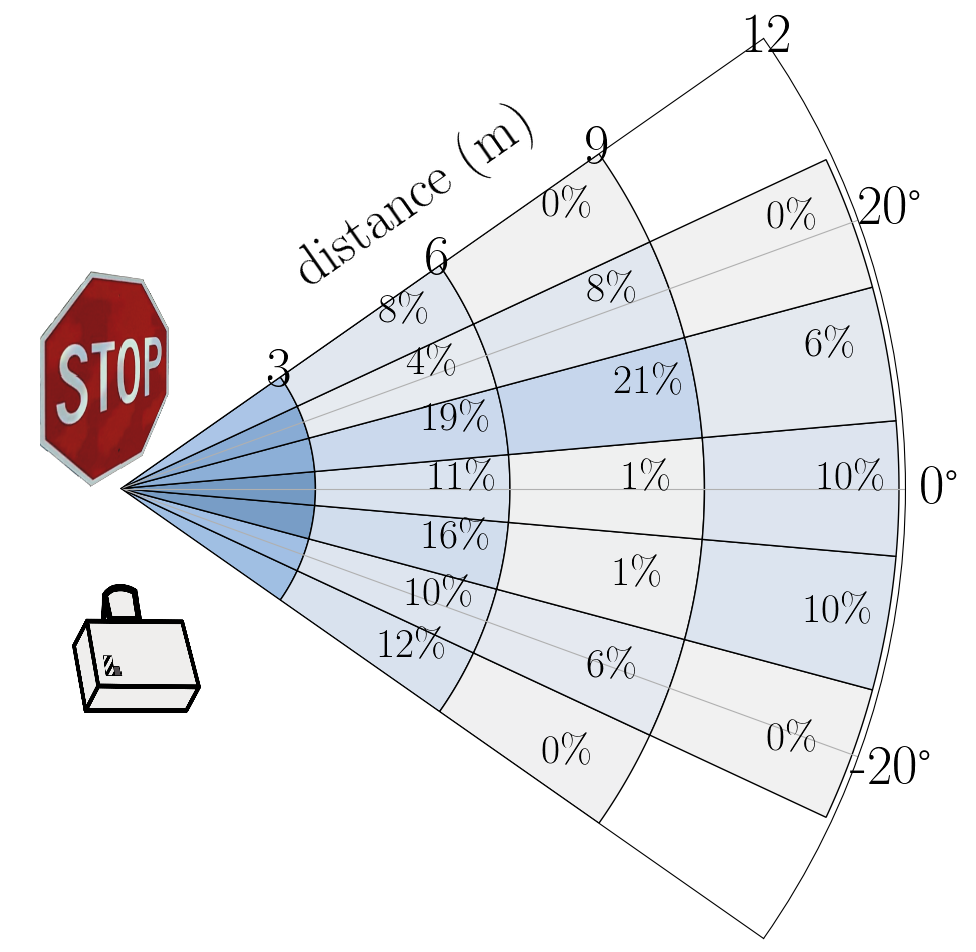}
		\caption[]%
		{{\small \gtsrb{}.}}
		\label{fig:600_cone_gtsrbcnn}
	\end{subfigure}
	\hfill
	\begin{subfigure}[t]{\mysize\textwidth}
		\centering
		\includegraphics[width=\textwidth]{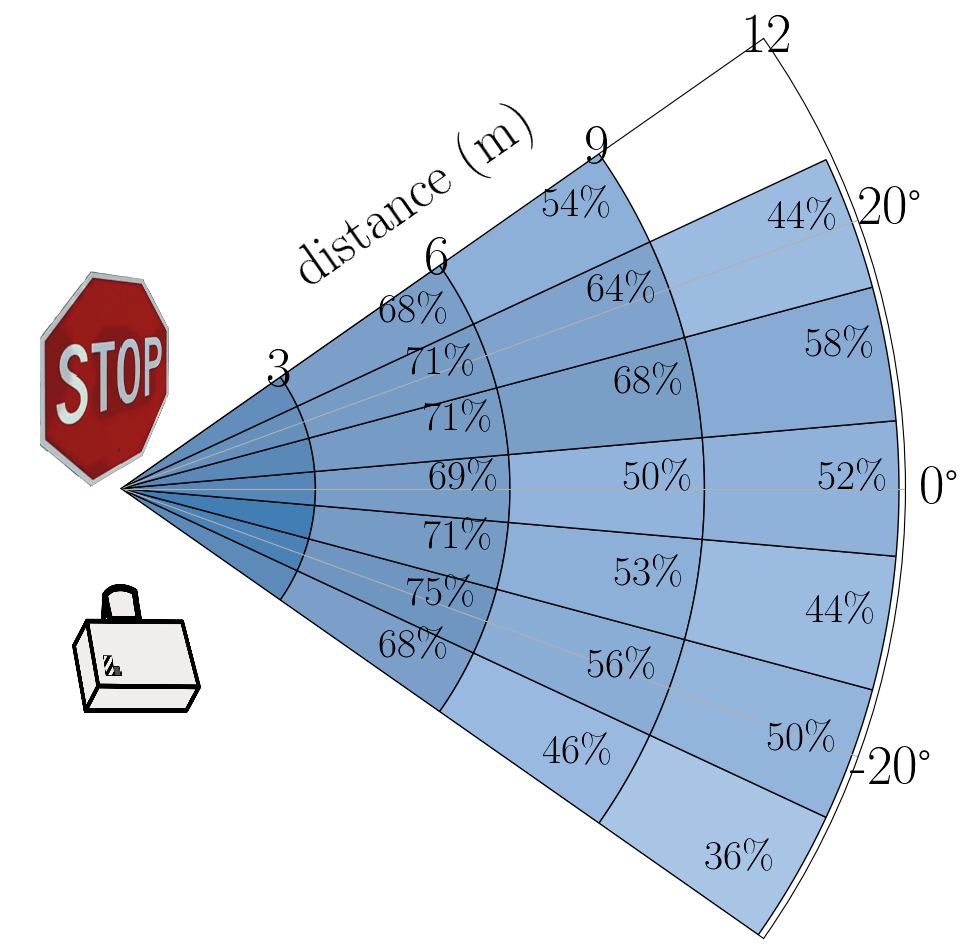}
		\caption[]%
		{{\small \lisa{}.}}
		\label{fig:600_cone_lisacnn}
	\end{subfigure}

	\caption{Attack success rate at different angles, distances and for different networks, as the percentage of frames where a stop sign is not detected. Darker shades represent higher success rates. Percentages for 0-3m are omitted for clarity, but the corresponding cone section is colored accordingly. The images of the stop signs in the figure are computed using the projection models for the two light settings, so they resemble what the adversarial stop sign looks like in practice. }
	\label{fig:cone}
\end{figure*}

In this section, we present the results of the detection for the controlled indoor experiment.
We also report in Figure~\ref{fig:baseline_cone} the baseline results of using the networks to detect/classify the stop sign, by recording videos of the stop sign unaltered.
Figure~\ref{fig:baseline_cone} shows that all networks work quite well in non-adversarial conditions, with the exception of \lisa{} which shows a few misdetections.

We report in Figure~\ref{fig:cone} the results of the detection for the 120,  300 and 600 lux setting for the different networks, as the percentage of frames where the stop sign was not detected by the network.
The minimum number of frames tested for a single model is 3,438, see Table~\ref{tab:transferability_app} for exact figures.
Figure~\ref{fig:cone} shows that the attack is extremely successful in dimmer lighting conditions, obtaining >99\% success rate for all networks except \mrcnn{}, which presents additional resilience at shorter distances.
The figure also shows how our method is able to create AE that generalize extremely well across all the measured distances from 1 to 12m and viewing angles -30$^\circ$ to 30$^\circ$.
As the ambient increases, the success rate quickly decreases accordingly.
Already at 300 lux, the attack success rate is greatly reduced for \mrcnn{} and \gtsrb{}, while \yolo{} and \lisa{} still remain vulnerable, but the attack degradation becomes evident at 600 lux.


Overall, we found that \mrcnn{} is consistently more resilient than the other networks in the detection.
In particular we found that \mrcnn{} sometimes recognizes stop signs just based on the octagonal silhouette of the sign or even just with faded reflections of the sign on windows.
This could be a combination of \mrcnn{} learning more robust features for the detection (possibly thanks to the higher model complexity) and of using a region proposal network for the detection~\cite{he2017mask}.
Nevertheless, such robustness comes at the cost of execution speed: \mrcnn{} requires up to 14 times the execution time of \yolo{} (300ms vs 22ms).



\begin{figure*}[t]
	\centering
	\includegraphics[width=1.0\textwidth]{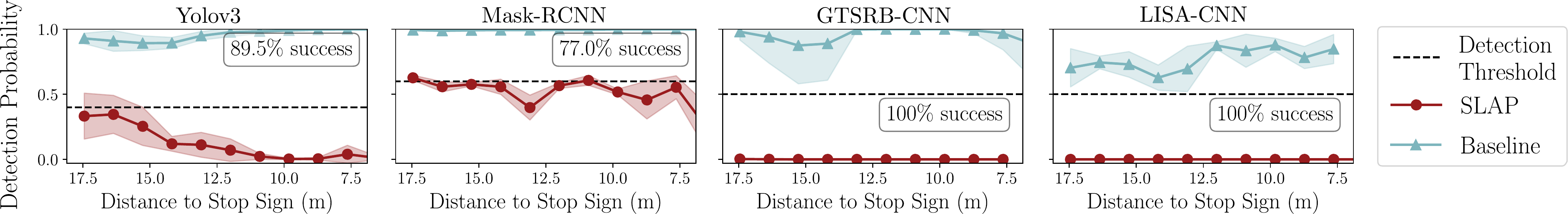}
	\caption{Detection probability for the stop sign during the road driving test. During the test the car approaches the stop sign while the attack is being carried out, the ambient light during the measurements is $\sim$120 lux, the car headlights are on. The data are grouped into 10 distance bins, the shaded areas indicate the standard deviation of the probability within that distance bin.}
	\label{fig:approach_road}
\end{figure*}

\subsection{Road Driving Test}\label{sec:road_driving_test}
To further test the feasibility of the attack, we carry out the attack outdoors in moving vehicle settings.

\myparagraph{Setup}
The experiment is carried out on a section of private road at our institution.
We mount the stop sign at 2m height and set the projector in front of it at a distance of approximately 2 metres.
The experiment was conducted shortly prior to sunset, at a longitude of 50 degrees north in early October\footnote{Exact longitude will be disclosed upon publication}.
At the time of the experiment the ambient light level measured at the surface of the sign is $\sim$ 120 lux.
We use a car to approach the stop sign at 10-15km/h, with the car headlights on during the approach.
Videos are recorded using the same iPhone X mounted inside the car at 240fps.
We follow the same pipeline described in Section~\ref{sec:experimental_setup}.
However, rather than carrying out the profiling step (Step 2 of the Experimental Procedure), we \textit{re-use} the 120 lux projections that were optimized for the controlled indoor conditions.

\myparagraph{Results}
We report the results from the driving test in Figure~\ref{fig:approach_road}, which shows the probability of detection for stop sign as the car approaches the sign.
The experiment measures up to 18m away to roughly 7m, when the stop sign exits the video frame (we keep the camera angle fixed during the approach).
The results closely match the findings indoor, with the attack being successful for most networks along the whole approach: we obtain 100\% success rate for \lisa{} and \gtsrb{} and over 77\% for \mrcnn{} and \yolo{}.
These results also confirm the generalizability of optimized projections: simply re-using projections without having to re-execute Step~2 and Step~3 of the experimental procedure at the time of attack led to similar success rates.
This means that adversaries could easily pre-compute a set of projections and quickly swap between them depending on the current light conditions.

\subsection{Defences}\label{sec:defences}

\begin{table*}[t]
	\small
	\centering
	\begin{tabular}{cc|ccccc}
	    & \textit{Ambient} & \textit{Attack} & \textit{Adversarial} & \textit{Input} & \multicolumn{2}{c}{\textit{SentiNet~\cite{Chou2018}}} \\
		\textit{Network} & \textit{Light} (lx) & \textit{Success} &  \textit{Learning~\cite{Szegedy2015}} & \textit{Randomization~\cite{xie2017mitigating}} & Random & Checkerboard \\ \toprule 
 \multirow{5}{*}{\textbf{\gtsrb{}}}  & 120 & 99.96\% & 20.23\% (-79.73\%) & 99.55\% (-0.40\%) & 93.43\% & 95.45\% \\ 
 & 180 & 90.53\% & 23.57\% (-66.97\%) & 90.02\% (-0.51\%) & 93.19\% & 93.72\% \\ 
 & 300 & 56.51\% & 48.18\% (-8.33\%) & 86.78\% (+30.27\%) & 96.97\% & 96.46\% \\ 
 & 440 & 56.34\% & 40.24\% (-16.10\%) & 82.96\% (+26.61\%) & 95.81\% & 96.34\% \\ 
 & 600 & 12.79\% & 10.91\% (-1.88\%) & 51.37\% (+38.58\%) & 95.29\% & 95.29\% \\ 
\midrule 
\multirow{5}{*}{\textbf{\lisa{}}}  & 120 & 100.00\% & 0.06\% (-99.94\%) & 100.00\% (+0.00\%) & 94.24\% & 95.29\% \\ 
 & 180 & 99.95\% & 0.88\% (-99.07\%) & 99.90\% (-0.05\%) & 100.00\% & 100.00\% \\ 
 & 300 & 99.81\% & 0.00\% (-99.81\%) & 99.98\% (+0.17\%) & 94.76\% & 96.86\% \\ 
 & 440 & 98.44\% & 0.59\% (-97.85\%) & 99.95\% (+1.51\%) & 100.00\% & 100.00\% \\ 
 & 600 & 69.05\% & 0.04\% (-69.01\%) & 95.71\% (+26.67\%) & 95.81\% & 96.86\% \\ 
\bottomrule 
\end{tabular}

	\caption{Attack success rate across the various evaluated defences, models and lux settings. Figures are reported as the percentage of frames in which the attack is successful, i.e., a stop sign is not detected. Differently, (*) figures for SentiNet are reported as percentage out of the 100 adversarial frames extracted from the videos, both overlaying patterns Random and Checkerboard are reported.}
	\label{tab:defenceresults}
\end{table*}

Generally, AE defences are aimed at detecting AE in a digital scenario, where adversaries have the capability to arbitrarily manipulate inputs, but are limited to an $L_p$-norm constraint.
In the case of physical AE, adversaries are not directly limited by an $L_p$-norm constraint but by the physical realizability of their AE.
Defences that are tailored to physical AE have not been investigated as much as general-scope AE defences.
For this reason, we specifically choose to evaluate our AE against Sentinet~\cite{Chou2018}: it is one of the few published works that addresses physical AE detection.
Additionally, we evaluate our attack against two other defences which could be used in the autonomous driving scenario as they not entail additional running time: the input randomization by Xie et al.~\cite{xie2017mitigating} and adversarial learning~\cite{Szegedy2015}.
In the following we describe our evaluation setup and results.

\begin{figure}[t]
	\centering
	\includegraphics[width=0.48\textwidth]{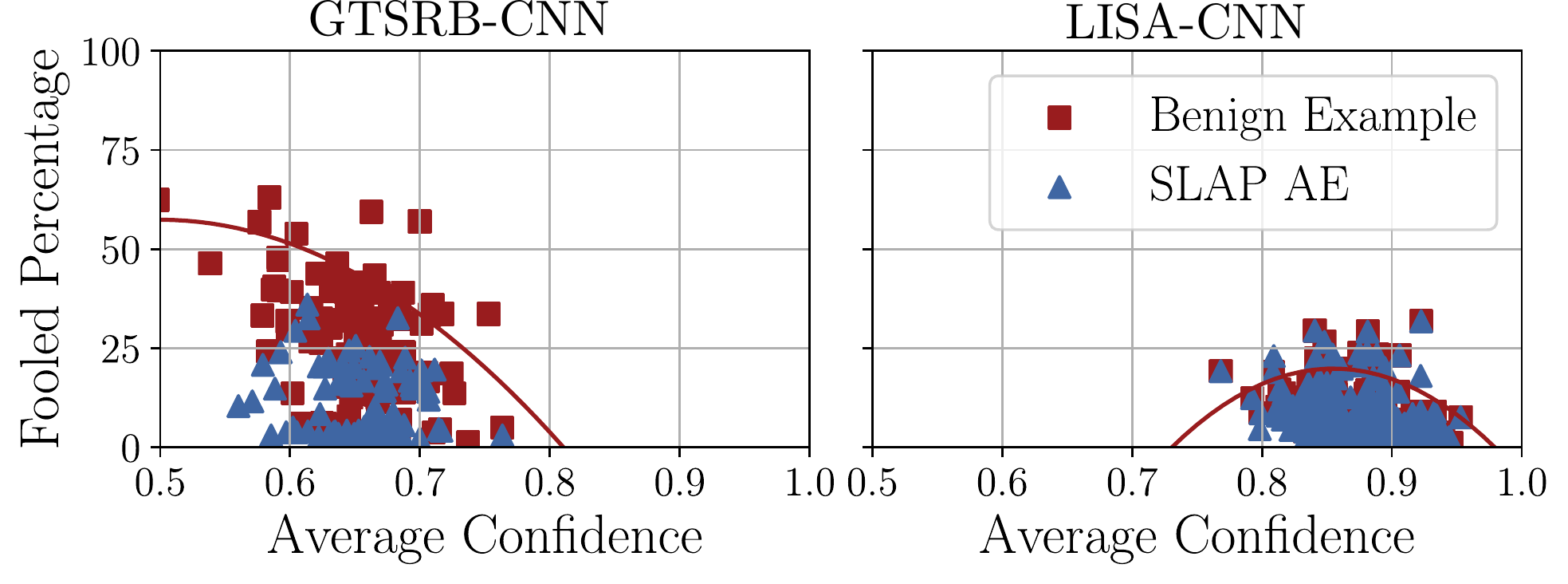}
	\caption{Visualization of the SentiNet detection results (from the 180 lux setting). The plot shows that the SLAP AE have a behaviour similar to benign examples across the two dimensions used by SentiNet, preventing detection.}
	\label{fig:sentinet_180_noise_sentinet}
\end{figure}

\myparagraph{Setup and Remarks}
We evaluate the three considered AE defences applied to  \gtsrb{} and \lisa{}, as all three defences are designed to work in image classification scenario; at the time of writing, defences for object detectors are not as well explored.
For SentiNet, we use 100 \textit{benign} images taken from the GTSRB and LISA dataset to compute the threshold function, 100 \textit{test} images where we overlay the suspected adversarial regions and 100 random frames containing a SLAP AE from the collected videos. 
For our Sentinet implementation, we use XRAI~\cite{kapishnikov2019xrai} to compute saliency masks as the original method used (GradCam~\cite{chattopadhay2018grad}) led to too coarse grained masks (see Appendix~\ref{sec:sentinet_description}).
For the input randomization of~\cite{xie2017mitigating}, we set the maximum size of the padded image to be 36 (from 32).
For adversarial learning~\cite{Szegedy2015}, we re-write the \lisa{} and \gtsrb{} models and we train them on the respective datasets from scratch adding an FGSM-adversarial loss to the optimization. 
We use Adam with learning rate 0.001, the weight of the adversarial loss is set to 0.2, the FGSM step size to 0.2, we use $L_{\text{inf}}$-norm and train for 50 epochs.
Adversarially trained models present a slight accuracy degradation on the test set compared to training them with categorial cross-entropy, \gtsrb{} goes from 98.47\% to 98.08\% (-.39\%) while \lisa{} from 95.9\% to 95.55\% (-.35\%).
For input randomization and adversarial learning we run the inference on all the collected video frames of the experiment.


\myparagraph{Results} 
We report the results in Table~\ref{tab:defenceresults}.
The table shows the attack success rate computed as the percentages of frames where a stop sign was not detected.
We also report the legitimate attack success for comparison.
We found that input randomization does not detect our attack, this is expected given that any type of input augmentation-defence is intrinsically compensated for by our optimization (see Section~\ref{sec:training_data_augmentation}).
Even worse, such method actually degrades the accuracy of the model, showing that the original models for \lisa{} and \gtsrb{} taken from~\cite{Eykholt2017} were not trained with sufficient data augmentation.
As expected, thanks to the larger affected areas of the SLAP AE, these adversarial samples can bypass detection by SentiNet in over 95\% of the evaluated frames, with no significant difference across the overlay pattern used (either Random or Checkerboard).
We also report a visualization of the threshold function fit in SentiNet in Figure~\ref{fig:sentinet_180_noise_sentinet}, showing that the behaviour of SLAP AE resembles those of normal examples.
We found that adversarial learning is a more suitable way to defend against SLAP, stopping a good portion of the attacks.
Nevertheless, the fact that we only evaluate an adaptive defender (not an adaptive adversary) and that adversarially-trained models suffer from benign accuracy degradation (performance of the model with no attack in place) highlights how SLAP still remains a potential threat.


\subsection{Attack Transferability}\label{sec:attack_transferability}

\begin{table*}[t]
	\centering\footnotesize
	\begin{tabular}{cc@{\hskip 6pt}c|c@{\hskip 6pt}c@{\hskip 6pt}c@{\hskip 6pt}c@{\hskip 6pt}c@{\hskip 6pt}c@{\hskip 6pt}c}
		& & & \multicolumn{7}{c}{\textit{Target Model}} \\
		\textit{lux} & \textit{Source Model} & \textit{no. frames} & \yolo{} & \mrcnn{} & \gtsrb{} & $\text{\gtsrb{}}^{(a)}$ & \lisa{} & $\text{\lisa{}}^{(a)}$ & Google Vision* \\ \toprule 
		\multirow{4}{*}{\textbf{120}}& \yolo{}& 4587 & \textbf{100.0\%}& 73.4\%& 0.0\%& 0.0\%& 21.5\%& 0.0\%& 100.0\%\\ 
		& \mrcnn{}& 3765 & 98.7\%& \textbf{97.1\%}& 0.0\%& 0.0\%& 15.5\%& 0.0\%& 100.0\%\\ 
		& \gtsrb{}& 3760 & 40.5\%& 37.0\%& \textbf{99.9\%}& 16.1\%& 51.4\%& 0.0\%& 72.4\%\\ 
		& \lisa{}& 4998 & 29.4\%& 28.1\%& 6.8\%& 0.0\%& \textbf{100.0\%}& 0.0\%& 77.1\%\\ 
		\midrule\multirow{4}{*}{\textbf{300}}& \yolo{}& 5169 & \textbf{96.5\%}& 3.6\%& 2.5\%& 0.0\%& 2.3\%& 0.0\%& 72.3\%\\ 
		& \mrcnn{}& 3543 & 32.0\%& \textbf{14.0\%}& 0.1\%& 0.0\%& 10.4\%& 0.0\%& 65.9\%\\ 
		& \gtsrb{}& 3438 & 2.0\%& 2.9\%& \textbf{48.0\%}& 43.1\%& 44.0\%& 0.0\%& 47.6\%\\ 
		& \lisa{}& 4388 & 0.7\%& 4.9\%& 8.6\%& 0.0\%& \textbf{100.0\%}& 0.0\%& 25.0\%\\ 
		\midrule\multirow{4}{*}{\textbf{600}}& \yolo{}& 5507 & \textbf{17.8\%}& 0.2\%& 32.5\%& 0.0\%& 27.4\%& 0.0\%& 23.7\%\\ 
		& \mrcnn{}& 5058 & 0.1\%& \textbf{0.4\%}& 5.3\%& 0.0\%& 4.6\%& 0.0\%& 16.7\%\\ 
		& \gtsrb{}& 4637 & 0.0\%& 0.9\%& \textbf{7.2\%}& 7.5\%& 4.9\%& 0.0\%& 21.1\%\\ 
		& \lisa{}& 4714 & 0.0\%& 0.9\%& 8.6\%& 0.0\%& \textbf{57.5\%}& 0.0\%& 15.8\%\\ 
		\bottomrule 
	\end{tabular}
	\caption{
		Transferability results. We test all the frames from the collected videos with a certain projection being shone against a
		different target model, figures in bold are white-box pairs. (*) For Google Vision we only test one frame every 30 frames,
		i.e., one per second. We also remove all frames that are further than 6m away as Google Vision does not detect most of  
		them in a baseline scenario. Models indicated with $\_^{(a)}$ indicate adversarially trained models.  
	}	
	\label{tab:transferability}
\end{table*}

\myparagraph{Setup}
In this section, we test the transferability of our attack across networks, testing all pairwise combinations of our models, including adversarially trained ones.
We also use the Google Vision API\cite{googlevisionapi} to test our projections against their proprietary models.
The API returns a list of labeled objects in the image with associated confidence scores and bounding boxes, "stop sign" is one of the labels.
We set the detection threshold for Google Vision API as 0.01, i.e., we count that a stop sign is detected in a frame if the API replies with a stop sign object with confidence greater than 0.01.

\myparagraph{Results}
We report the results in Table~\ref{tab:transferability}.
The table shows the source (white-box) model on the left, which identifies the projection shown in the tested videos.
We also report the number of frames tested, taken from the videos from the indoor experiment.
Table~\ref{tab:transferability} reprorts success rates of the attack as a percentage of the frames where the stop sign was undetected.
Table~\ref{tab:transferability} shows that our attack transfers well for low light settings, but the transferability degrades quickly for the 300 lux setting and above.
We find that \mrcnn{} transfers better to \yolo{} compared to the opposite direction, the same happens for \gtsrb{} and \lisa{}, suggesting that fitting AE on complex models favours the attacker.
Table~\ref{tab:transferability} also shows that adversarially trained model have benefits by reducing the transferability of attacks fit on surrogate models.

\section{Discussion}\label{sec:discussion}
In this section, we discuss the attack feasibility.

\myparagraph{Attack Feasibility}
\begin{figure}[t]
	\centering
	\includegraphics[width=0.46\textwidth]{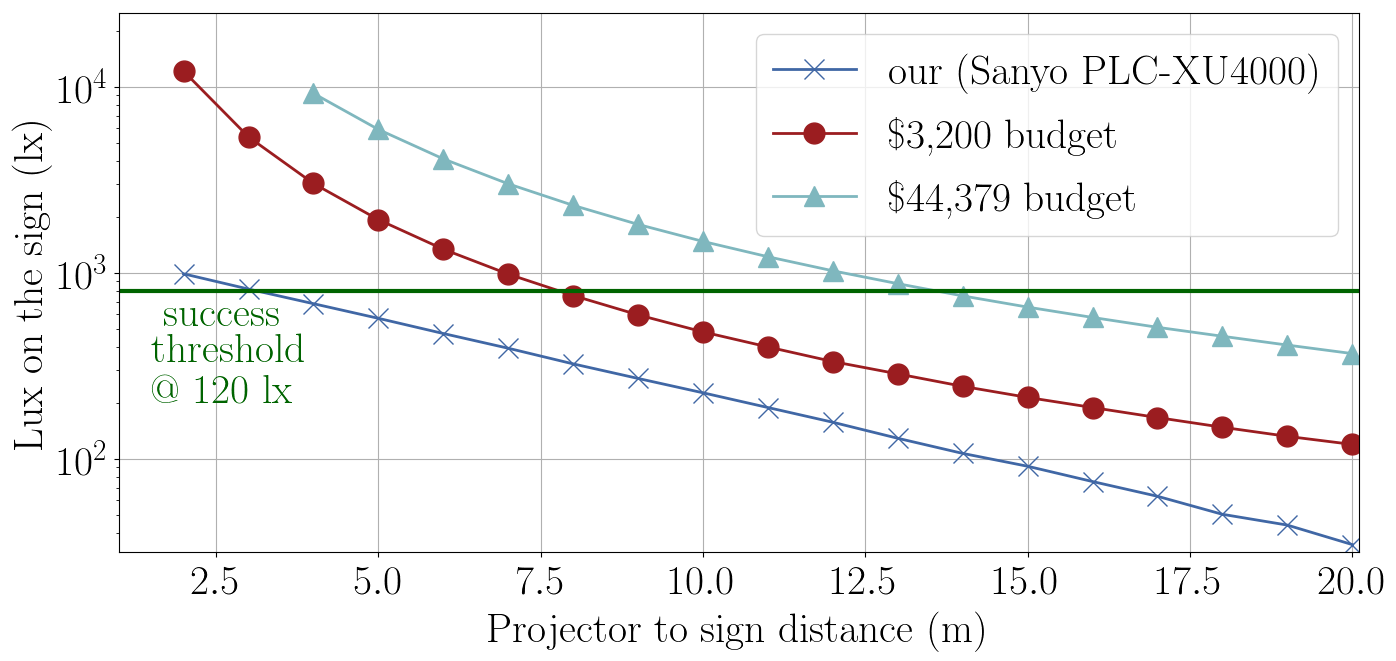}
	\caption{Amount of lux achievable on the stop sign surface for increasing projection distances and different projectors. The horizontal line shows the threshold for success measured in our experiments (800 lux at 120 lux ambient light).}
	\label{fig:attack_feasibility}
\end{figure}
Our experiments demonstrate that increasing ambient light quickly stops the feasibility of the attack in bright conditions.
In practice, during daytime, the attack could be conducted on non-bright days, e.g., dark overcast days or close to sunset or sunrise, when the ambient light is low (<400 lux).
Regarding the effect of car headlights, our outdoor experiments show that the car headlights-emitted light is negligible compared to the projection luminosity and does not influence the attack success.
While car headlights on high-beam would compromise the projection appearance and degrade the attack success rates, we did not consider these lights to be on as stop signs would be mainly present in urban areas, where high-beam headlights would be off.
In general, the amount of projector-emitted light that reaches the sign depends on three factors: (i) the distance between projector and sign, (ii) the throw ratio of the projector  and (iii) the amount of lumens the projector can emit.
We report in Figure~\ref{fig:attack_feasibility} a representation of how the distance between the projector and the stop sign relates to the attack success rate.
We consider two additional projectors with long throw distance, the Panasonic PT-RZ570BU and the  NEC PH1202HL1, available for \$3,200 and \$44,379 respectively.
We use the projector's throw ratios (2.93 and 3.02) and their emitted lumens (5,000 and 12,000 lumens) to calculate how many lux of light the projector can shine on the sign surface from increasing distances.
We consider the success as measured in 120 lux ambient light, where obtaining 800 lux of light on the sign with the projector is sufficient to achieve consistent attack success (see Section~\ref{sec:experimental_setup}).
Figure~\ref{fig:attack_feasibility} shows that the attack could be carried out from 7.5m away with the weaker projector and up to 13m away with the more expensive one.
Additionally, adversaries could also use different lenses to increase the throw ratio of cheaper projectors (similarly to ~\cite{man2020ghostimage}).


\myparagraph{Attack Generalizability}
We show results for attacks on other objects (give way sign, bottle) in Appendix~\ref{sec:app:other_objects}, however, to extend the attack to \textit{any} object, the adversary will have to consider the distortion introduced by the projection surface (not necessary for flat traffic signs).
The attacker will have to augment the projection model used in this paper with differentiable transformations which model the distortion caused by the non-flat surface.
In general, the size of the projectable area limits the feasibility of the attack against certain objects (e.g., hard to project on a bike); this drawback is shared across all vectors that create physically robust AE, including adversarial patches.
We also found that the properties of the material where the projection is being shone will impact the attack success: traffic signs are an easier target because of their high material reflectivity.
When executing the attack on other objects, we found that certain adaptations lead to marginal attack improvements, in particular context information (e.g., the pole for the stop sign, the table where the bottle is placed).
Generally, for object detectors, adversaries will have to tailor certain parameters of the optimization to the target object.

\section{Conclusions}
In this paper we presented SLAP, a new attack vector to realize short-lived physical adversarial examples by using a light projector.
We investigate the autonomous driving scenario, where the attacker's goal is to change the appearance of a stop sign by shining a crafted projection onto it so that it is undetected by the DNNs mounted on passing vehicles.

Given the non-trivial physical constraints of projecting specific light patterns on various materials in various conditions, we proposed a method to generate projections based on fitting a predictive three-way color model and using an AE generation pipeline that enhances the AE robustness.
We evaluated the proposed attack in a variety of light conditions, including outdoors, and against state-of-the-art object detectors \yolo{} and \mrcnn{} and traffic sign recognizers \lisa{} and \gtsrb{}.
Our results show that SLAP generates AEs that are robust in the real-world.
We evaluated defences, highlighting how existing defences tailored to physical AE will not work against AE generated by SLAP, while finding that an adaptive defender using adversarial learning can successfully hamper the attack effect, at the cost of reduced accuracy.

Nevertheless, the novel capability of modifying how an object is detected by DNN models, combined with the capability of carrying out opportunistic attacks, makes SLAP a powerful new attack vector that requires further investigation.
This paper makes an important  step towards increasing the awareness and further research of countermeasures against light-projection adversarial examples.



\section*{Source Code Availability}
The experiments code is available online.\footnote{\href{https://github.com/ssloxford/short-lived-adversarial-perturbations}{\textit{https://github.com/ssloxford/short-lived-adversarial-perturbations}}{}}
\Urlmuskip=0mu plus 1mu\relax
\bibliographystyle{plain}
\small
\bibliography{references}
\balance

\normalsize

\appendix
\clearpage

\section{Attack on Different Objects}\label{sec:app:other_objects}

Our evaluation focuses on the detection of stop signs, as this is generally an important component of road safety. 
Nevertheless the introduced attack can generalize in principle to any kind of deep neural network which uses RGB-camera inputs to make decisions.
To show how our attack generalizes, we also investigate the feasibility of the attack on different objects.

\myparagraph{Setup} For \lisa{} and \gtsrb{}, we choose another traffic sign, ``give way'', while for \yolo{} and \mrcnn{} we choose the ``bottle'' class.
For the give way sign and the bottle, we run a reduced evaluation: we execute all the experiment procedure steps reported in Section~\ref{sec:experimental_setup} and we test the correct (mis-)classification across a set of photos of the altered objects.
Extending our method to other objects is straightforward, it only requires to change the input mask of the projection.
When projecting on non-flat surfaces, the adversary will also have to consider the distortion introduced by those surfaces, this is briefly discussed in Section~\ref{sec:discussion}.

\begin{figure*}[t]
	\centering
	\begin{subfigure}[t]{.34\textwidth}
		\centering
		\includegraphics[width=\textwidth]{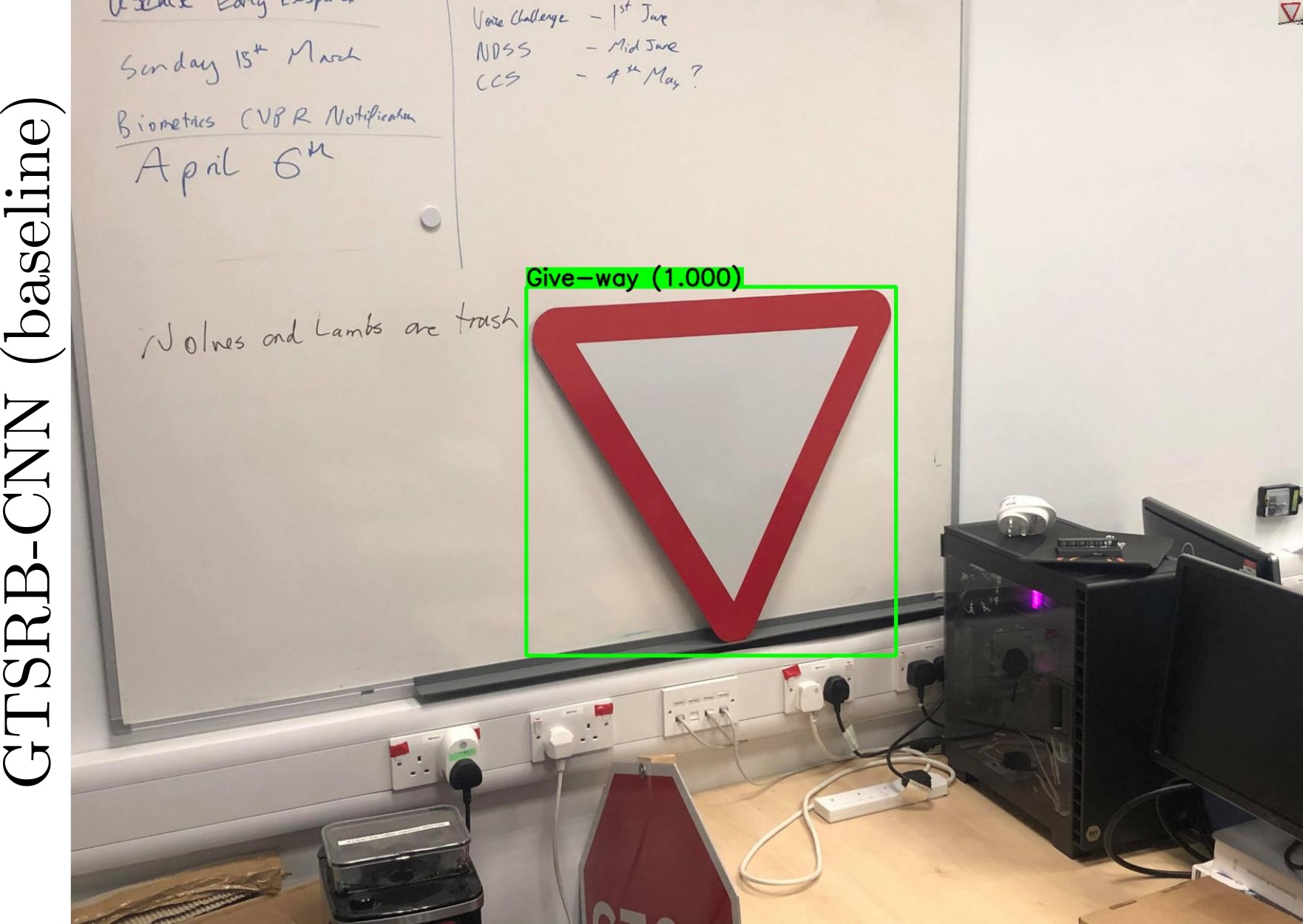}
	\end{subfigure}
	\hfill
	\begin{subfigure}[t]{.32\textwidth}
		\centering
		\includegraphics[width=\textwidth]{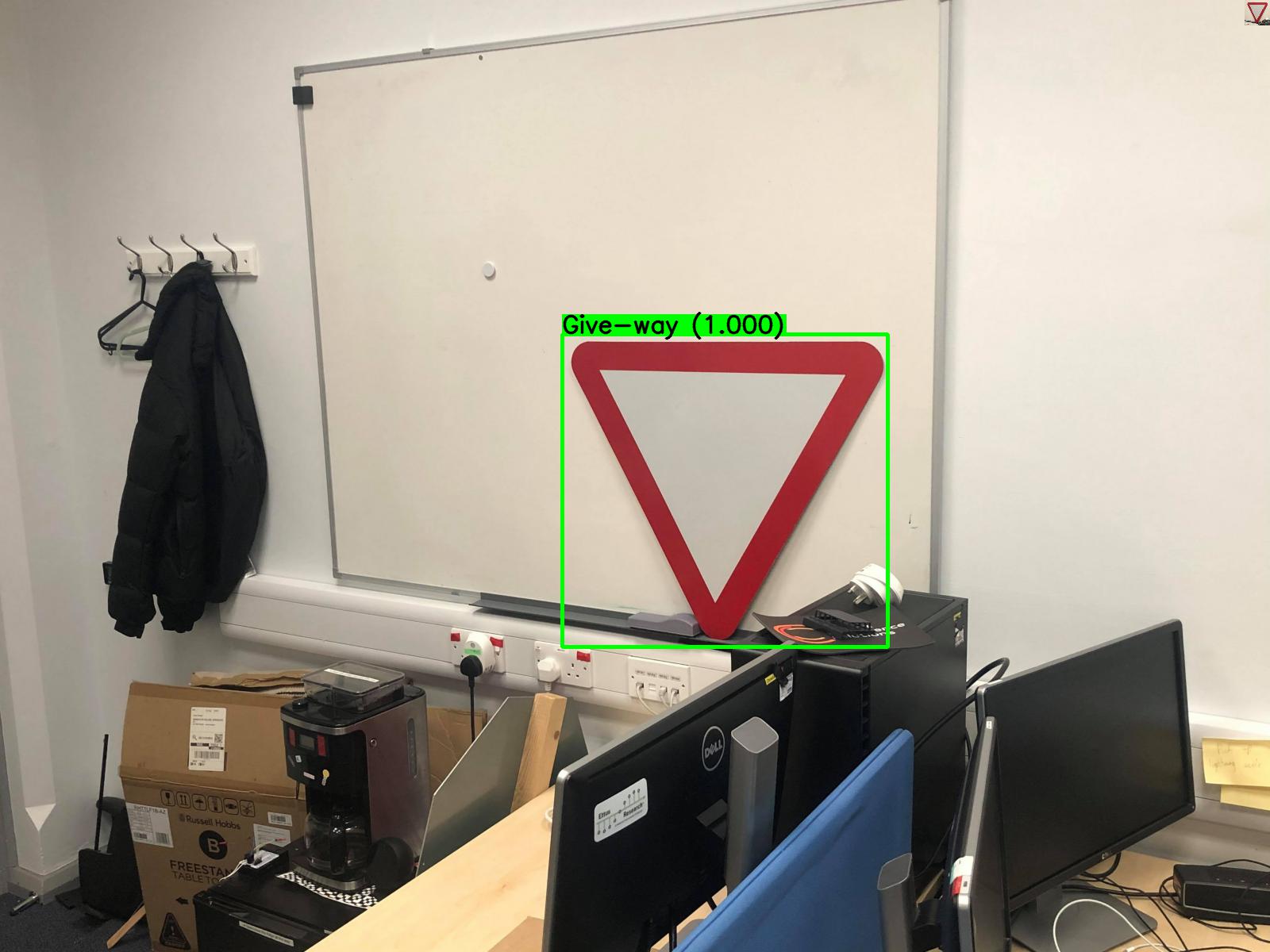}
	\end{subfigure}
	\hfill
	\begin{subfigure}[t]{.32\textwidth}
		\centering
		\includegraphics[width=\textwidth]{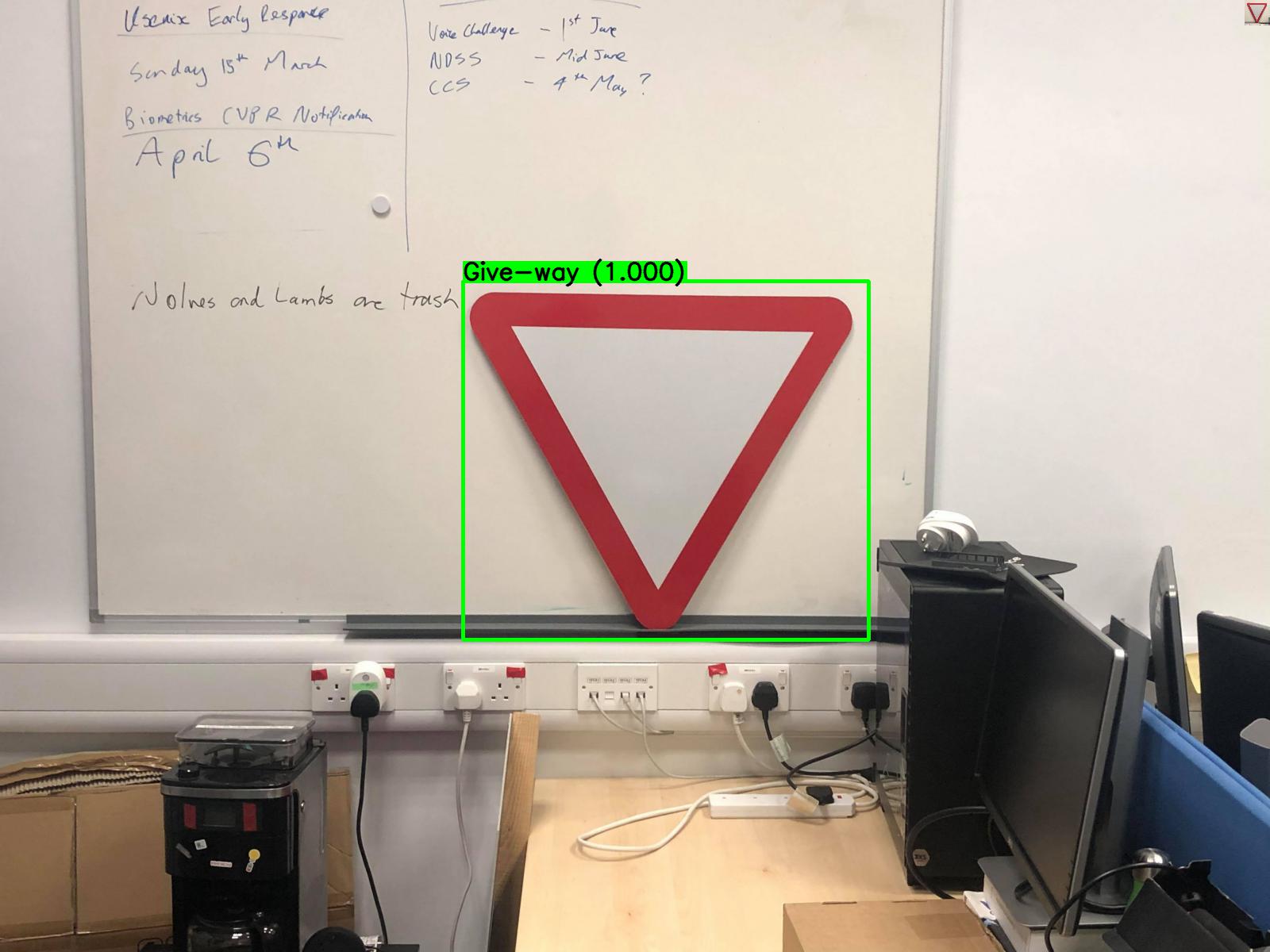}
	\end{subfigure}
	
	\begin{subfigure}[t]{.34\textwidth}
		\centering
		\includegraphics[width=\textwidth]{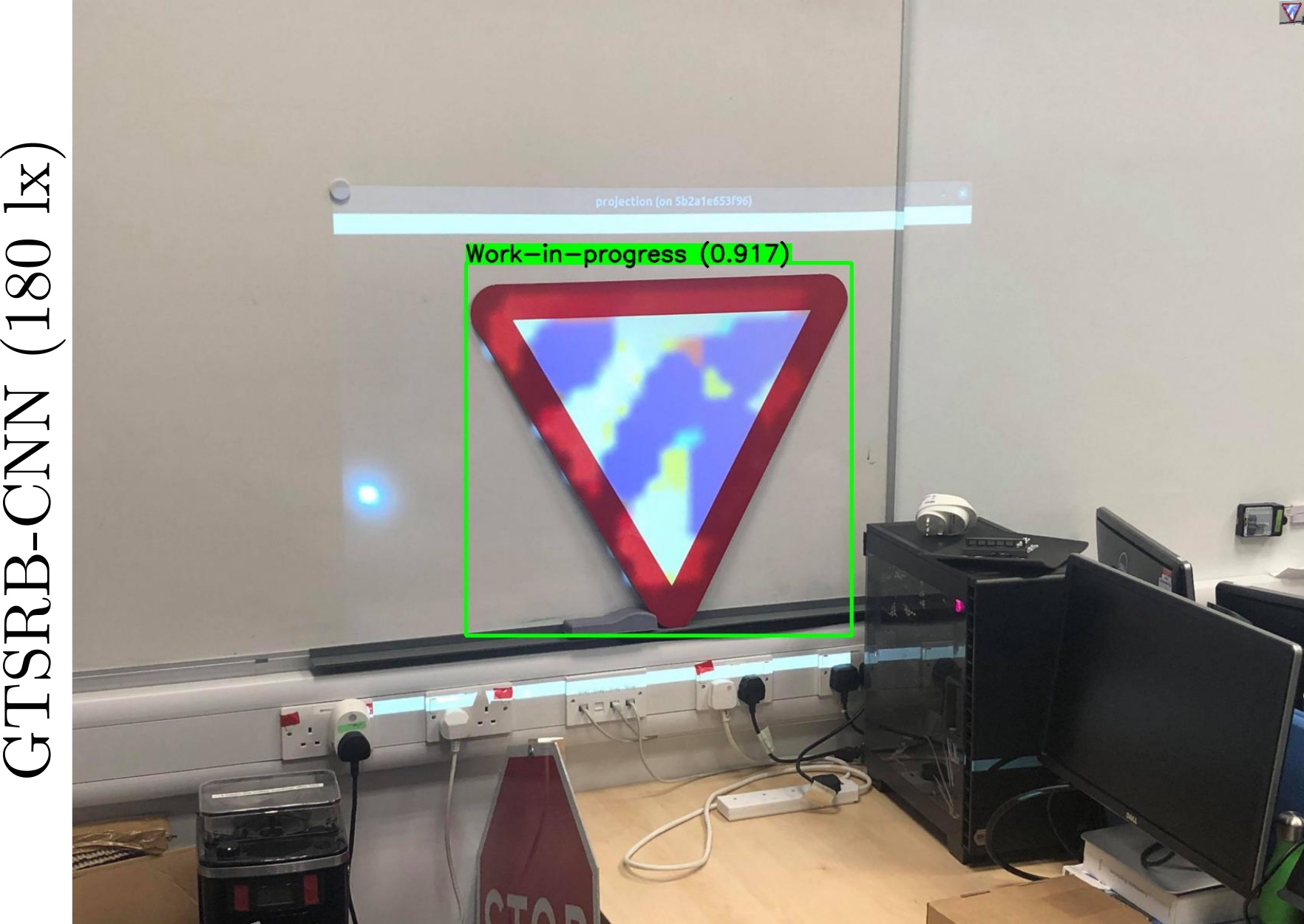}
	\end{subfigure}
	\hfill
	\begin{subfigure}[t]{.32\textwidth}
		\centering
		\includegraphics[width=\textwidth]{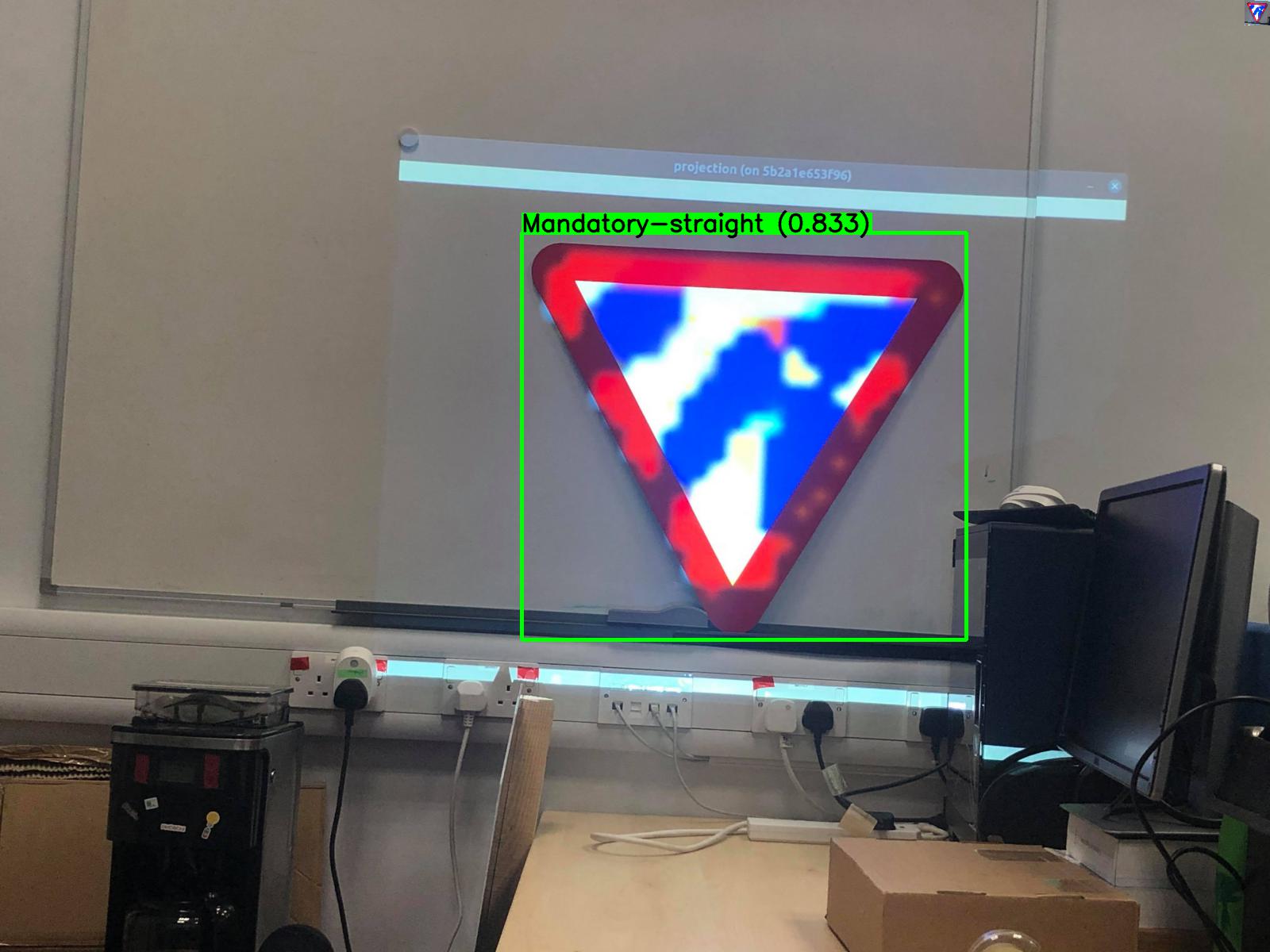}
	\end{subfigure}
	\hfill
	\begin{subfigure}[t]{.32\textwidth}
		\centering
		\includegraphics[width=\textwidth]{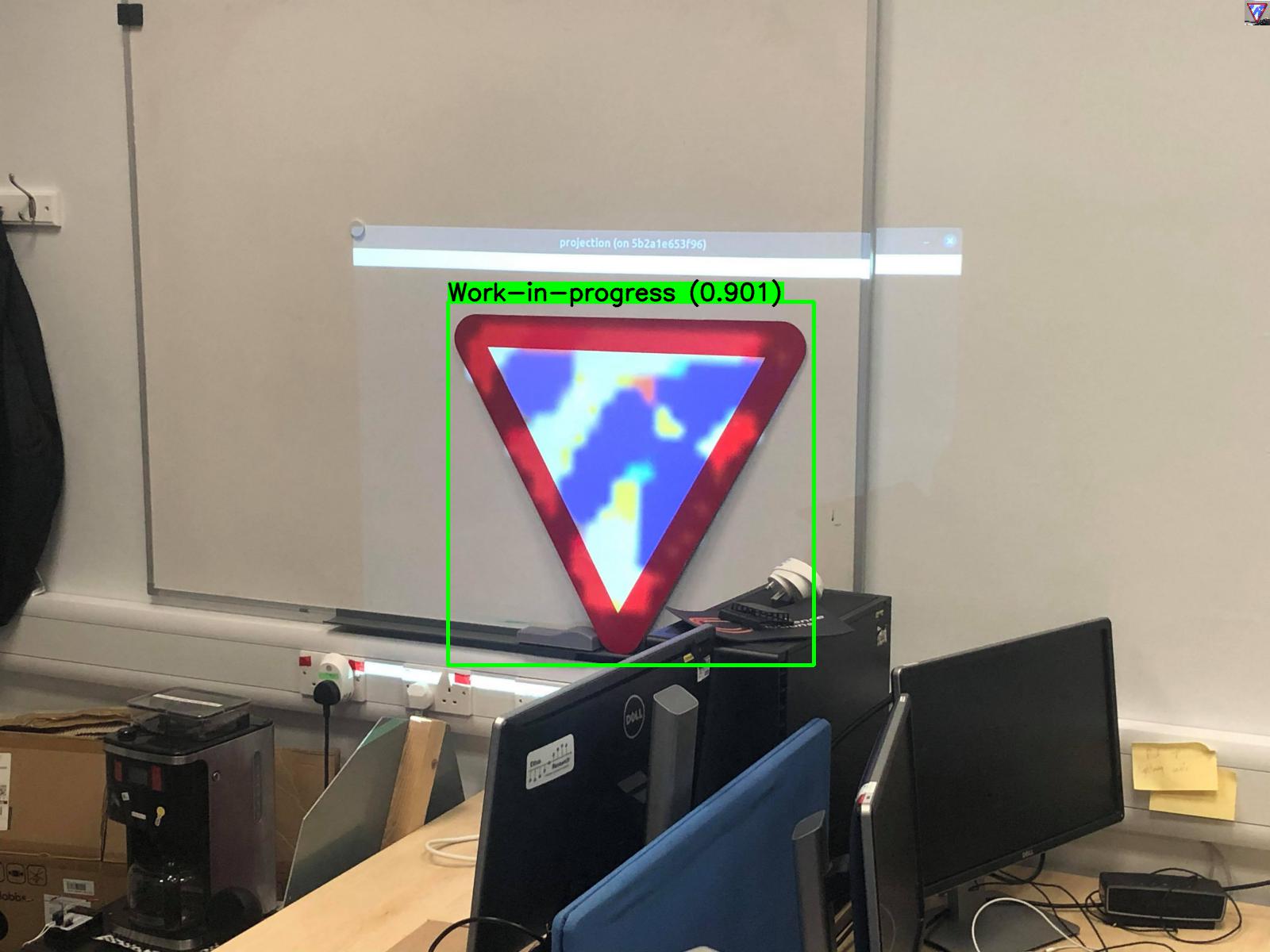}
	\end{subfigure}

	\begin{subfigure}[t]{.34\textwidth}
	\centering
	\includegraphics[width=\textwidth]{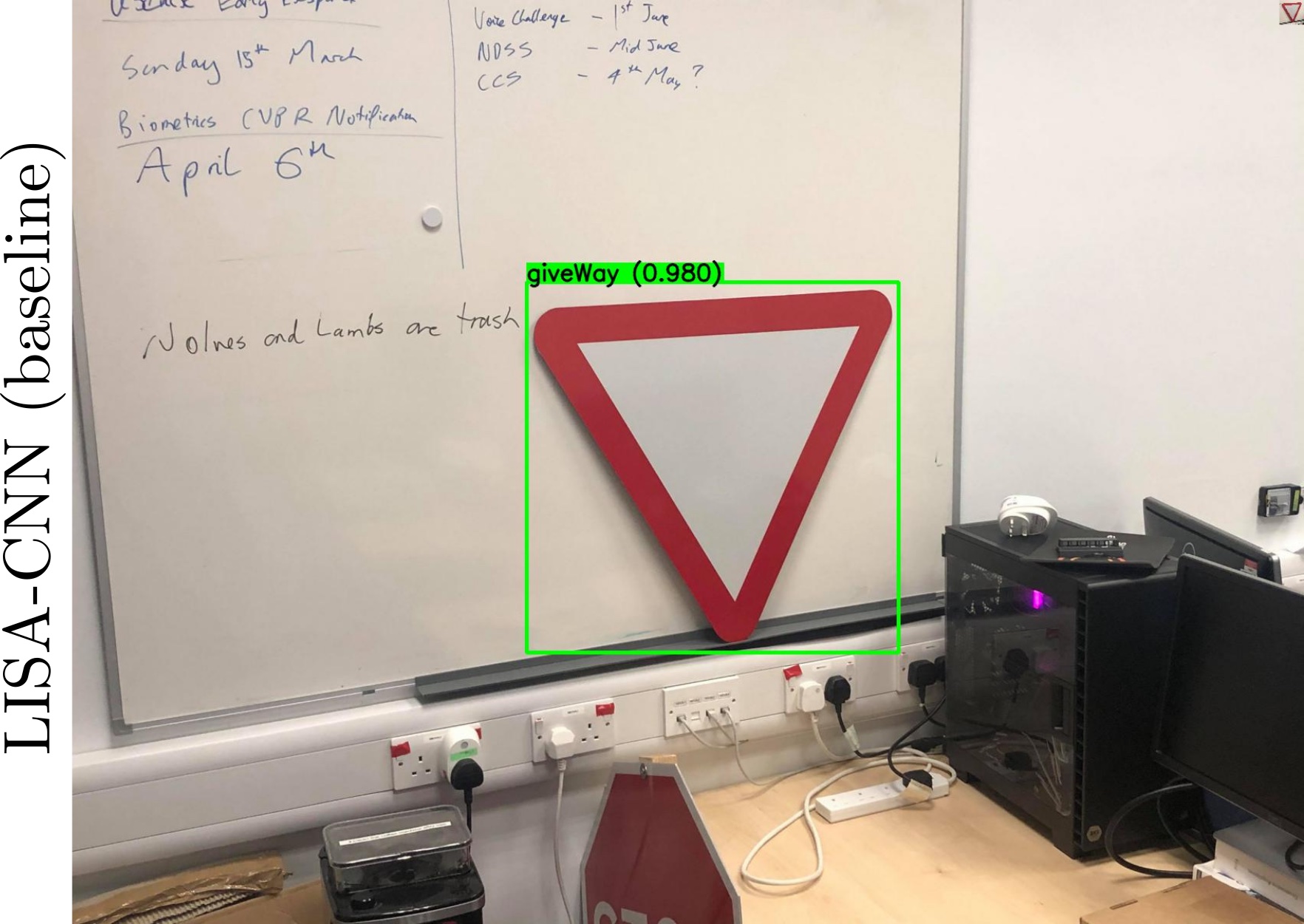}
	\end{subfigure}
	\hfill
	\begin{subfigure}[t]{.32\textwidth}
		\centering
		\includegraphics[width=\textwidth]{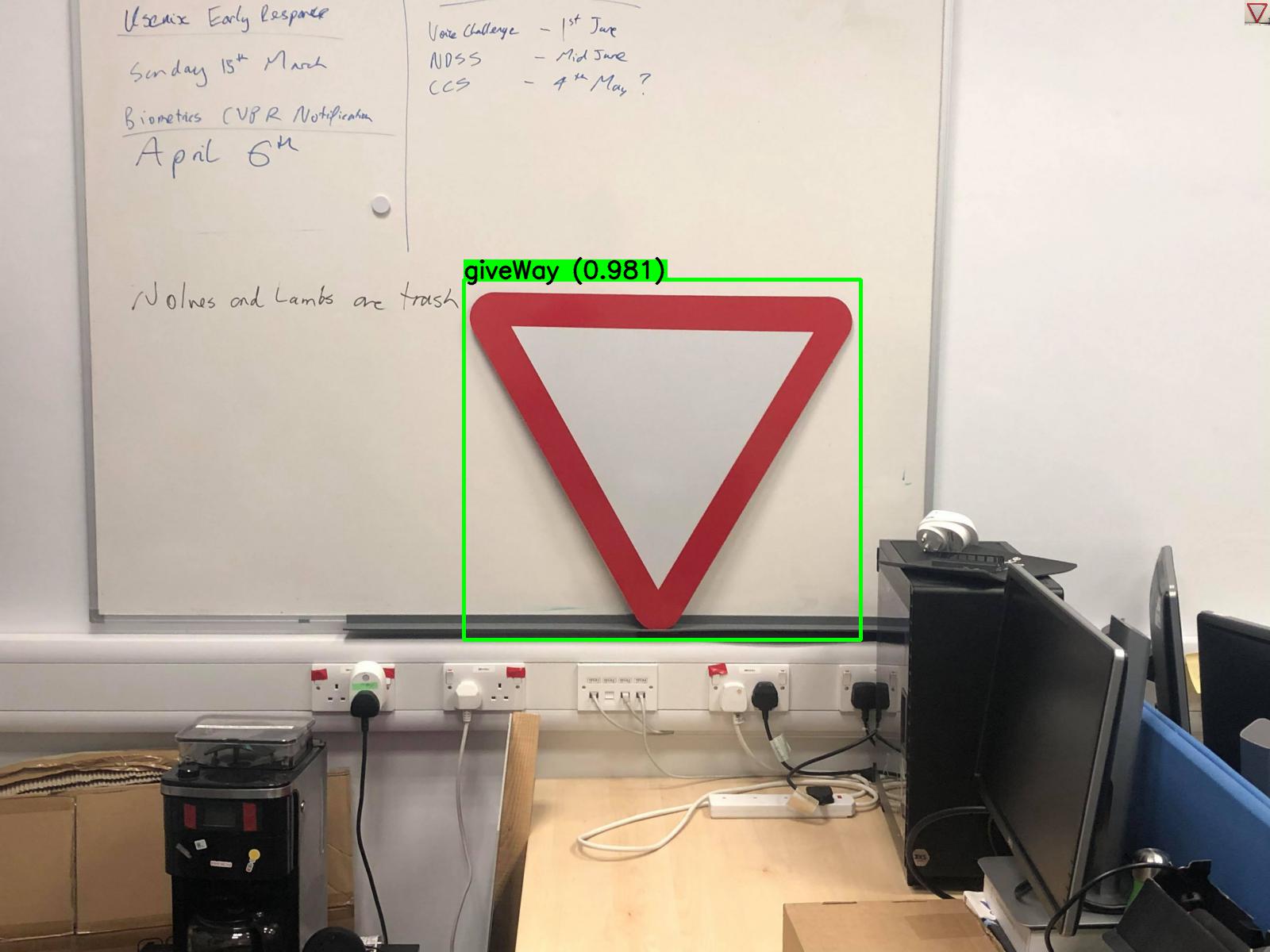}
	\end{subfigure}
	\hfill
	\begin{subfigure}[t]{.32\textwidth}
		\centering
		\includegraphics[width=\textwidth]{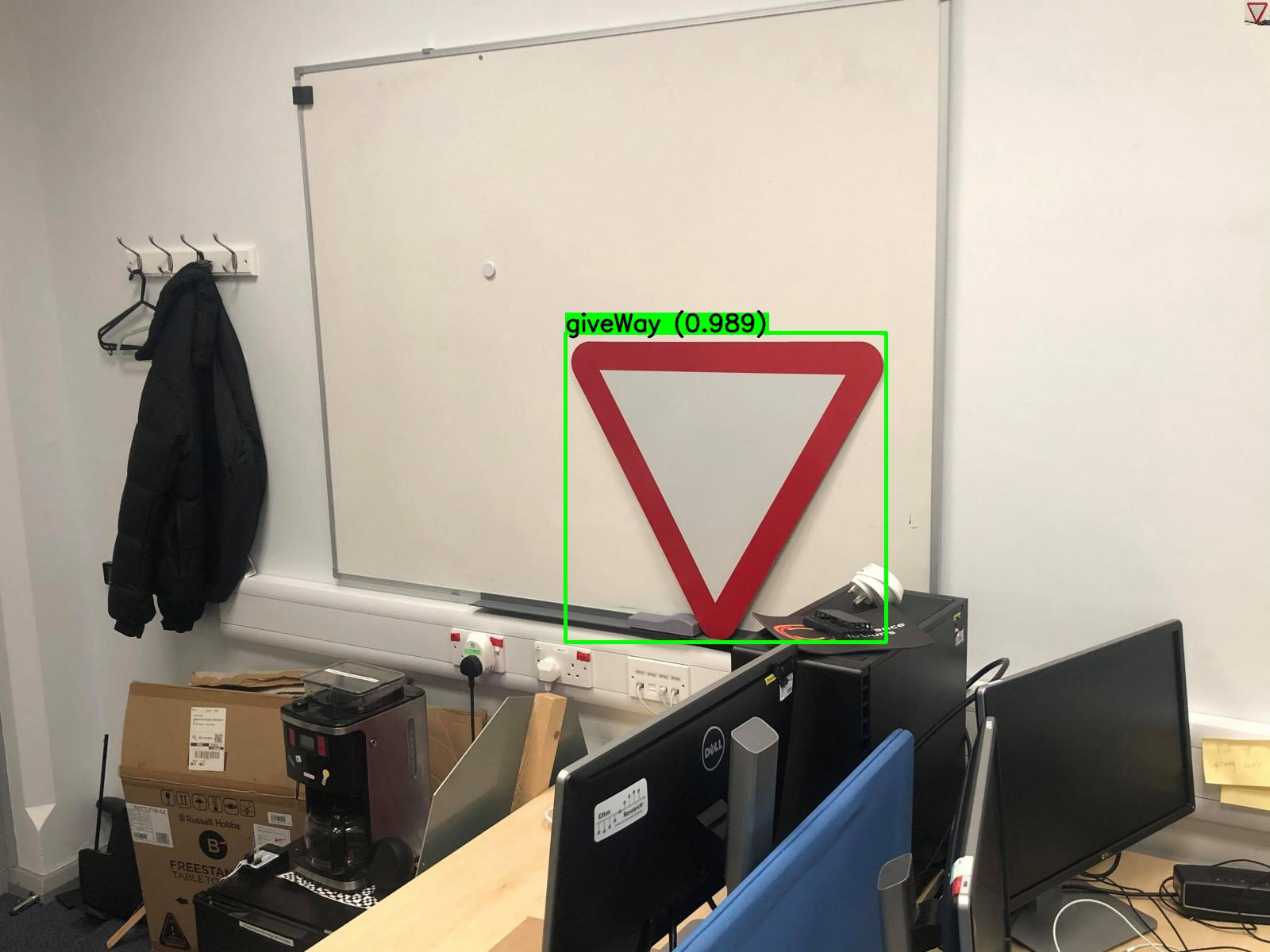}
	\end{subfigure}
	
	\begin{subfigure}[t]{.34\textwidth}
		\centering
		\includegraphics[width=\textwidth]{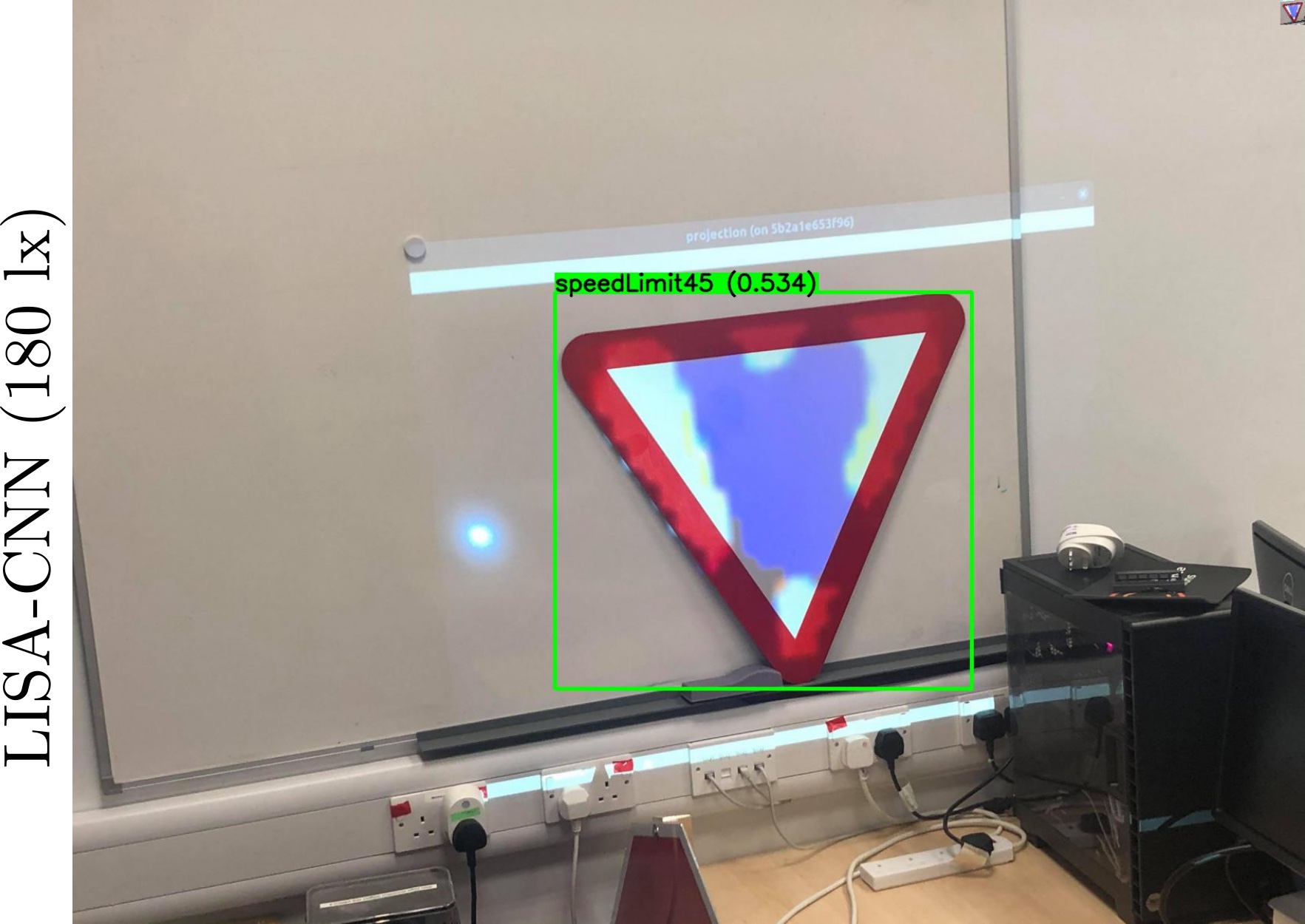}
	\end{subfigure}
	\hfill
	\begin{subfigure}[t]{.32\textwidth}
		\centering
		\includegraphics[width=\textwidth]{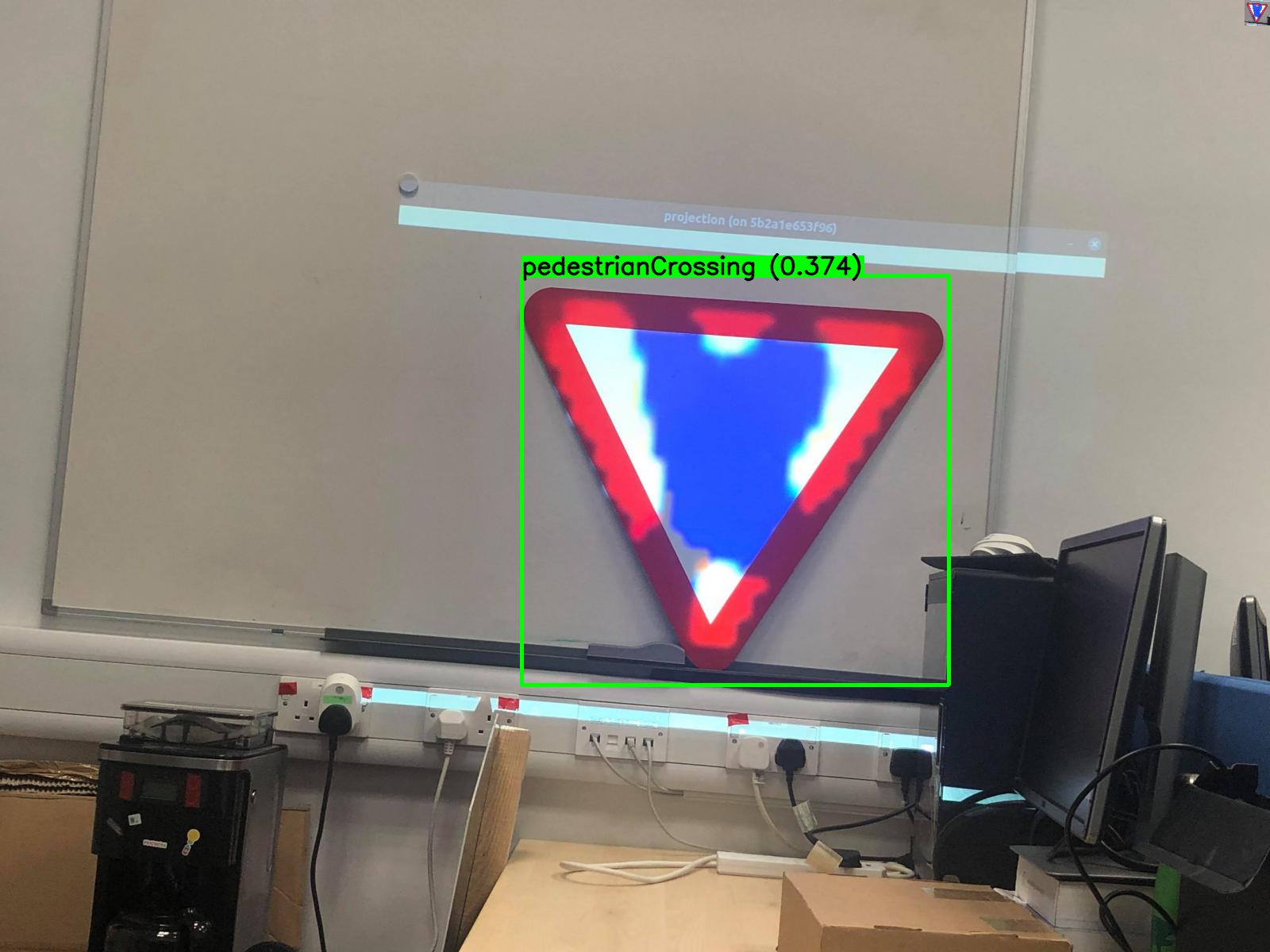}
	\end{subfigure}
	\hfill
	\begin{subfigure}[t]{.32\textwidth}
		\centering
		\includegraphics[width=\textwidth]{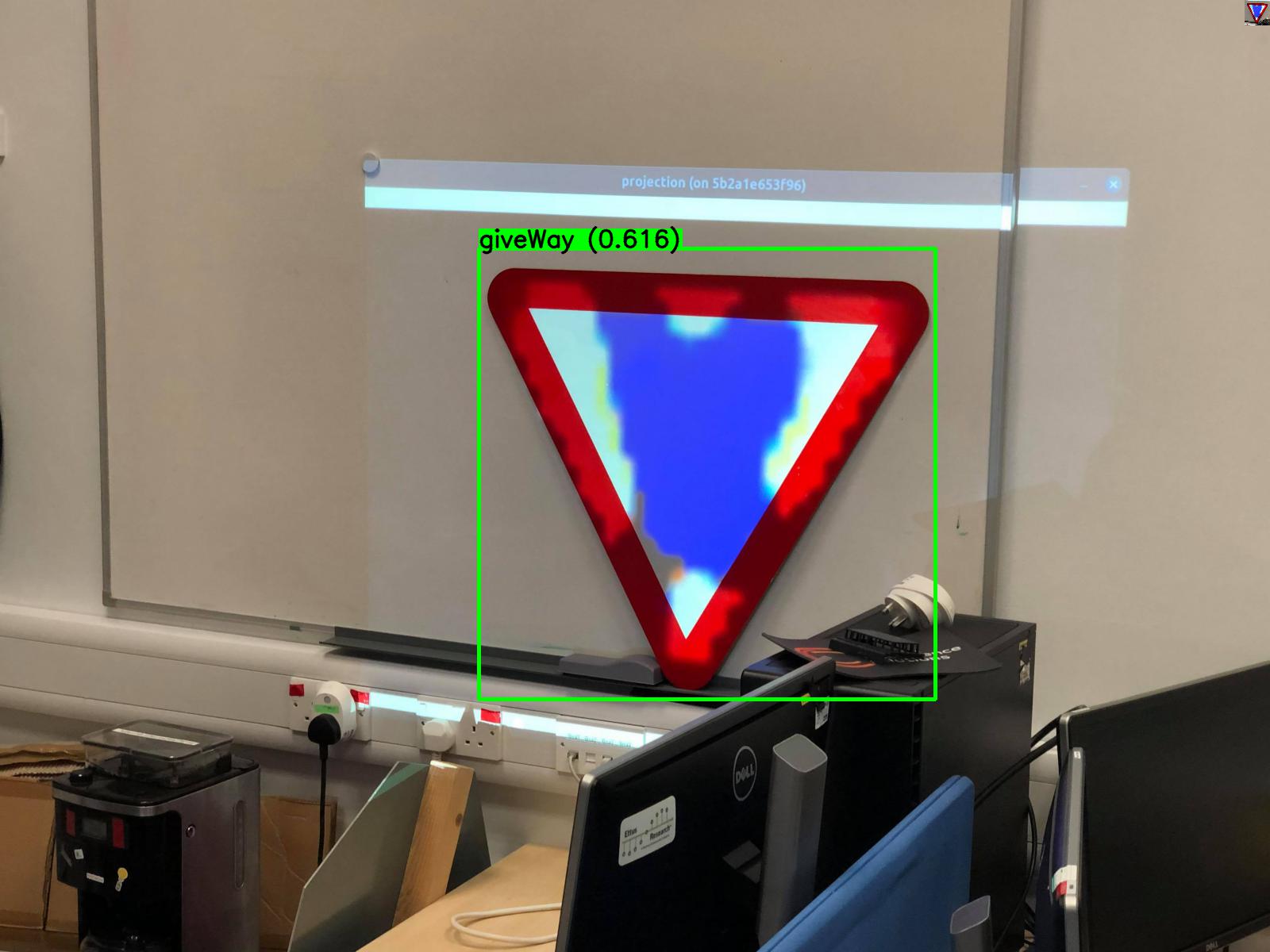}
	\end{subfigure}

	\caption{Attack on class ``Give Way'' for \gtsrb{} and \lisa{}.}
	\label{fig:other_target_app_giveway}
\end{figure*}

\begin{figure*}[t]
	\centering
	\begin{subfigure}[t]{.34\textwidth}
		\centering
		\includegraphics[width=\textwidth]{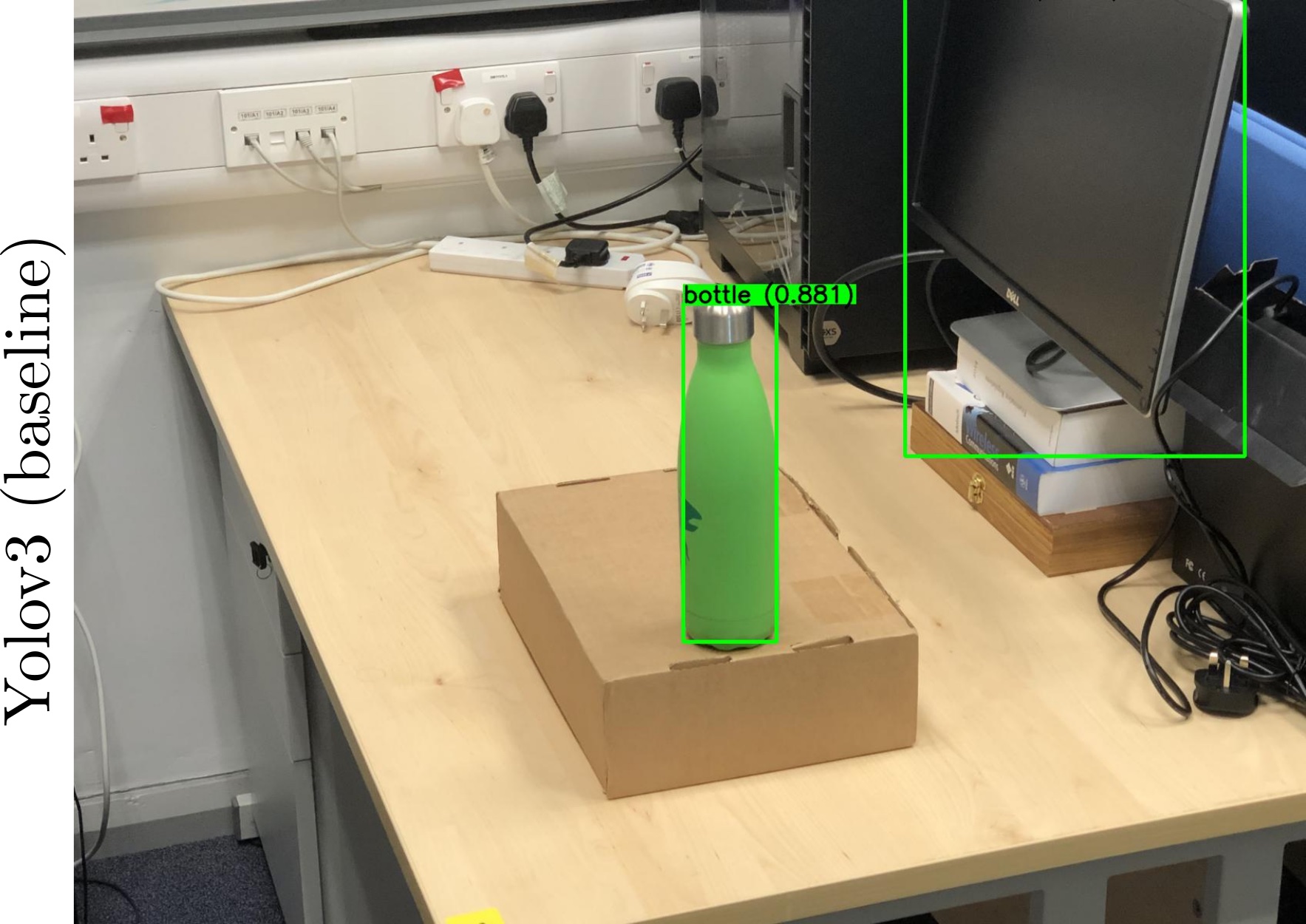}
	\end{subfigure}
	\hfill
	\begin{subfigure}[t]{.32\textwidth}
		\centering
		\includegraphics[width=\textwidth]{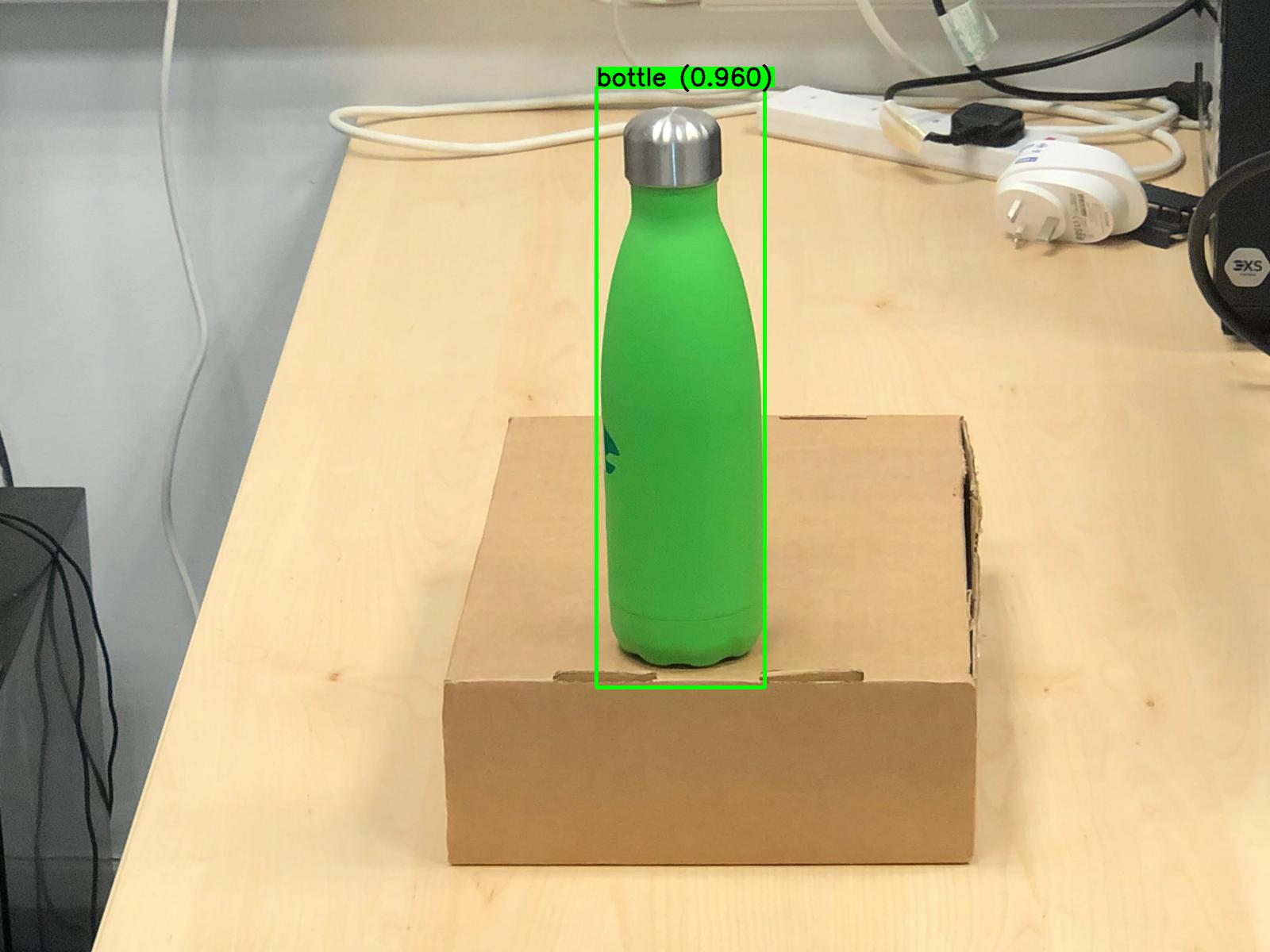}
	\end{subfigure}
	\hfill
	\begin{subfigure}[t]{.32\textwidth}
		\centering
		\includegraphics[width=\textwidth]{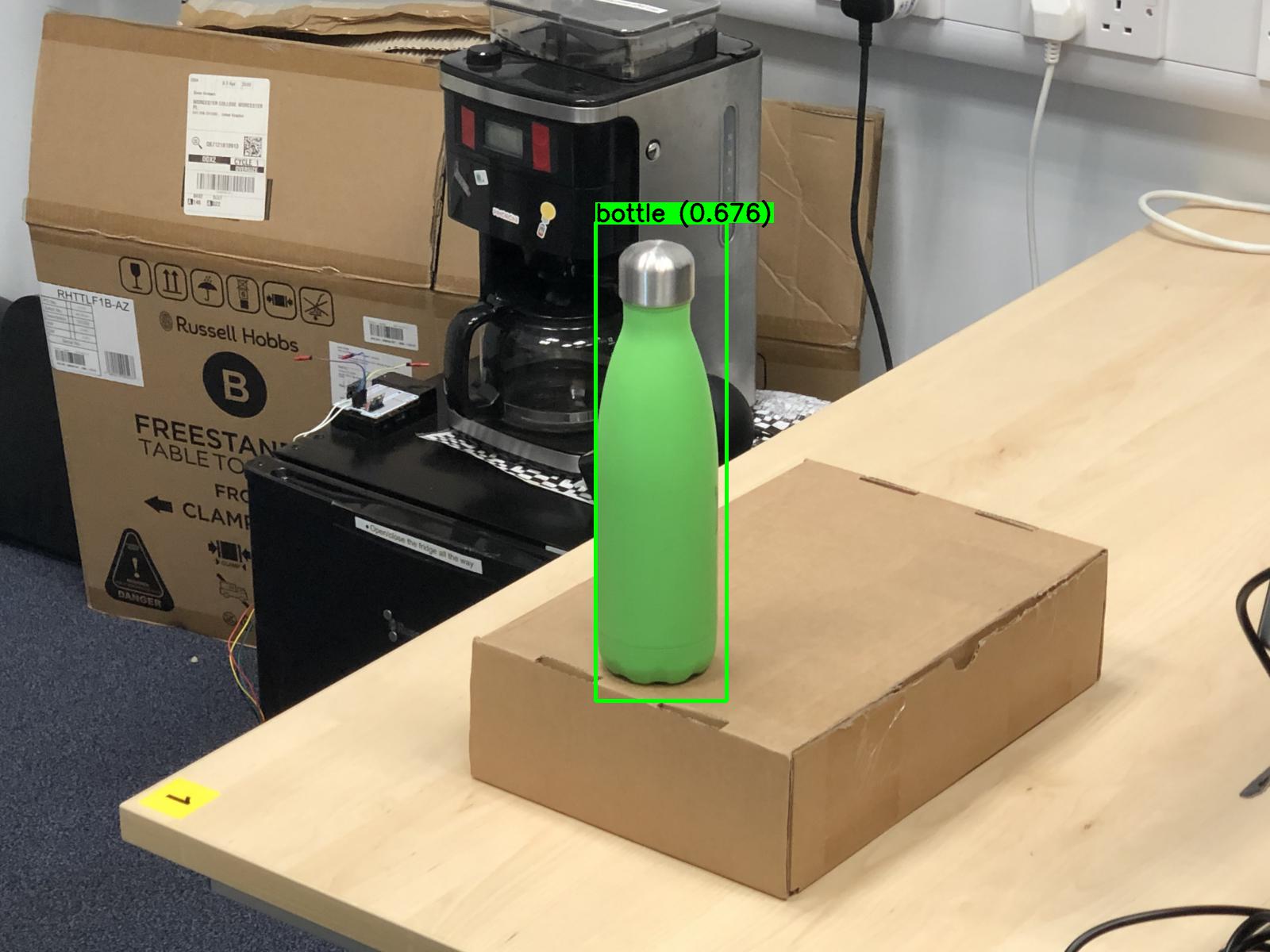}
	\end{subfigure}
	
	\begin{subfigure}[t]{.34\textwidth}
		\centering
		\includegraphics[width=\textwidth]{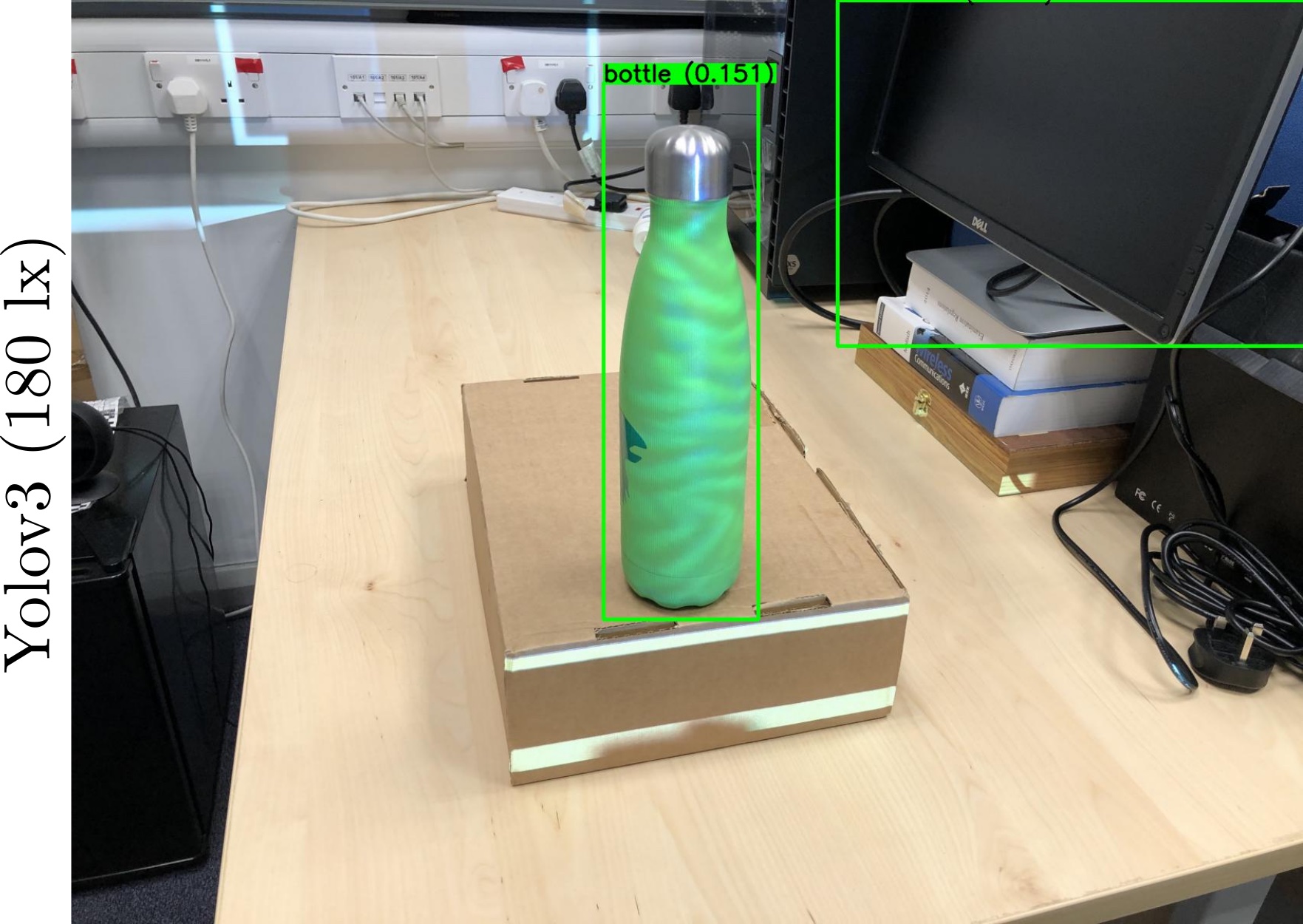}
	\end{subfigure}
	\hfill
	\begin{subfigure}[t]{.32\textwidth}
		\centering
		\includegraphics[width=\textwidth]{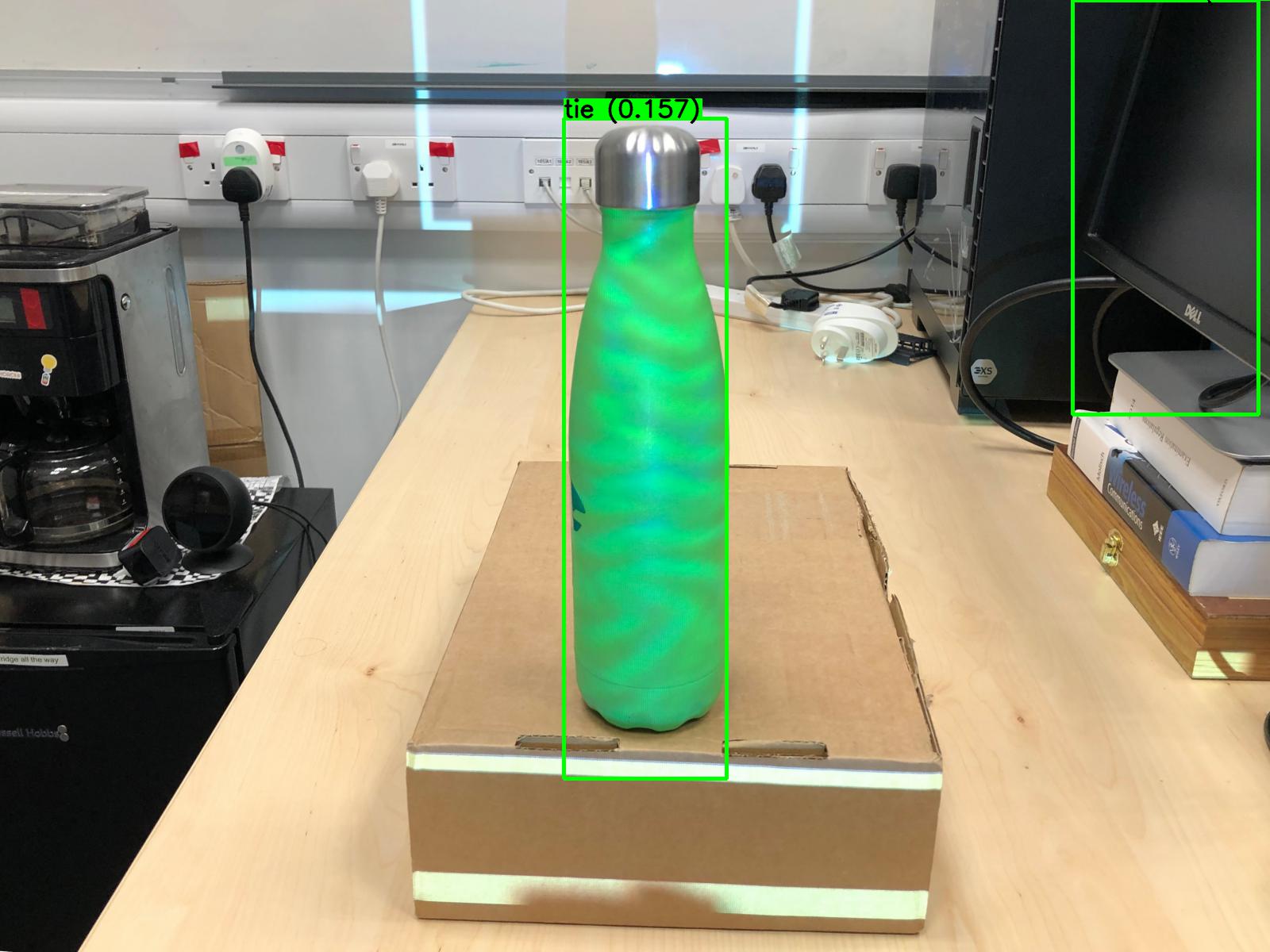}
	\end{subfigure}
	\hfill
	\begin{subfigure}[t]{.32\textwidth}
		\centering
		\includegraphics[width=\textwidth]{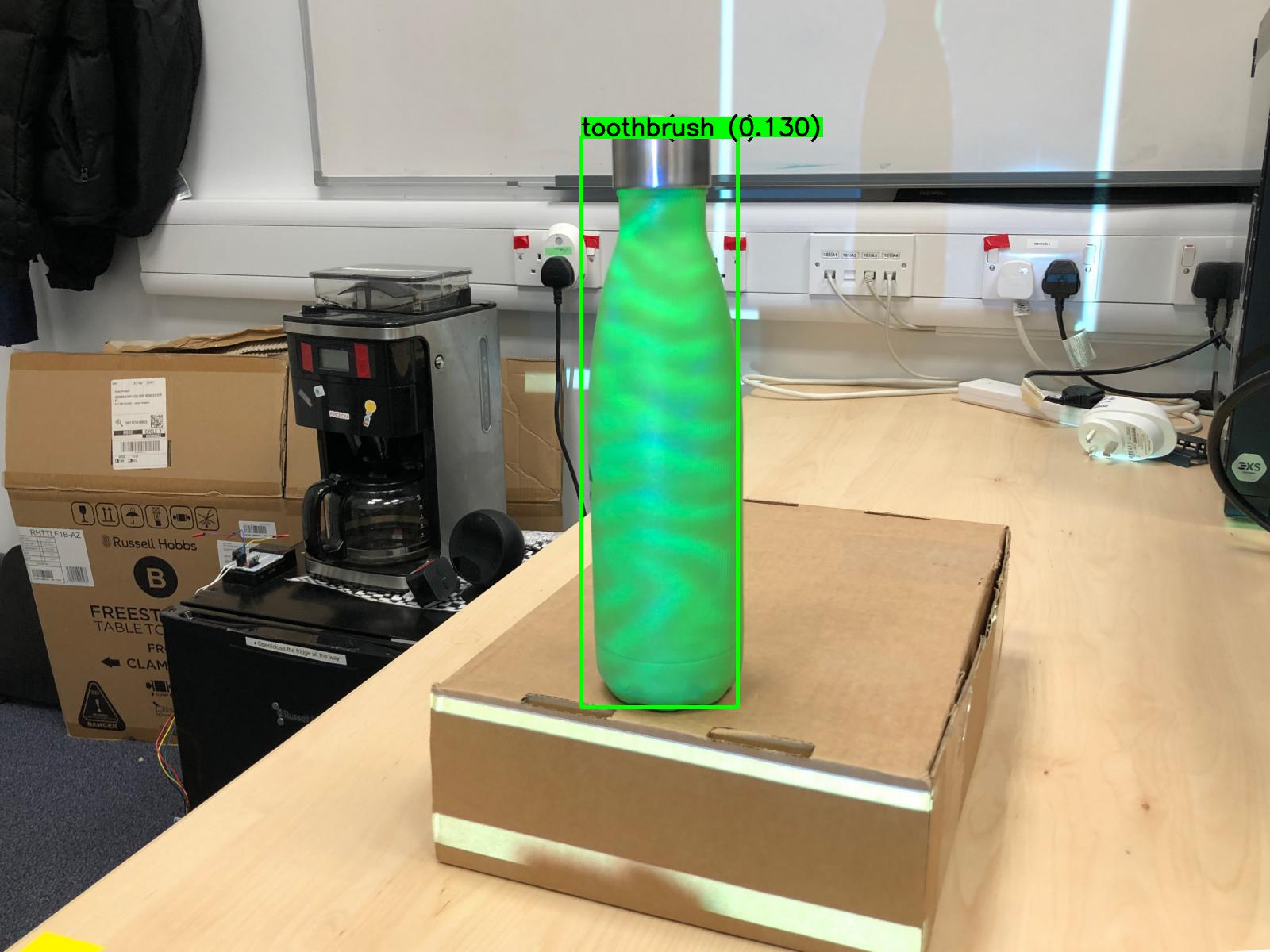}
	\end{subfigure}
	
	\begin{subfigure}[t]{.34\textwidth}
		\centering
		\includegraphics[width=\textwidth]{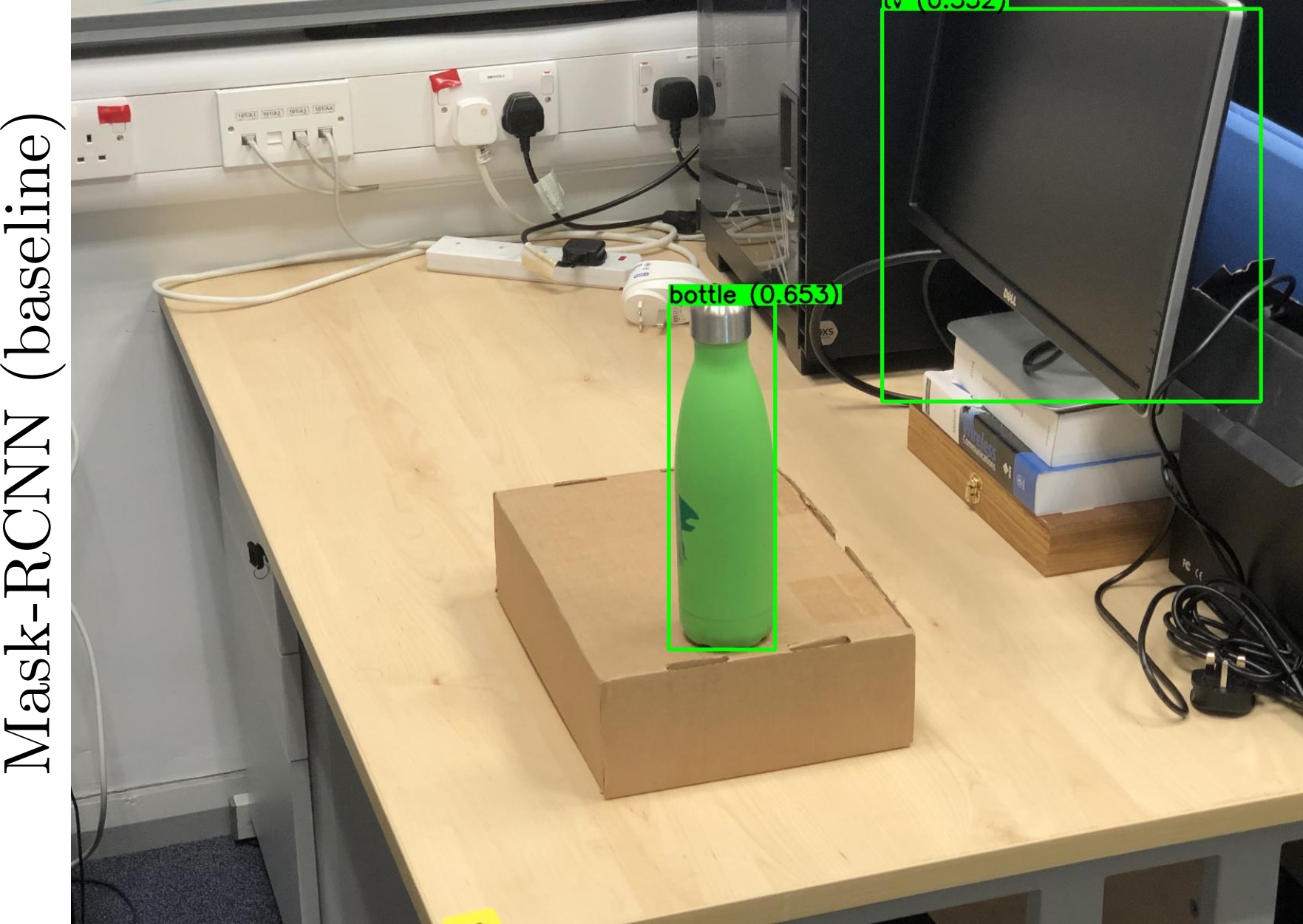}
	\end{subfigure}
	\hfill
	\begin{subfigure}[t]{.32\textwidth}
		\centering
		\includegraphics[width=\textwidth]{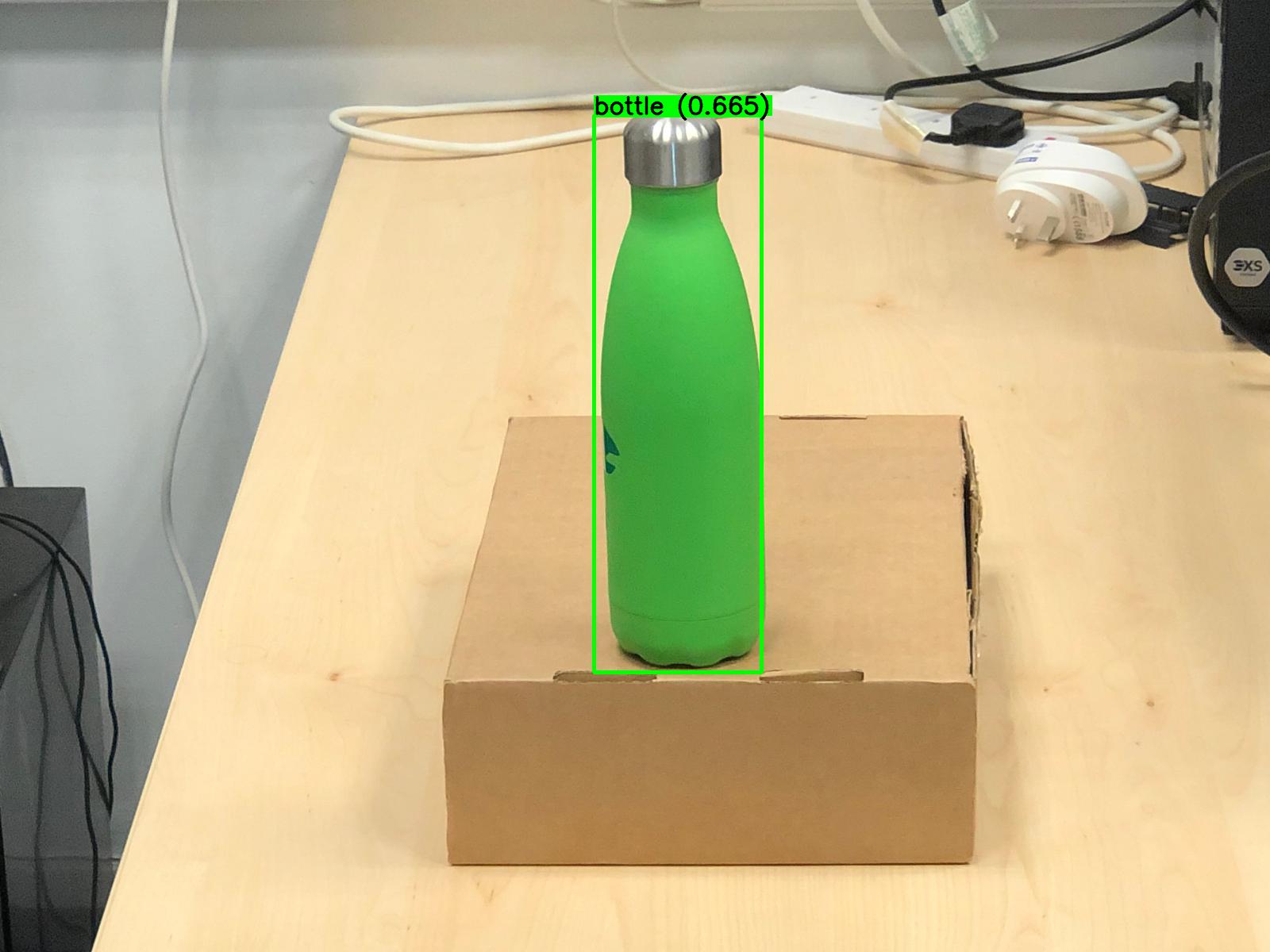}
	\end{subfigure}
	\hfill
	\begin{subfigure}[t]{.32\textwidth}
		\centering
		\includegraphics[width=\textwidth]{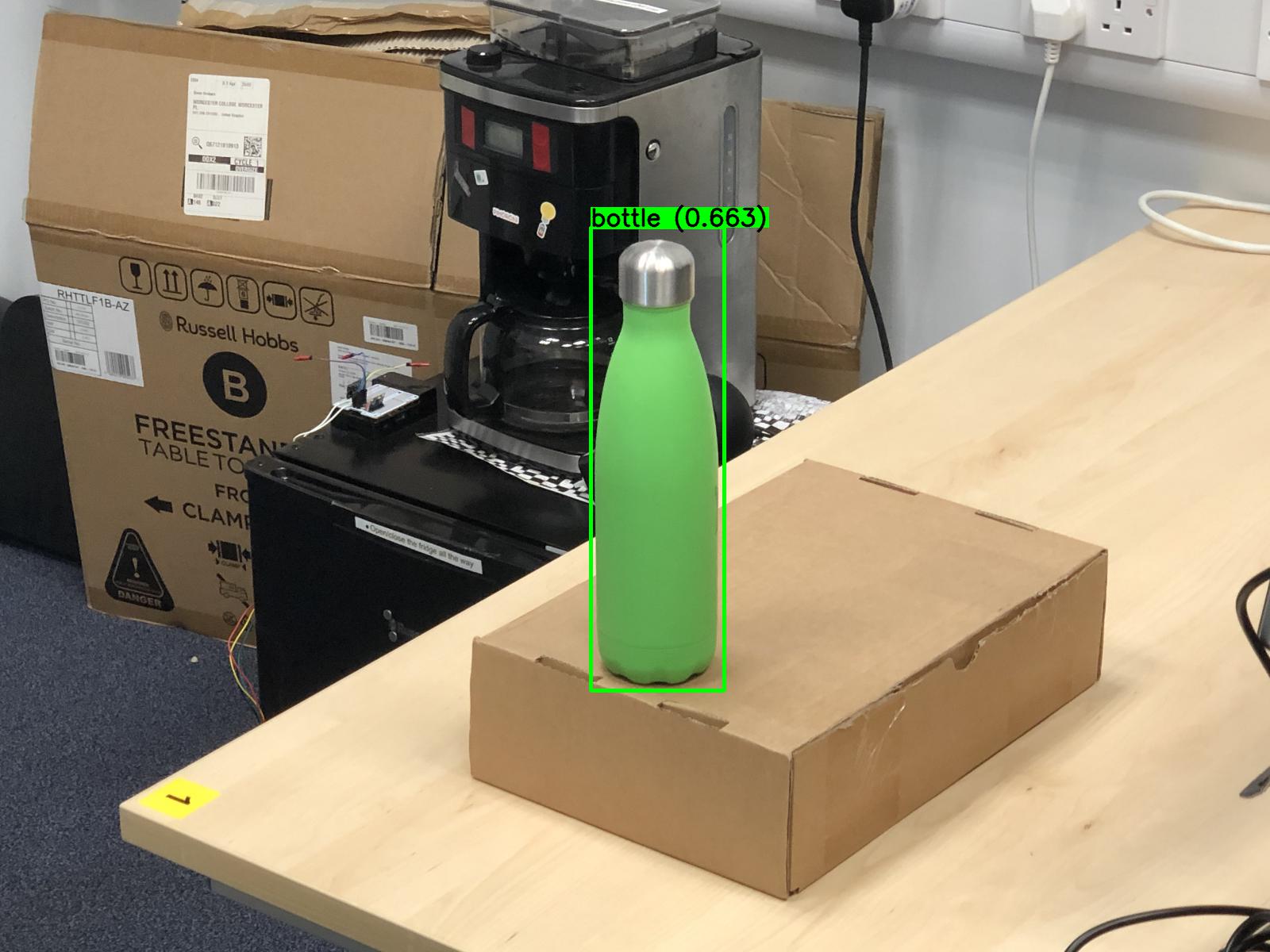}
	\end{subfigure}
	
	\begin{subfigure}[t]{.34\textwidth}
		\centering
		\includegraphics[width=\textwidth]{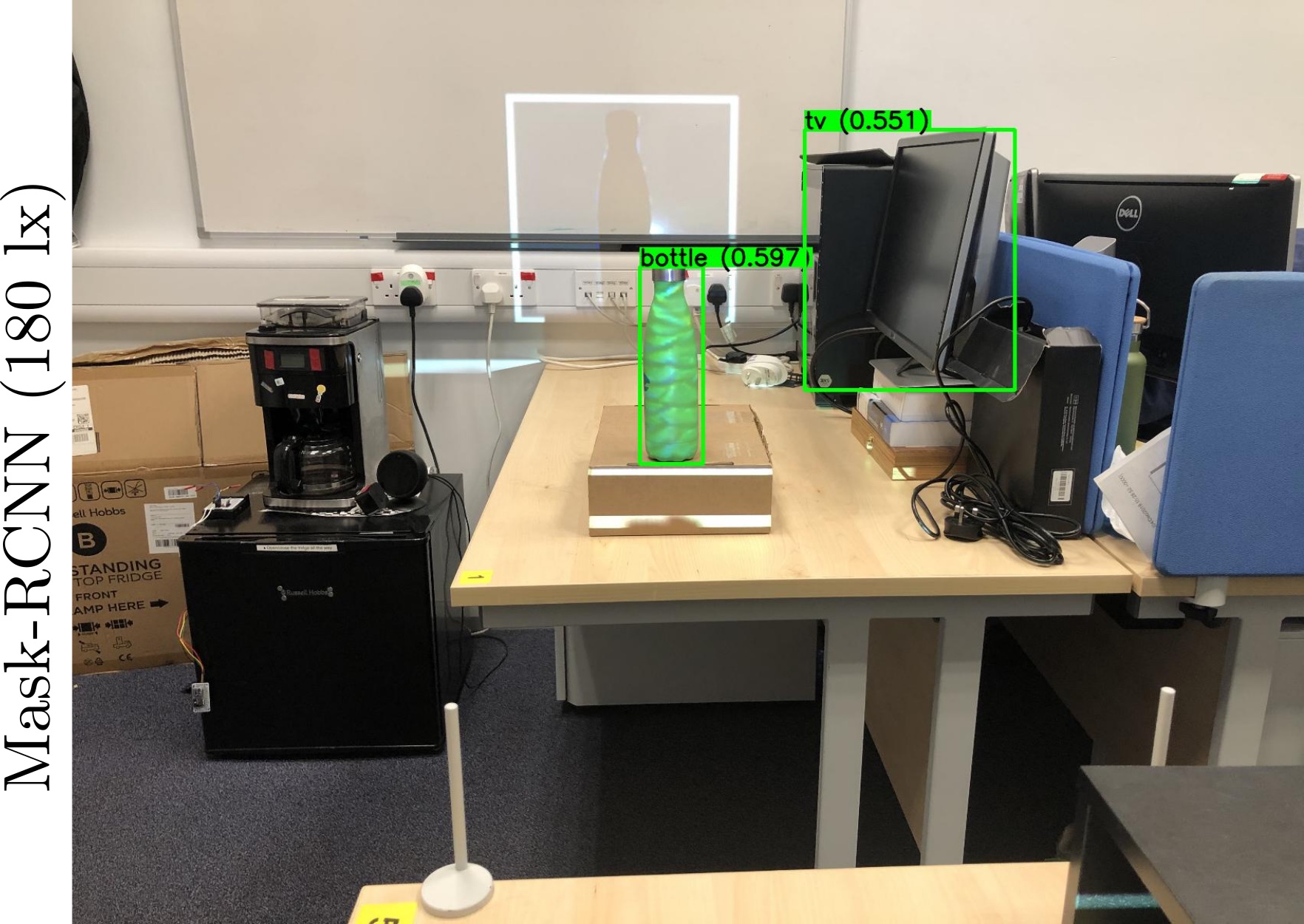}
	\end{subfigure}
	\hfill
	\begin{subfigure}[t]{.32\textwidth}
		\centering
		\includegraphics[width=\textwidth]{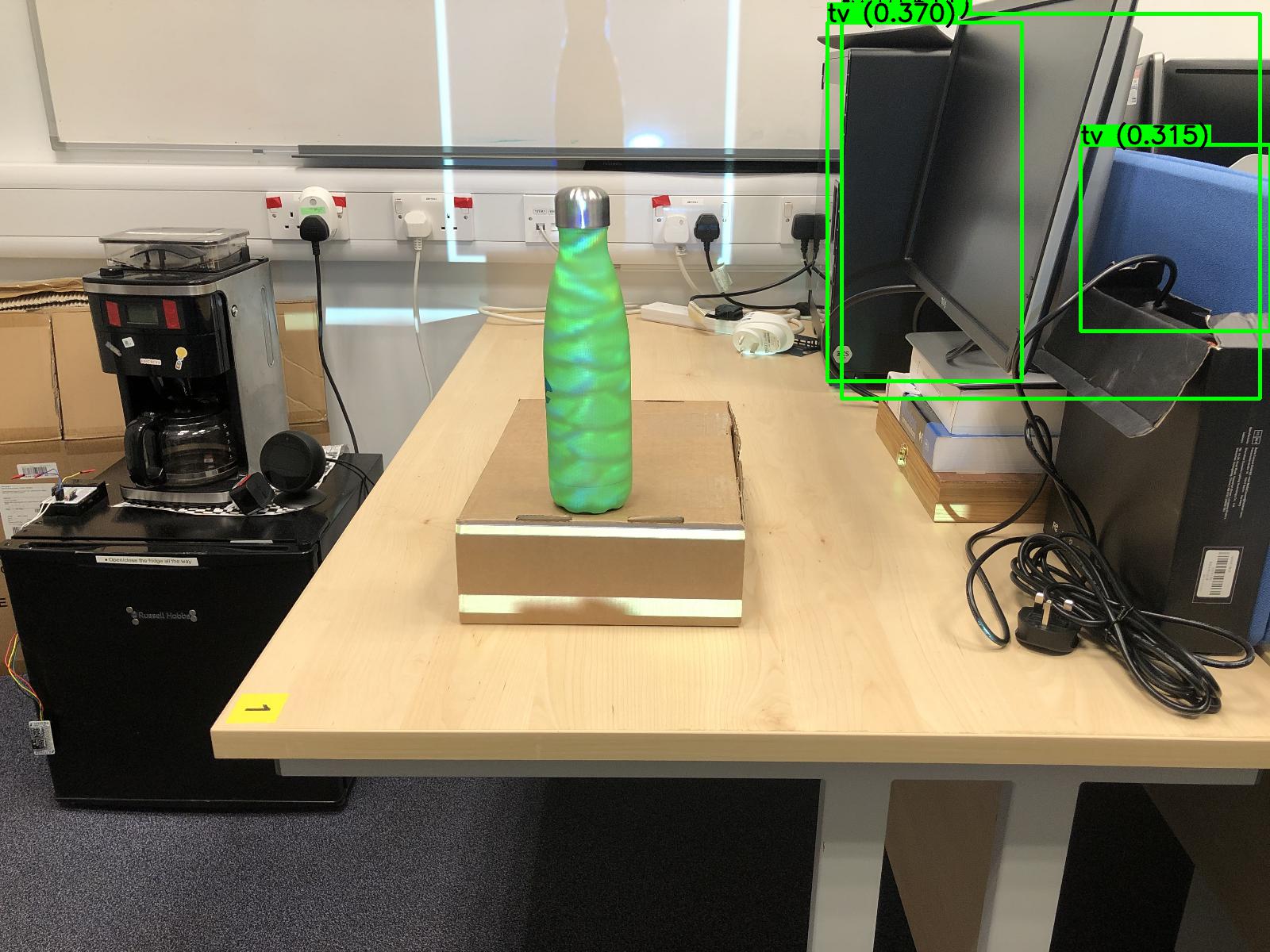}
	\end{subfigure}
	\hfill
	\begin{subfigure}[t]{.32\textwidth}
		\centering
		\includegraphics[width=\textwidth]{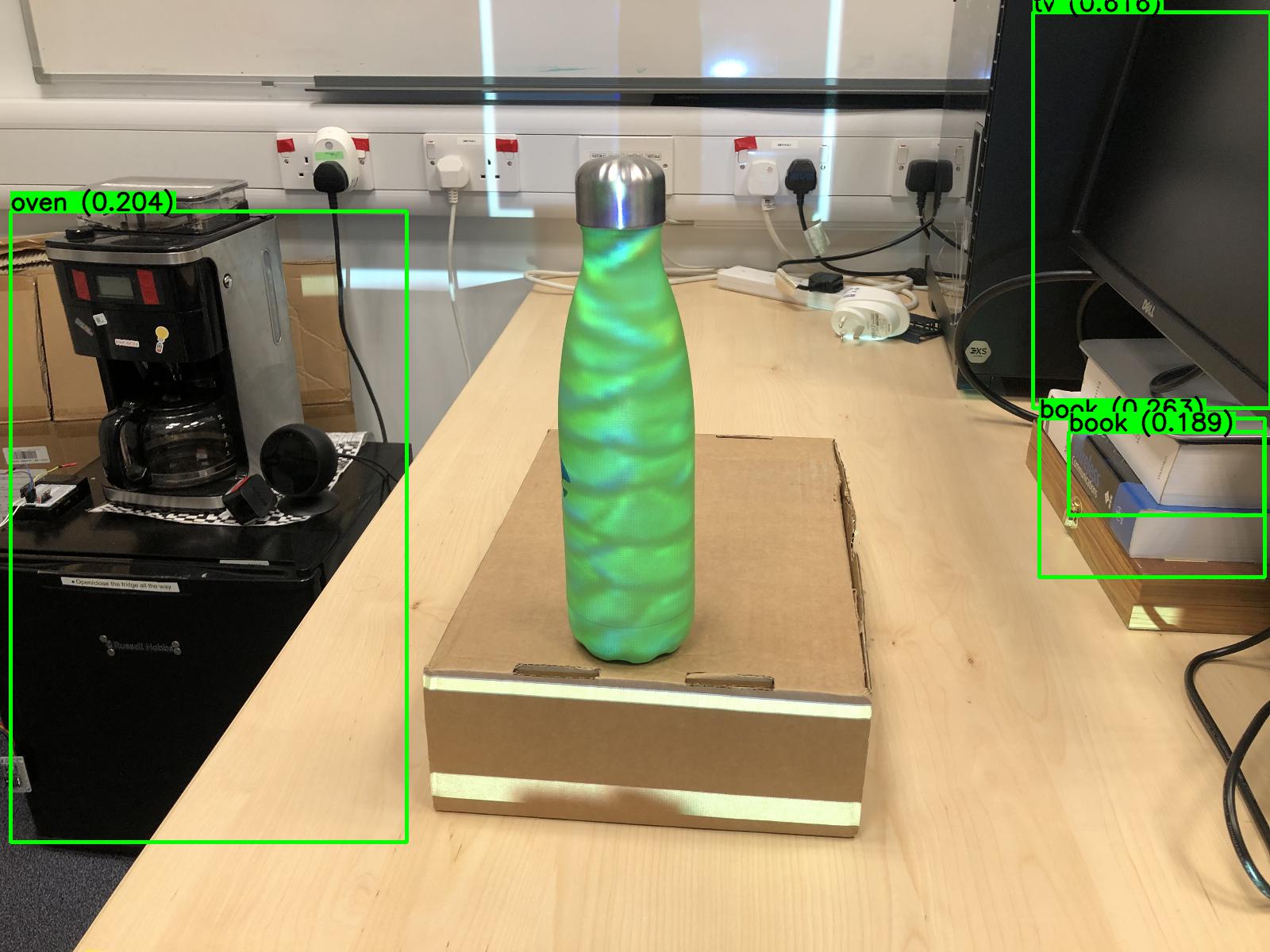}
	\end{subfigure}
	
	\caption{Attack on class ``Bottle'' for \yolo{} and \mrcnn{}. The detection thresholds used in the paper are 0.4 and 0.6, respectively.}
	\label{fig:other_target_app_bottle}
\end{figure*}

\myparagraph{Results}
We report example frames of successful attack on other objects in Figure~\ref{fig:other_target_app_giveway} and in Figure~\ref{fig:other_target_app_bottle}.
These include legitimate frames where the classification works correctly.
All the pictures are taken in 180 lux ambient light.
For \mrcnn{} and \yolo{} we restricted the bottle size to [150, 250], meaning that the bottle is generally in the foreground; we noticed that bottles of smaller sizes (in the background) are hardly detected.

\section{Additional Results}\label{sec:additional_results_app}

In the following subsections we report the entirety of the results for each of the experiment subsections in Section~\ref{sec:evaluation}. This includes extended results for the detection performance in controlled settings(Section~\ref{sec:app:det_results}), extended results for the road driving test (Section~\ref{sec:app:rdt}) and extended results for the transferability-based cross-network attack  (Section~\ref{sec:app:at}).

\subsection{Detection Results}\label{sec:app:det_results}
We report in Figure~\ref{fig:cone_app} the results for all the measured light settings in the controlled indoor experiment. These include 120, 180, 300, 440 and 600 lux.

\begin{figure*}[t]
	\centering
	\includegraphics[width=\textwidth]{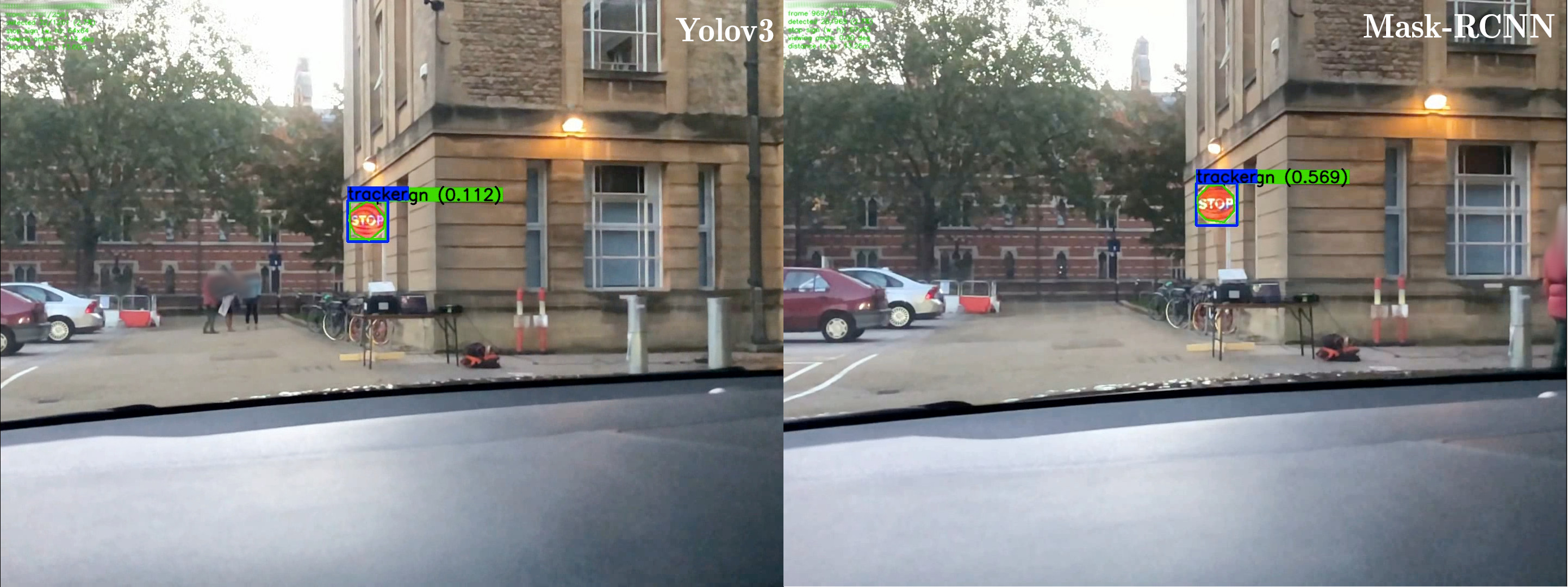}
	\caption{Sample video frames extracted from the outdoor experiment, for the 120 lux setting. Both frames show the stop sign undetected under the threshold set in the experiments. The blue `tracker` box is set manually and tracks the location of the sign.}
	\label{fig:approach_road_sc}
\end{figure*}

\begin{figure*}[t]
	\centering
	\includegraphics[width=\textwidth]{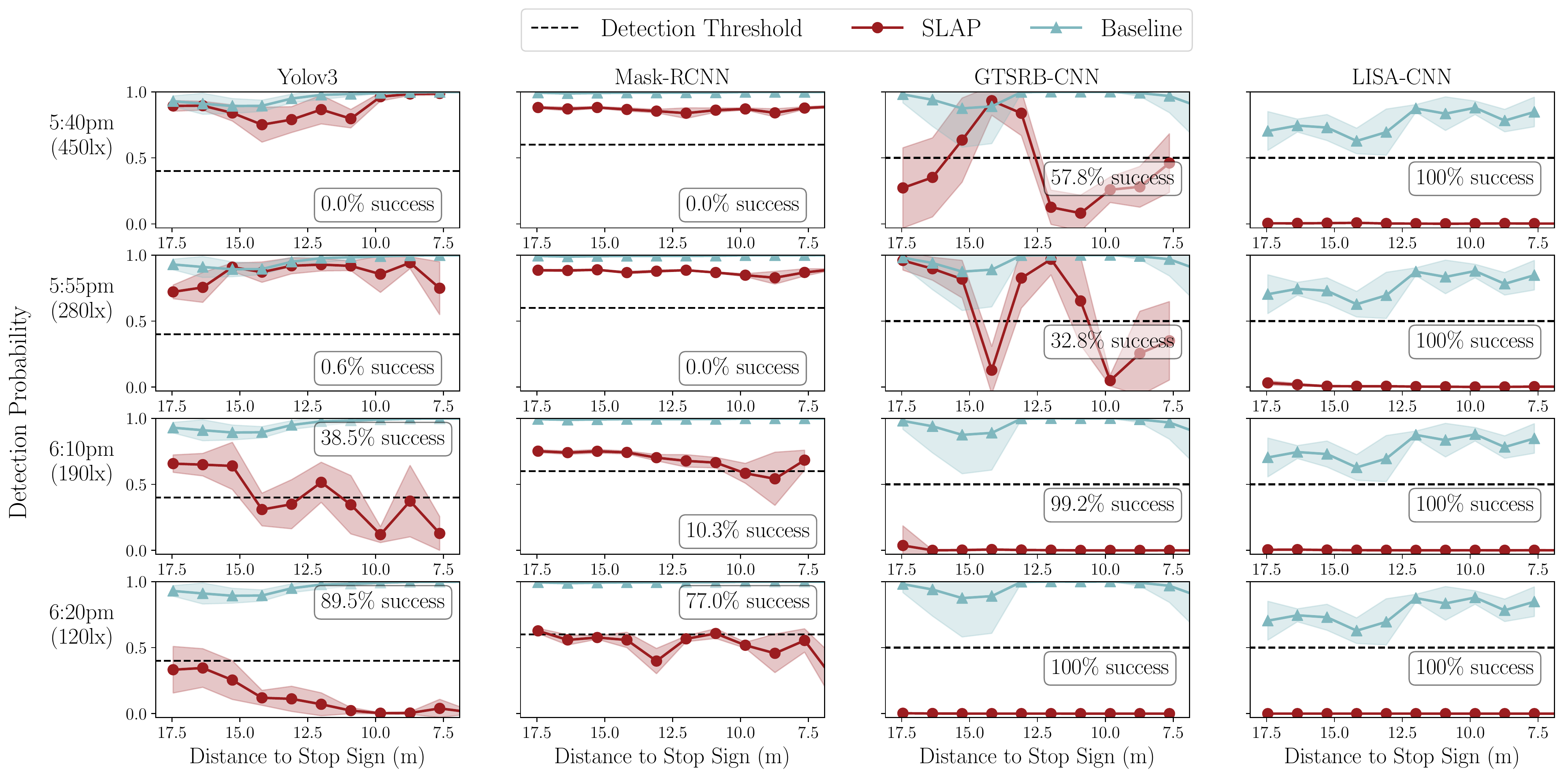}
	\caption{Full results for the experiment outdoors with a moving vehicle. We measured four different lighting conditions, 450, 280, 190 and 120 lux, each is reported in one row. Each column corresponds to one of the models investigated. The experiments are collected on a clear day shortly before the time of sunset (that is reported for 6:23pm on that day). Each plot report the detection threshold used by the network and the baseline performance of the model in non-adversarial settings.}
	\label{fig:approach_road_app}
\end{figure*}

\newcommand{\conessize}{0.18}

\subsection{Road Driving Test}\label{sec:app:rdt}
We report in Figure~\ref{fig:approach_road_app} the full results for all the measured light settings in the outdoors driving test.
We also report in Figure~\ref{fig:approach_road_sc} example frames taken from the outdoor driving test, for the 120 lux settings (these are taken roughly at 5:40pm local time).

\subsection{Attack Transferability}\label{sec:app:at}

We report in Table~\ref{tab:transferability_app} all the results for the transferability. This includes each pair of the evaluated models, including \lisa{}$^{(s)}$ and  \gtsrb{}$^{(s)}$ which are trained with cross-entropy loss from scratch.

\section{SentiNet Description}\label{sec:sentinet_description}
\myparagraph{Rationale} We picked SentiNet for the evaluation because it was one of the few (if not the only) defences that was \textit{specifically} designed to detect physical adversarial examples (AE).
In fact, there is a plethora of works that creates physical AE by using stickers (or patches) that are placed on the targeted objects.
The insight behind SentiNet is that these patches are the most common way to create physical AE, but generate small image areas with large saliency.
This is not only a \textit{detectable} behavior in general, but it is also \textit{unavoidable} for the attacker to escape such behavior when creating a physical AE (without replacing the entire object).

\myparagraph{Description} SentiNet is a system designed to detect adversarial examples leveraging the intuition of locality of patches.
If an adversarial sample contains a patch which causes a misclassification, then the saliency of the area containing the patch will be high.
Therefore, the salient area will cause misclassifications on other legitimate samples when overlayed onto them.
To compute the salient areas in input SentiNet uses GradCam++~\cite{chattopadhay2018grad}, which backpropagates the outputs to the last convolutional layer of the network and checks which region of the input lead to greater activations.
Since the resolution of this layer is only 4x4 for both \gtsrb{} and \lisa{}, we instead use XRAI~\cite{kapishnikov2019xrai}, a newer and more accurate method to compute salient areas.
We found that using GradCam made the output masks unusable as a resolution of 4x4 leads to coarse block like regions where salient areas cannot be accurately identified (resolution also is pointed out as a problem in the original paper~\cite{Chou2018}).
XRAI on the other hand produces saliency regions at the input resolution, leading to more granular salient areas, using an algorithm that incrementally grows salient regions.
As a consequence of this improved technique XRAI has been shown to outperform older saliency algorithms, producing higher quality, tightly bound saliency regions~\cite{kapishnikov2019xrai}.

SentiNet computes a threshold function which separates AE from benign images.
The threshold function is computed using: (i) the \textit{Average Confidence}, i.e., the average confidence of the network prediction made on benign test images where salient masks are replaced with inert patterns added to them and  (ii) the \textit{Fooled Percentage}, i.e., the percentage of benign test images where overlaying the salient mask leads the network to predict the suspected adversarial class.
These two scores characterize benign behaviour and can almost perfectly separate benign from adversarial inputs in SentiNet.
We follow the same technique as in the original paper for fitting the threshold function that separates the malicious and benign data.

\begin{figure*}[t]
	\centering
	\begin{subfigure}[t]{\conessize\textwidth}
		\centering
		\includegraphics[width=\textwidth]{figures/lta6_yolov3_yolov3_cone}
		\caption%
		{{\small \yolo{}.}}
		\label{fig:120_cone_yolov3_app}
	\end{subfigure}
	\hfill
	\begin{subfigure}[t]{\conessize\textwidth}
		\centering
		\includegraphics[width=\textwidth]{figures/lta11_maskrcnn_maskrcnn_cone}
		\caption[]%
		{{\small \mrcnn{}.}}
		\label{fig:120_cone_mrcnn_app}
	\end{subfigure}
	\hfill
	\begin{subfigure}[t]{\conessize\textwidth}
		\centering
		\includegraphics[width=\textwidth]{figures/lta11_gtsrbcnn_gtsrbcnn_cone}
		\caption[]%
		{{\small \gtsrb{}.}}
		\label{fig:120_cone_gtsrbcnn_app}
	\end{subfigure}
	\hfill
	\begin{subfigure}[t]{\conessize\textwidth}
		\centering
		\includegraphics[width=\textwidth]{figures/lta6_lisacnn_lisacnn_cone}
		\caption[]%
		{{\small \lisa{}.}}
		\label{fig:120_cone_lisacnn_app}
	\end{subfigure}
	\hfill
	
	\begin{subfigure}[t]{\conessize\textwidth}
		\centering
		\includegraphics[width=\textwidth]{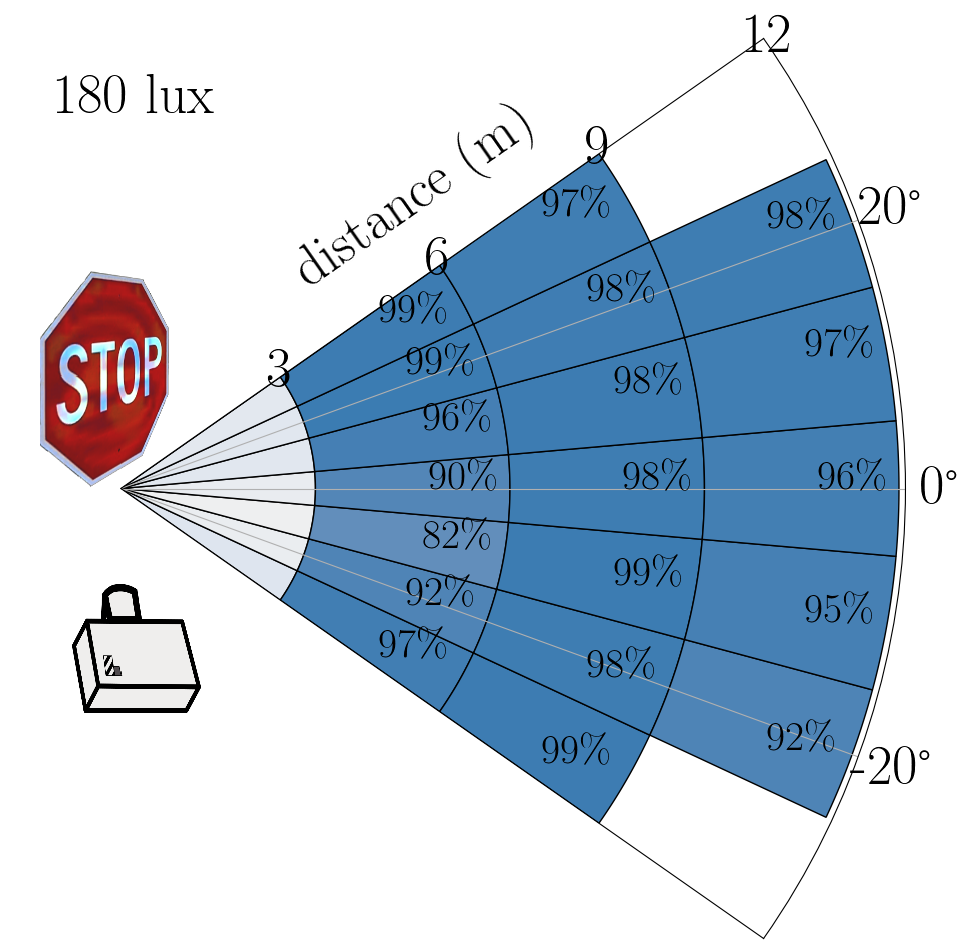}
		\caption%
		{{\small \yolo{}.}}
		\label{fig:180_cone_yolov3_app}
	\end{subfigure}
	\hfill
	\begin{subfigure}[t]{\conessize\textwidth}
		\centering
		\includegraphics[width=\textwidth]{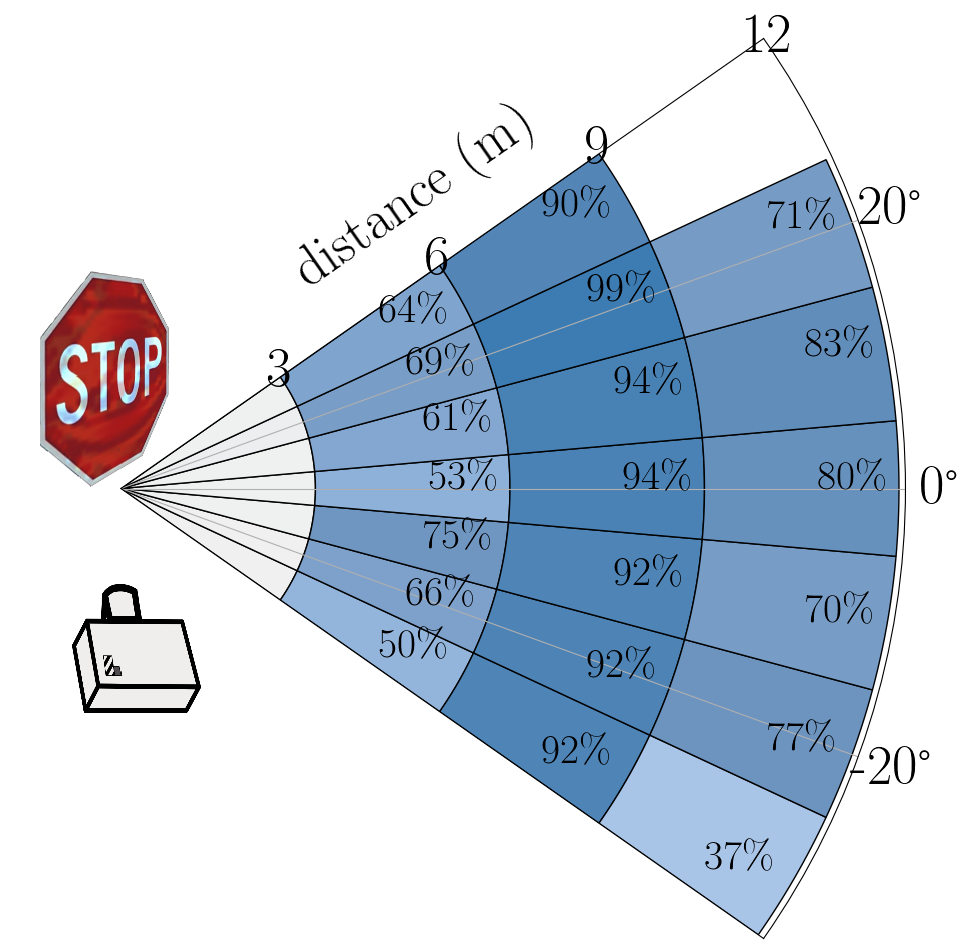}
		\caption[]%
		{{\small \mrcnn{}.}}
		\label{fig:180_cone_mrcnn_app}
	\end{subfigure}
	\hfill
	\begin{subfigure}[t]{\conessize\textwidth}
		\centering
		\includegraphics[width=\textwidth]{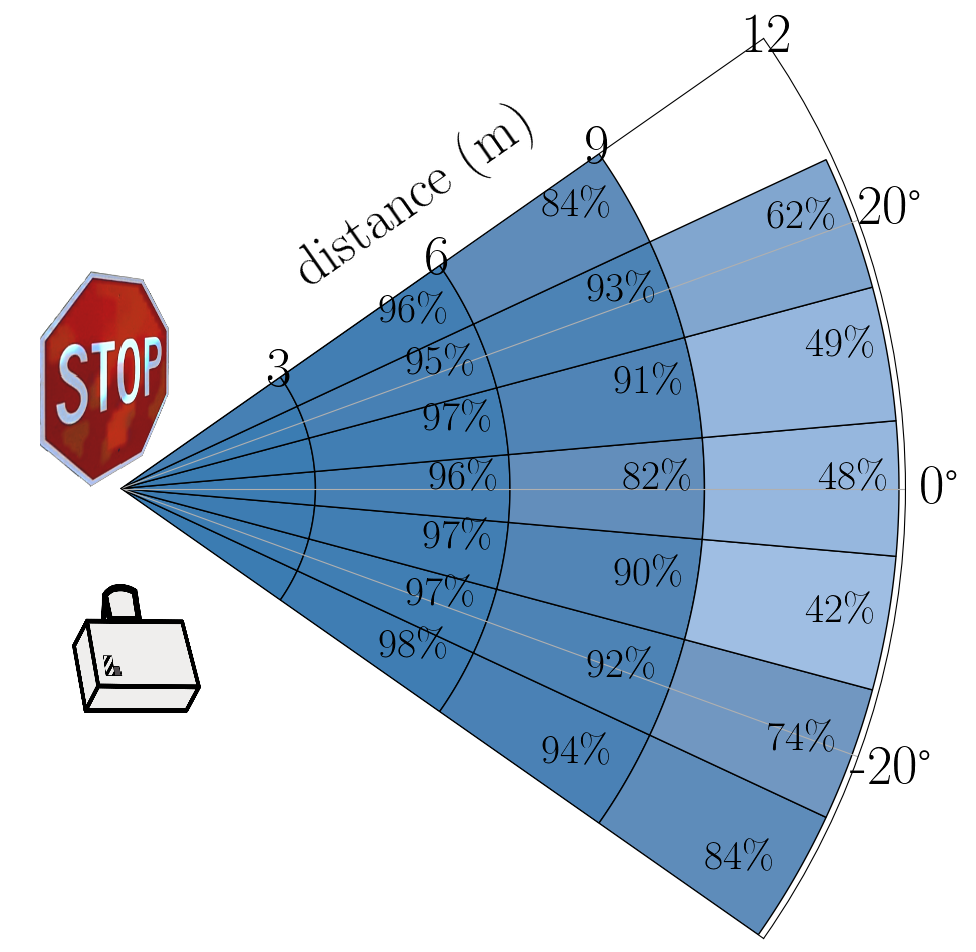}
		\caption[]%
		{{\small \gtsrb{}.}}
		\label{fig:180_cone_gtsrbcnn_app}
	\end{subfigure}
	\hfill
	\begin{subfigure}[t]{\conessize\textwidth}
		\centering
		\includegraphics[width=\textwidth]{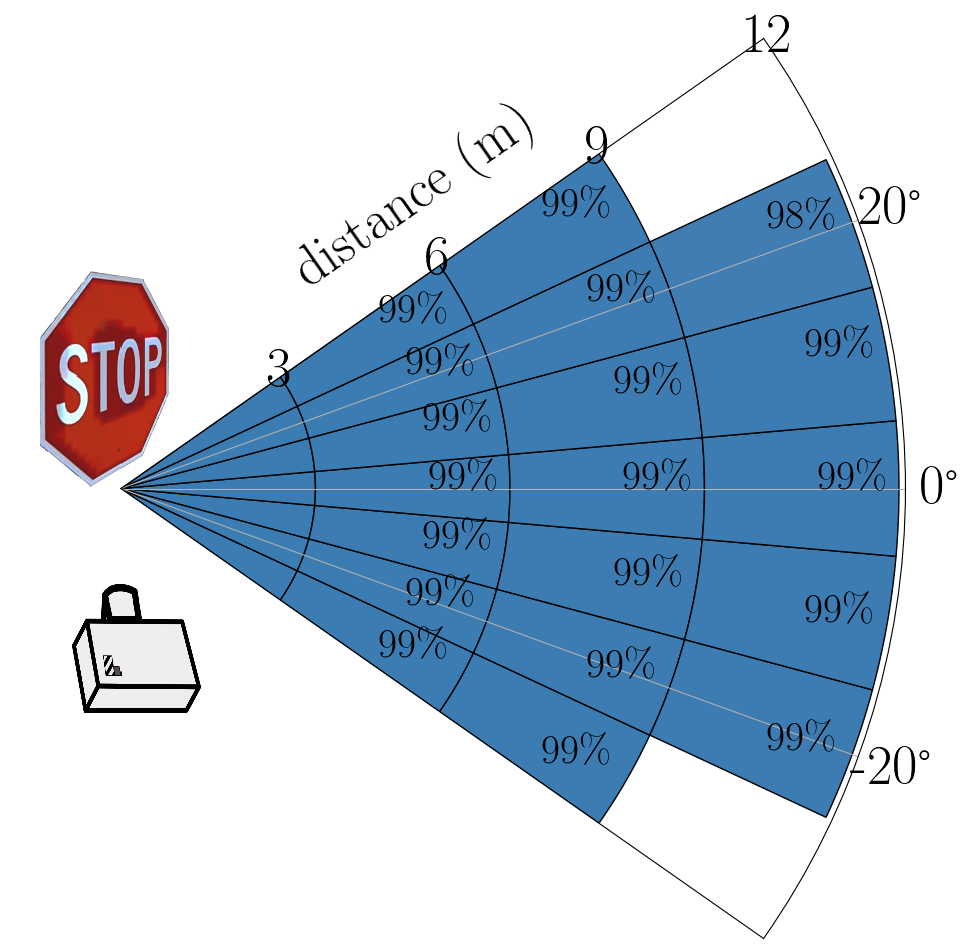}
		\caption[]%
		{{\small \lisa{}.}}
		\label{fig:180_cone_lisacnn_app}
	\end{subfigure}
	\hfill
	
	\begin{subfigure}[t]{\conessize\textwidth}
		\centering
		\includegraphics[width=\textwidth]{figures/lta5_yolov3_yolov3_cone}
		\caption%
		{{\small \yolo{}.}}
		\label{fig:300_cone_yolov3_app}
	\end{subfigure}
	\hfill
	\begin{subfigure}[t]{\conessize\textwidth}
		\centering
		\includegraphics[width=\textwidth]{figures/lta10_maskrcnn_maskrcnn_cone}
		\caption[]%
		{{\small \mrcnn{}.}}
		\label{fig:300_cone_mrcnn_app}
	\end{subfigure}
	\hfill
	\begin{subfigure}[t]{\conessize\textwidth}
		\centering
		\includegraphics[width=\textwidth]{figures/lta10_gtsrbcnn_gtsrbcnn_cone}
		\caption[]%
		{{\small \gtsrb{}.}}
		\label{fig:300_cone_gtsrbcnn_app}
	\end{subfigure}
	\hfill
	\begin{subfigure}[t]{\conessize\textwidth}
		\centering
		\includegraphics[width=\textwidth]{figures/lta5_lisacnn_lisacnn_cone}
		\caption[]%
		{{\small \lisa{}.}}
		\label{fig:300_cone_lisacnn_app}
	\end{subfigure}
	\hfill
	
	\begin{subfigure}[t]{\conessize\textwidth}
		\centering
		\includegraphics[width=\textwidth]{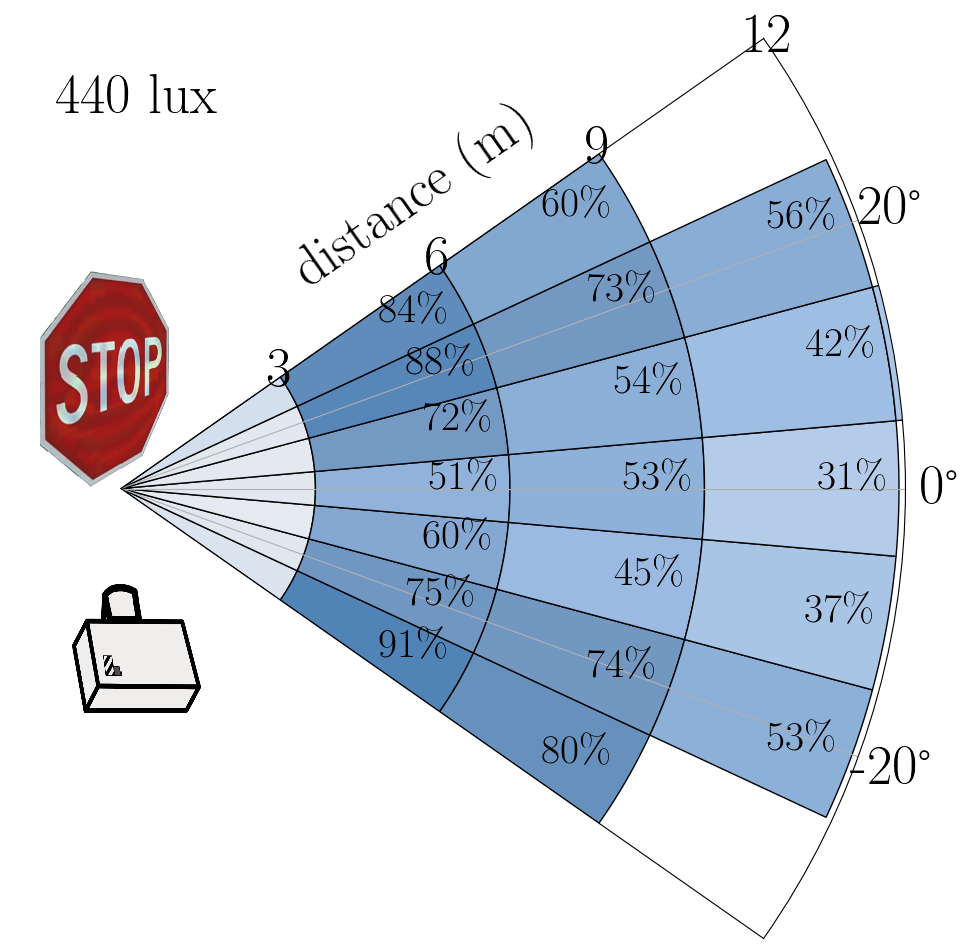}
		\caption%
		{{\small \yolo{}.}}
		\label{fig:440_cone_yolov3_app}
	\end{subfigure}
	\hfill
	\begin{subfigure}[t]{\conessize\textwidth}
		\centering
		\includegraphics[width=\textwidth]{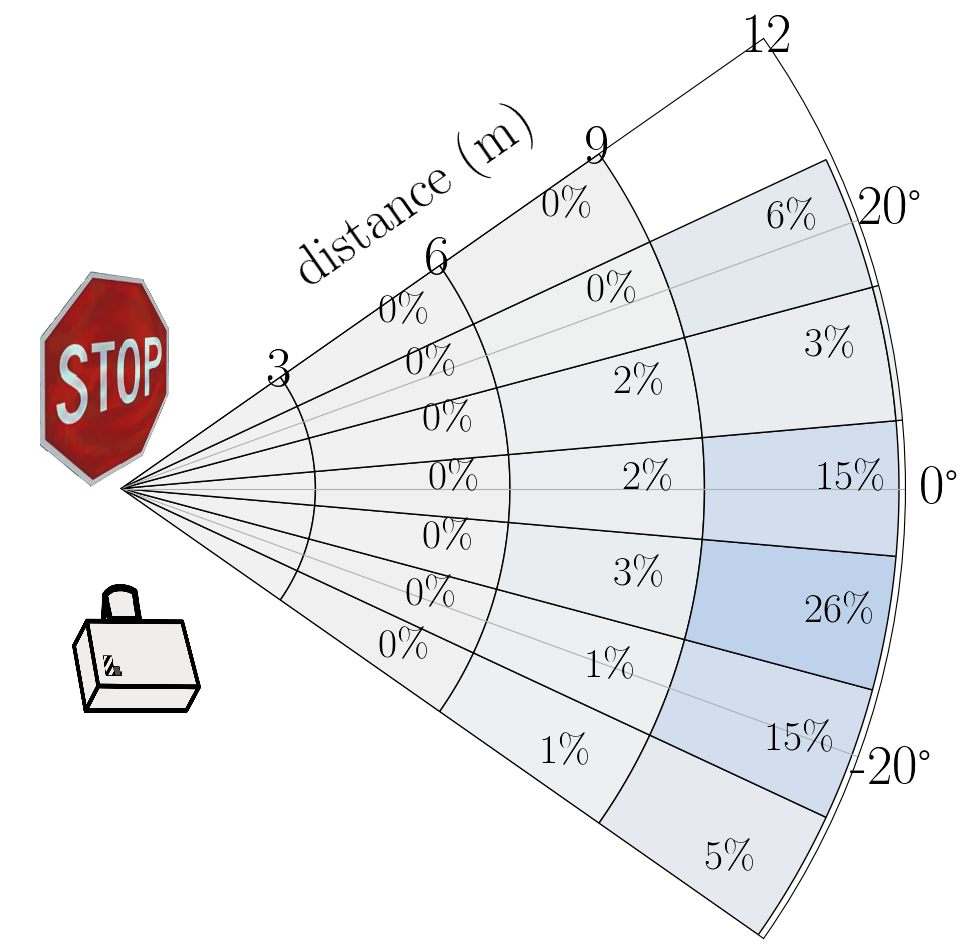}
		\caption[]%
		{{\small \mrcnn{}.}}
		\label{fig:440_cone_mrcnn_app}
	\end{subfigure}
	\hfill
	\begin{subfigure}[t]{\conessize\textwidth}
		\centering
		\includegraphics[width=\textwidth]{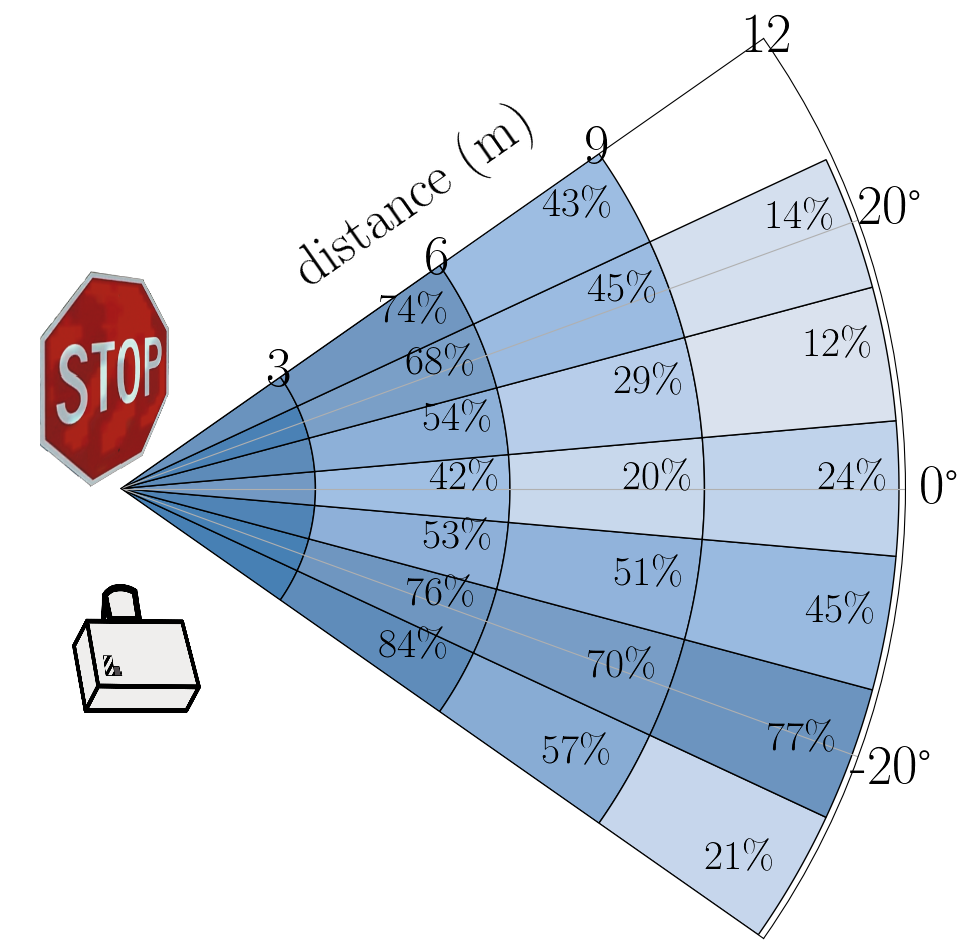}
		\caption[]%
		{{\small \gtsrb{}.}}
		\label{fig:440_cone_gtsrbcnn_app}
	\end{subfigure}
	\hfill
	\begin{subfigure}[t]{\conessize\textwidth}
		\centering
		\includegraphics[width=\textwidth]{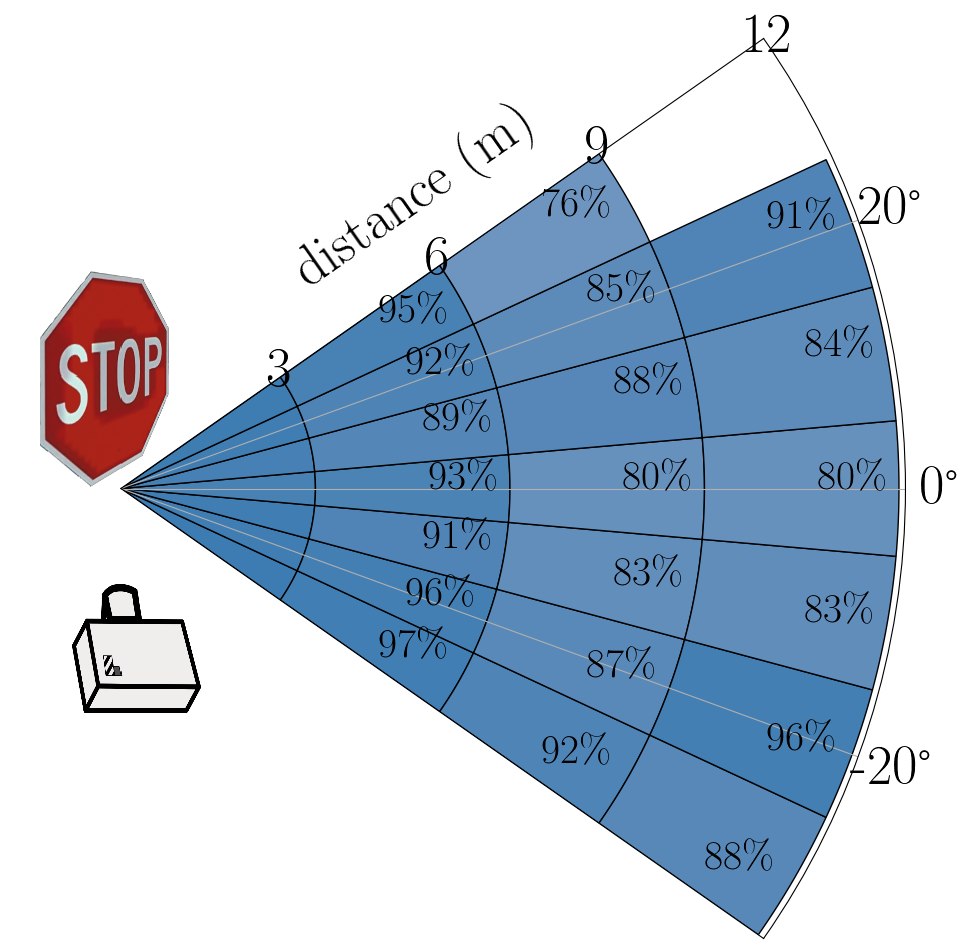}
		\caption[]%
		{{\small \lisa{}.}}
		\label{fig:440_cone_lisacnn_app}
	\end{subfigure}
	\hfill
	
	\begin{subfigure}[b]{\conessize\textwidth}
		\centering
		\includegraphics[width=\textwidth]{figures/lta4_yolov3_yolov3_cone}
		\caption%
		{{\small \yolo{}.}}
		\label{fig:600_cone_yolov3_app}
	\end{subfigure}
	\hfill
	\begin{subfigure}[b]{\conessize\textwidth}
		\centering
		\includegraphics[width=\textwidth]{figures/lta4_maskrcnn_maskrcnn_cone}
		\caption[]%
		{{\small \mrcnn{}.}}
		\label{fig:600_cone_mrcnn_app}
	\end{subfigure}
	\hfill
	\begin{subfigure}[b]{\conessize\textwidth}
		\centering
		\includegraphics[width=\textwidth]{figures/lta4_gtsrbcnn_gtsrbcnn_cone}
		\caption[]%
		{{\small \gtsrb{}.}}
		\label{fig:600_cone_gtsrbcnn_app}
	\end{subfigure}
	\hfill
	\begin{subfigure}[b]{\conessize\textwidth}
		\centering
		\includegraphics[width=\textwidth]{figures/lta4_lisacnn_lisacnn_cone}
		\caption[]%
		{{\small \lisa{}.}}
		\label{fig:600_cone_lisacnn_app}
	\end{subfigure}
	
	\caption{Attack success rates for various viewing angles and distances, for all models and all measured lux settings and the indoors experiment.}
	\label{fig:cone_app}
\end{figure*}

\begin{table*}[t]
	\centering\footnotesize
	\begin{tabular}{c@{\hskip 6pt}c@{\hskip 6pt}c|c@{\hskip 6pt}c@{\hskip 6pt}c@{\hskip 6pt}c@{\hskip 6pt}c@{\hskip 6pt}c@{\hskip 6pt}c@{\hskip 6pt}c@{\hskip 6pt}c}
		& & & \multicolumn{9}{c}{\textit{Target Model}} \\
		\textit{lux} & \textit{Source Model} & \textit{no. frames} & \rotatebox{60}{\yolo{}} & \rotatebox{60}{\mrcnn{}} & \rotatebox{60}{\gtsrb{}} & \rotatebox{60}{$\text{\gtsrb{}}^{(a)}$} & \rotatebox{60}{$\text{\gtsrb{}}^{(s)}$} & \rotatebox{60}{\lisa{}} & \rotatebox{60}{$\text{\lisa{}}^{(a)}$} & \rotatebox{60}{$\text{\lisa{}}^{(s)}$} & \rotatebox{60}{Google Vision*} \\ \toprule 
		\multirow{4}{*}{\textbf{120}}& \yolo{}& 4587 & \textbf{100.0\%}& 73.4\%& 0.0\%& 0.0\%& 0.0\%& 21.5\%& 0.0\%& 0.0\%& 100.0\%\\ 
		& \mrcnn{}& 3765 & 98.7\%& \textbf{97.1\%}& 0.0\%& 0.0\%& 0.0\%& 15.5\%& 0.0\%& 0.0\%& 100.0\%\\ 
		& \gtsrb{}& 3760 & 40.5\%& 37.0\%& \textbf{99.9\%}& 0.0\%& 16.1\%& 51.4\%& 0.0\%& 0.0\%& 72.4\%\\ 
		& \lisa{}& 4998 & 29.4\%& 28.1\%& 6.8\%& 0.0\%& 0.0\%& \textbf{100.0\%}& 0.0\%& 0.0\%& 77.1\%\\ 
		\midrule\multirow{4}{*}{\textbf{180}}& \yolo{}& 7862 & \textbf{99.9\%}& 4.0\%& 14.6\%& 0.0\%& 0.0\%& 17.5\%& 0.0\%& 0.0\%& 90.6\%\\ 
		& \mrcnn{}& 4083 & 96.3\%& \textbf{91.0\%}& 0.2\%& 0.0\%& 0.0\%& 54.8\%& 0.0\%& 0.0\%& 98.9\%\\ 
		& \gtsrb{}& 7426 & 12.3\%& 2.0\%& \textbf{85.7\%}& 0.0\%& 27.7\%& 13.4\%& 0.0\%& 0.0\%& 44.4\%\\ 
		& \lisa{}& 6268 & 9.0\%& 0.6\%& 35.7\%& 0.0\%& 0.0\%& \textbf{100.0\%}& 0.0\%& 0.0\%& 26.2\%\\ 
		\midrule\multirow{4}{*}{\textbf{300}}& \yolo{}& 5169 & \textbf{96.5\%}& 3.6\%& 2.5\%& 0.0\%& 0.0\%& 2.3\%& 0.0\%& 0.0\%& 72.3\%\\ 
		& \mrcnn{}& 3543 & 32.0\%& \textbf{14.0\%}& 0.1\%& 0.0\%& 0.0\%& 10.4\%& 0.0\%& 0.0\%& 65.9\%\\ 
		& \gtsrb{}& 3438 & 2.0\%& 2.9\%& \textbf{48.0\%}& 0.0\%& 43.1\%& 44.0\%& 0.0\%& 0.0\%& 47.6\%\\ 
		& \lisa{}& 4388 & 0.7\%& 4.9\%& 8.6\%& 0.0\%& 0.0\%& \textbf{100.0\%}& 0.0\%& 0.0\%& 25.0\%\\ 
		\midrule\multirow{4}{*}{\textbf{440}}& \yolo{}& 6716 & \textbf{49.5\%}& 0.8\%& 40.3\%& 0.0\%& 0.0\%& 40.9\%& 0.0\%& 0.0\%& 35.3\%\\ 
		& \mrcnn{}& 6023 & 5.4\%& \textbf{3.3\%}& 41.1\%& 0.0\%& 0.0\%& 35.6\%& 0.0\%& 0.0\%& 33.6\%\\ 
		& \gtsrb{}& 6565 & 0.7\%& 0.7\%& \textbf{43.7\%}& 0.0\%& 44.1\%& 35.6\%& 0.0\%& 0.0\%& 33.1\%\\ 
		& \lisa{}& 6287 & 1.0\%& 2.4\%& 26.4\%& 0.0\%& 0.0\%& \textbf{97.4\%}& 0.0\%& 0.0\%& 26.8\%\\ 
		\midrule\multirow{4}{*}{\textbf{600}}& \yolo{}& 5507 & \textbf{17.8\%}& 0.2\%& 32.5\%& 0.0\%& 0.0\%& 27.4\%& 0.0\%& 0.0\%& 23.7\%\\ 
		& \mrcnn{}& 5058 & 0.1\%& \textbf{0.4\%}& 5.3\%& 0.0\%& 0.0\%& 4.6\%& 0.0\%& 0.0\%& 16.7\%\\ 
		& \gtsrb{}& 4637 & 0.0\%& 0.9\%& \textbf{7.2\%}& 0.0\%& 7.5\%& 4.9\%& 0.0\%& 0.0\%& 21.1\%\\ 
		& \lisa{}& 4714 & 0.0\%& 0.9\%& 8.6\%& 0.0\%& 0.0\%& \textbf{57.5\%}& 0.0\%& 0.0\%& 15.8\%\\ 
		\bottomrule 
	\end{tabular}
	\caption{
		Transferability results. We test all the frames from the collected videos with a certain projection being shone against a
		different target model, figures in bold are white-box pairs. (*) For Google Vision we only test one frame every 30 frames,
		i.e., one per second. We also remove all frames that are further than 6m away as Google Vision does not detect most of  
		them in a baseline scenario. Models indicated with $\_^{(a)}$ indicate adversarially trained models.  
		Models indicated with $\_^{(s)}$ indicate models we re-trained from scratch. }	
	\label{tab:transferability_app}
\end{table*}

\end{document}